\documentclass{article}

\usepackage{arxiv}

\usepackage[utf8]{inputenc} % allow utf-8 input
\usepackage[T1]{fontenc}    % use 8-bit T1 fonts
\usepackage{hyperref}       % hyperlinks
\usepackage{url}            % simple URL typesetting
\usepackage{booktabs}       % professional-quality tables
\usepackage{amsfonts}       % blackboard math symbols
\usepackage{nicefrac}       % compact symbols for 1/2, etc.
\usepackage{microtype}      % microtypography
\usepackage{amsmath} % for symbols like \uparrow and \downarrow

\usepackage{cleveref}       % smart cross-referencing
\usepackage{lipsum}         % Can be removed after putting your text content
\usepackage{graphicx}
\usepackage{natbib}
\usepackage{doi}

\usepackage{comment}
\usepackage{listings}
\usepackage{booktabs}
\usepackage{multirow}
\usepackage{amsfonts}
\usepackage{pifont} % For check and X marks
\usepackage{colortbl} % For color
\usepackage{tabularray}
\usepackage{titletoc}
\usepackage{amssymb}

\usepackage{tcolorbox}
\tcbuselibrary{breakable}
\usepackage{caption}

\usepackage{titletoc} % add this once

\newcommand{\cmark}{{\ding{51}}} % checkmark
\newcommand{\xmark}{{\ding{55}}} %  X mark

\usepackage{url} % NeurIPS already includes this, but harmless to repeat

\newenvironment{links}{%
  \begin{center}
  \begin{minipage}{0.85\linewidth} % adjust width: 0.6–0.9\linewidth works well
  \begin{list}{}{\setlength\itemsep{0pt}\setlength\leftmargin{0pt}}
}{%
  \end{list}
  \end{minipage}
  \end{center}
}
\newcommand{\link}[2]{\item \textbf{#1:} \url{#2}}

\title{EgoEMS: A High-Fidelity Multimodal Egocentric Dataset for Cognitive Assistance in \\ Emergency Medical Services}

\usepackage{authblk}

\newbox{\orcid}\sbox{\orcid}{\includegraphics[scale=0.06]{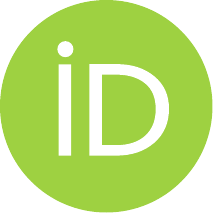}} % optional icon; remove if not needed

\author[1]{Keshara Weerasinghe}
\author[1]{Xueren Ge}
\author[1]{Tessa Heick}
\author[1]{Lahiru Nuwan Wijayasingha}
\author[2]{Anthony Cortez}
\author[1]{\\Abhishek Satpathy}
\author[1]{John Stankovic}
\author[1]{Homa Alemzadeh}

\affil[1]{School of Engineering and Applied Science}
\affil[2]{School of Medicine}
\affil[ ]{University of Virginia}
\affil[ ]{\texttt{\{cjh9fw, zar8jw, vht2gm, lnw8px, aec3gp, cqa3ym, jas9f, ha4d\}@virginia.edu}}

\renewcommand{\shorttitle}{}

\hypersetup{
    pdftitle={EgoEMS: A High-Fidelity Multimodal Egocentric Dataset for Cognitive Assistance in Emergency Medical Services},
    pdfsubject={cs.CV, cs.HC, cs.AI},
    pdfauthor={Keshara Weerasinghe, Xueren Ge, Tessa Heick, Lahiru Nuwan Wijayasingha, Anthony Cortez, Abhishek Satpathy, John A. Stankovic, Homa Alemzadeh},
    pdfkeywords={Egocentric video, Emergency Medical Services, Multimodal dataset, Cognitive assistance, Action recognition, CPR quality estimation}
}

\usepackage{chngcntr}

\newcommand{\appendixformat}{
  \appendix
  \renewcommand{\thesection}{\Alph{section}}
  \renewcommand{\thesubsection}{\thesection.\arabic{subsection}}
  \renewcommand{\thesubsubsection}{\thesubsection.\arabic{subsubsection}}
}

\begin{document}
\maketitle

\begin{abstract}
Emergency Medical Services (EMS) are critical to patient survival in emergencies, but first responders often face intense cognitive demands in high-stakes situations. AI cognitive assistants, acting as virtual partners, have the potential to ease this burden by supporting real-time data collection and decision making. In pursuit of this vision, we introduce EgoEMS, the first end-to-end, high-fidelity, multimodal, multiperson dataset capturing over 20 hours of realistic, procedural EMS activities from an egocentric view in 233 simulated emergency scenarios performed by 62 participants, including 46 EMS professionals. Developed in collaboration with EMS experts and aligned with national standards, EgoEMS is captured using an open-source, low-cost, and replicable data collection system and is annotated with keysteps, timestamped audio transcripts with speaker diarization, action quality metrics, and bounding boxes with segmentation masks. Emphasizing realism, the dataset includes responder-patient interactions reflecting real-world emergency dynamics. 
We also present a suite of benchmarks for real-time multimodal keystep recognition and action quality estimation, essential for developing AI support tools for EMS. 
We hope EgoEMS inspires the research community to push the boundaries of intelligent EMS systems and ultimately contribute to improved patient outcomes.
\end{abstract}

\begin{links}
    \link{Code \& Dataset}{https://uva-dsa.github.io/EgoEMS}
    % \link{Code}{https://github.com/UVA-DSA/EgoEMS}
    % \link{Datasets}{https://uva-dsa.github.io/EgoEMS/}
\end{links}

% keywords can be removed

% ==== INTRODUCTION ===== %

\section{Introduction}
\label{main:intro}

Every year more than 28 million emergency medical incidents are responded to in the U.S. \cite{nasemso2020emsassessment}. Upon arrival at an incident scene, 
Emergency Medical Services (EMS)
personnel must rapidly assess the situation, process complex information about victims and the environment, and provide emergency care before transferring patients to the hospital. In these safety-critical scenarios
, patient survival hinges on rapid and accurate
decision making. 
However, EMS responders often face overwhelming physical, mental, and emotional demands, resulting in cognitive overload, burnout, and increased risk of errors \cite{cognitiveload, crowe2018association}. 

With the recent rise of embedded Artificial Intelligence (AI) and large language models (LLMs) along with the rapid advancements of Augmented Reality (AR) technologies, there is tremendous potential to develop Intelligent Cognitive Assistants (ICAs) that can act as virtual partners to enhance situational awareness, guide critical procedures, and support training~\cite{preum2021review}. Yet, the medical and EMS domains remain significantly underserved due to a lack of large-scale, high-fidelity labeled datasets and major privacy and security challenges.

Recent works have proposed innovative support systems and technologies to aid first responders in high-stakes environments.
One example is the development of ICAs for real-time diagnosis and treatment decision support~\cite{preum2021review,jin2023emsassist,weerasinghe2024real,preum2019cognitiveems,preum2018towards,shu2019behavior, ge-etal-2024-dkec}. These systems can also serve as virtual coaches for training, helping novice responders build expertise through real-time feedback. However, existing ICAs are typically developed using datasets with limited fidelity and single modalities (primarily speech), which fail to capture the procedural complexity and unpredictability of real-world EMS settings.
To provide accurate predictions and effective support in practical scenarios, ICAs must perceive the environment through multimodal sensing and interpret multiple responder activities from a first-person perspective in real-time. 

\begin{figure*}[t!]
    \centering
        \includegraphics[width=\textwidth]{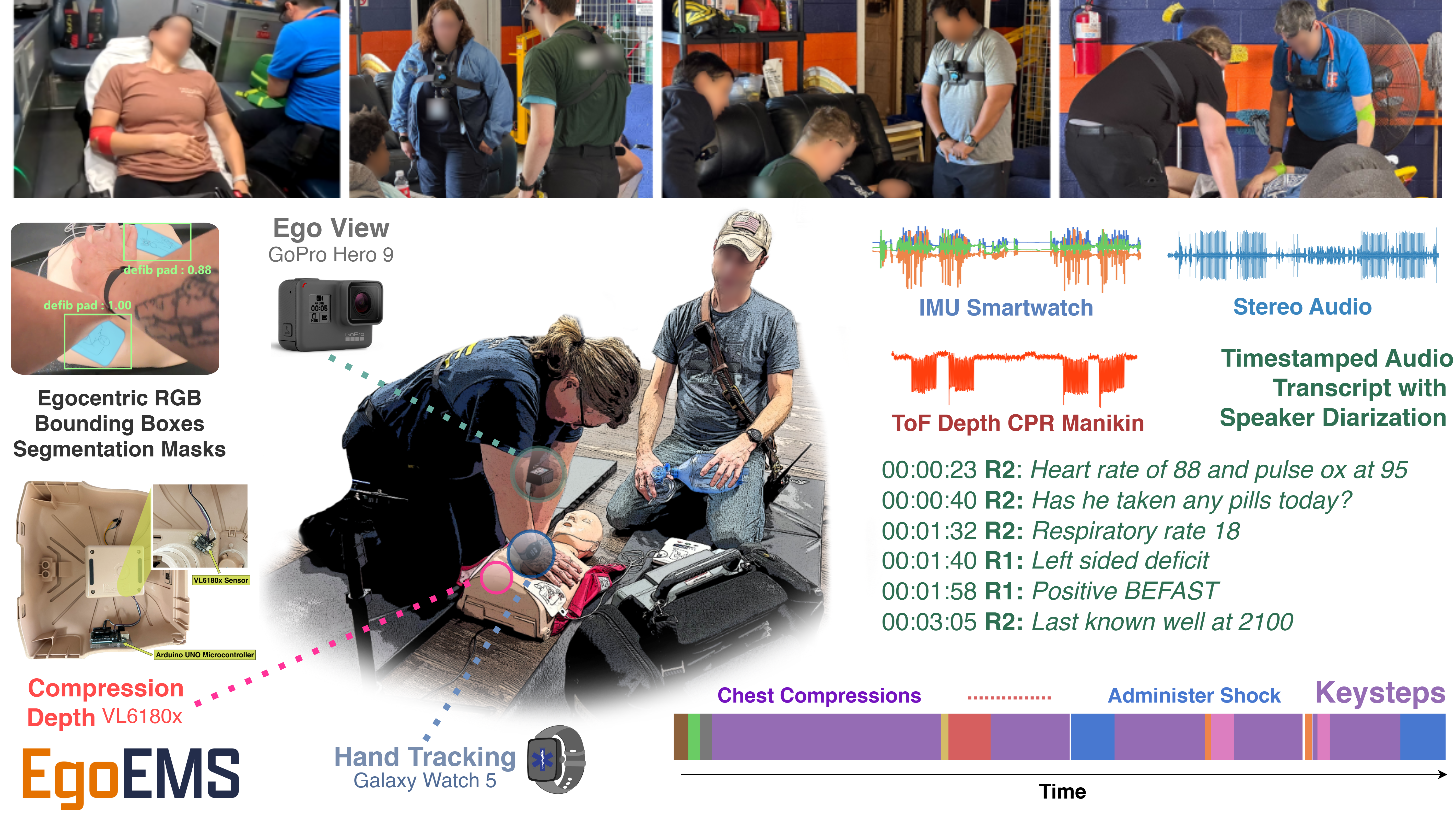}
    \caption{EgoEMS dataset provides synchronized egocentric multiperson views, along with rich high-fidelity multimodal data capturing EMS professionals in highly procedural tasks resulting in a total of 233 trials (20 hours) including 2694 keystep instances from commonly executed EMS interventions. The dataset includes annotations of keysteps, timestamped audio transcripts with speaker diarization, and semi-automatically generated bounding boxes and segmentation masks for key EMS objects, offering a comprehensive resource for understanding EMS workflows and developing AI solutions.
    }
    \label{fig:main}
\end{figure*}

Advances in egocentric datasets~\cite{yang2025egolife,liu2022hoi4d,bansal2022my,wang2024egovid} encourage the development of personal AI assistants for daily life and procedural task automation. Unlike exocentric (third-person) recordings, egocentric perspectives align more naturally with wearable AI systems and are particularly effective at capturing fine-grained hand-object interactions that are often occluded or outside the field of view in external camera setups~\cite{bansal2022my}.  However, existing datasets  mostly focus on routine daily activities, lack multimodal integration, and are not designed for high-stakes domains like emergency medicine, where actions are complex, time-critical, and performed by teams.
Ego-Exo4D~\cite{grauman2024ego} includes a limited number of medical procedures, such as Cardiopulmonary Resuscitation (CPR) and COVID-19 testing. Other datasets, like EgoSurgery~\cite{fujii2024egosurgery} and Trauma Thompson~\cite{birch2023trauma}, although focused on health domains, remain narrow in scope and limited to single views, specific procedures, and single modalities. 
Synthetic datasets, such as~\cite{wang2023pov}, instead emphasize hand–tool pose estimation in controlled surgical scenes, but lack realism, procedural breadth, and real-world variability. %of real-world emergencies. 

EMS incidents unfold in safety-critical environments where a team of first responders performs coordinated \textit{keysteps} within standard procedures (\textit{EMS interventions}) as defined by established \textit{EMS protocol} guidelines under strict time constraints.
These settings inherently generate rich multimodal information (e.g., egocentric views of medical interventions, conversational audio, and motion data) that can be leveraged for real-time decision support.
For instance, Cardiac Arrest protocol requires maintaining correct CPR compression rate and depth for patient survival~\cite{odemsa,AmericanRedCross2022Steps} and proper timing of ventilations (e.g., 30 compressions to 2 breaths) for adequate oxygenation (see Appendix A for an overview of CPR procedure).  Similarly, in stroke emergencies, rapid diagnosis and hospital transfer within clinically recommended time windows are required to minimize brain damage and improve outcomes~\cite{saver2006time,odemsa}. Although responders are trained on such guidelines, stress and cognitive load can impact their performance. ICAs can help track procedural keysteps and quality metrics, 
providing real-time reminders or feedback to improve adherence to the protocols. However, the development of such systems is hindered by the absence of suitable datasets. 

To address the aforementioned gaps, we introduce \textit{EgoEMS, the first high-fidelity egocentric dataset designed specifically for cognitive assistance in EMS.} EgoEMS captures 20 hours of synchronized and labeled multimodal
data of multiperson interactions in end-to-end EMS workflows from initial patient assessment to intervention, involving 62 subjects with varying skill levels (including EMS professionals and members of the public) performing 233 trials. Unlike prior datasets, EgoEMS is structured around nationally standardized EMS protocols (in the U.S.), spanning some of the most common EMS scenarios (cardiac arrest, cardiac suspected and stroke) and 9 critical interventions (including Airway-Breathing-Circulation (ABCs), 12-lead Electrocardiogram (ECG), CPR, Ventilation, Defibrillation, Stroke Assessment, Patient History, Vital Sign Assessment and Transport) according to NEMSIS \cite{nemsis} database and expert feedback.

The dataset provides 
(i) egocentric views of the scene captured from responders’ body‑worn cameras, 
(ii) audio recordings of conversations at the scene,
(iii) smartwatch IMU data capturing the responder’s hand movements, and (iv) ground-truth quality metrics (compression rate and depth during CPR procedures), synchronously collected using off-the-shelf components
and custom open-source software (see Figure \ref{fig:main}). We also provide annotations for EMS keysteps performed by responders, timestamped transcripts of responders' conversations with speaker diarization (i.e., automatic labeling of “who spoke when” in multi-speaker audio), and object bounding boxes with segmentation masks for the medical tools used by the responders in cardiac arrest emergencies.
These annotations are generated using manual and semi-automatic approaches, based on the NREMT \cite{nremt} 
guidelines and in collaboration with EMS experts (see Figure~\ref{fig:taxonomy}). Together, these elements enable end-to-end modeling of emergency response scenarios, from high-level protocol decisions to fine-grained action execution, with the hierarchical taxonomy providing structured representations of realistic EMS workflows.

We also present three benchmark tasks (see Figure \ref{fig:benchmarks}), that reflect core real-time context inference capabilities essential for ICAs to support EMS responders: \textit{Keystep Classification} for recognizing the specific keystep performed by a responder, \textit{Keystep Segmentation} for detecting the start and end times of each keystep, and \textit{Quality Evaluation} by continuous estimation of activity quality metrics (e.g., CPR compression rate and depth) utilizing multimodal data to provide feedback to responders.

In summary, our contributions are the following:
\begin{itemize}
    \item \textit{The first synchronized and labeled multimodal dataset of %single and 
    multiperson EMS procedural activities}, %developed in collaboration with EMS professionals, 
    capturing collaborative dynamics of real-world scenarios with varied experience levels and certifications of EMS personnel.

    \item \textit{A taxonomy of EMS activities, keysteps, and objects/tools}, developed in collaboration with EMS professionals and aligned with NREMT,
    which is used to generate ground-truth annotations for activity recognition, 
    and object detection and segmentation along with audio transcription and speaker diarization.

    \item \textit{A suite of benchmarks for real-time activity recognition and quality estimation}, leveraging both single and combined modalities,
    % each modality and different combinations of multiple modalities 
    to explore the performance of state-of-the-art (SOTA) supervised deep learning models compared to zero-shot methods including LLMs.

    \item \textit{An open-source, low-cost, and easily replicable multimodal data collection system} based on off-the-shelf 
    % sensors 
    devices
    (e.g., GoPro Hero) and custom hardware
    integration
    (e.g., VL6180X ToF sensor) for synchronized capture of procedural activities through egocentric video and conversational audio recordings from the scene, smartwatch IMU data from hand movements, and ground-truth quality metrics from patient simulators.
    
\end{itemize}

% ========== END OF INTRODUCTION ============== %

% ========== BEGIN RELATED WORKS ============== %

\begin{table*}[t]
\centering
\setlength{\tabcolsep}{1mm}
\small
\resizebox{\columnwidth}{!}{%
\begin{tabular}{@{}lccc|ccc|ccccc@{}}
\toprule
\textbf{Dataset}  & {\shortstack{\textbf{Activity} \\ \textbf{Setting}}}  & \textbf{Synced} & {\textbf{MP}}    & \textbf{IMU} & \textbf{Audio} & \textbf{RGB} & \shortstack{\textbf{Fine} \\ \textbf{Act.}} & \shortstack{\textbf{Tran-} \\ \textbf{scripts}} & \shortstack{\textbf{Obj.} \\ \textbf{BB}}& \textbf{Skill}  & \shortstack{\textbf{Dur.} \\ (hrs)} \\
\midrule
Epic-Kitchens~\cite{damen2018scaling}  & Daily Life   & \xmark &  \xmark   & \xmark & \cmark & \cmark & \cmark & \cmark & \xmark & \xmark & 100 \\

HOI4D~\cite{liu2022hoi4d}  & Object Manipulation   & \xmark &  \xmark   & \xmark & \xmark & \cmark & \cmark & \xmark & \cmark & \xmark & 44.4 \\
% Ego4D~\cite{grauman2022ego4d}     & \cmark  & \cmark & Daily Life  &  \cmark  & \xmark & \cmark & \cmark & \cmark & \cmark & \cmark & \xmark \\
EgoProceL~\cite{bansal2022my} & Daily Life  & \xmark & \xmark  & \xmark & \xmark & \cmark & \cmark & \xmark & \xmark & \xmark & 62 \\
HoloAssist~\cite{wang2023holoassist}  & Daily Life  & \cmark & \cmark  & \cmark & \cmark & \cmark & \cmark & \cmark & \xmark & \cmark & 166 \\
EgoVid-5M~\cite{wang2024egovid}& Daily Life  &  \xmark  & \xmark & \xmark & \xmark & \cmark & \cmark & \xmark & \xmark & \xmark  & 5550 \\
EgoLife~\cite{yang2025egolife}  & Daily Life &  \cmark & \cmark & \xmark & \cmark & \cmark & \cmark & \cmark & \xmark & \xmark & 300 \\
% EgoExoLearn~\cite{huang2024egoexolearn} & Daily/Lab   & \cmark  & \xmark  & \xmark & \xmark & \cmark & \cmark & \xmark & \xmark & \cmark & 120 \\
Ego-Exo4D~\cite{grauman2024ego} & Skilled Activities  &  \cmark & \xmark  & \cmark & \cmark & \cmark & \cmark & \cmark & \cmark & \cmark & 1442 \\
\midrule
Trauma-Thompson~\cite{birch2023trauma}& Medical Emergency &  \xmark  & \xmark   & \xmark & \xmark & \cmark & \cmark & \xmark & \xmark & \xmark & $\sim$1.6 \\
EgoSurgery~\cite{fujii2024egosurgery} & Surgery  & \xmark  & \cmark   & \xmark & \xmark & \cmark & \cmark & \xmark & \xmark & \xmark & 15 \\
% EgoFalls~\cite{wang2023egofalls} & Mobility & \xmark  & \xmark   & \xmark & \cmark & \cmark & \cmark & \xmark & \xmark & \xmark & 15 \\
POV-Surgery~\cite{wang2023pov}  & Synthetic Surgery  & \xmark & \xmark & \xmark & \xmark & \cmark & \xmark & \xmark & \cmark & \xmark & $\sim$1\\
\midrule
\textbf{EgoEMS (Ours)}  & \textbf{Medical Emergency} &  \cmark & \cmark  & \cmark & \cmark & \cmark & \cmark & \cmark & \cmark & \cmark & \textbf{20} \\
\bottomrule
\end{tabular}
}
\caption{Comparison of egocentric datasets by modality availability and annotation types. Synced: Synchronized modalities, MP: Multiperson, Obj. BB: Object bounding boxes and segmentation masks,  Fine Act: Fine-grained action annotations, Skill: Ground-truth for skill estimation, Dur: Approximate dataset duration ($\sim$ estimated from reported frame counts at 30fps)}
\label{tab:related_works}
\end{table*}
\vspace{-1em}

\section{Background and Related Work}
\label{main:related-works}

\noindent\textbf{Egocentric datasets.}
AI assistants that support real-world decision making require multimodal understanding of human activities from a first-person perspective~\cite{preum2021review}. Existing egocentric datasets such as Epic-Kitchens~\cite{damen2018scaling}, HOI4D~\cite{liu2022hoi4d}, HoloAssist~\cite{wang2023holoassist}, EgoVid-5M~\cite{wang2024egovid}, EgoProceL~\cite{bansal2022my}, and EgoLife~\cite{yang2025egolife} largely capture daily activities, object interactions, or scripted behaviors. Ego-Exo4D~\cite{grauman2024ego} combines egocentric and exocentric views and includes limited medical procedures (e.g., CPR, COVID testing) performed by participants with basic training and accredited nurses.
Other egocentric datasets specifically focused on the medical domain include POV-Surgery~\cite{wang2023pov} and 
EgoSurgery~\cite{fujii2024egosurgery} for open surgery 
and Trauma Thompson~\cite{birch2023trauma} for life-saving interventions such as tube thoracostomy and tracheostomy.

Despite these advances, none of the existing datasets %are confined to routine tasks, controlled surgical scenes, or patient-centric contexts, and none %
capture the procedural structure, high‑stakes environment, and coordinated multiperson nature of end‑to‑end EMS workflows (see Table \ref{tab:related_works}). 
Curation of such datasets is particularly challenging due to privacy concerns and high annotation costs in medical settings, resulting in much smaller datasets. 
No prior dataset offers multimodal, multiperson egocentric recordings of simulated emergencies with certified responders and the ground truth necessary for modeling decision‑making and cognitive assistance in critical care.

\noindent\textbf{Activity recognition.}
Action recognition spans both classification, which assigns labels to video segments, and segmentation, which temporally localizes and labels actions over time. Prior work on egocentric video has addressed keystep or action classification
by leveraging supervised deep learning models \cite{plizzari2023can, dessalene2023therbligs, plizzari2022e2, escorcia2022sos, bansal2022my, grauman2024ego}, with fewer efforts leveraging multimodal fusion of video and audio \cite{radevski2023multimodal, gong2023mmg}. Segmentation has also been widely studied \cite{zhang2022actionformer, yi2021asformer, li2020ms, lea2017temporal, wang2016temporal}, though only a few works consider egocentric multimodal settings by combining video, audio, and IMU signals \cite{grauman2024ego, huang2024egoexolearn}. 

In this paper, we take the first step towards benchmarking SOTA deep learning models for multimodal EMS activity recognition. We evaluate diverse representative models, covering supervised (unimodal and multimodal), cross-domain few-shot, and zero-shot (LLM) approaches. See EgoEMS Benchmarks Section and Appendix D for more details.

% In this paper, we take the first step towards benchmarking SOTA deep learning models for activity recognition using multimodal data for the EMS domain.
% We select strong, representative baselines that cover complementary approaches, including supervised models based on transformer architectures \cite{gberta_2021_ICML,weerasinghe2024multimodal} capable of modeling long temporal dependencies and multimodal fusion, a convolutional TSN model \cite{wang2016temporal} as a widely used CNN baseline, and zero-shot vision–language models such as Qwen-2.5 \cite{bai2025qwen2} and VideoLLaMA-3.3 \cite{zhang2025videollama} to assess the potential of large pretrained models. For audio-only recognition, we also include a custom zero-shot pipeline using WhisperTimestamped \cite{lintoai2023whispertimestamped} and GPT-4o \cite{achiam2023gpt}. 

\noindent\textbf{CPR quality estimation.}
CPR is one of the most safety-critical interventions in emergency care, where proper compression rate and depth are vital for patient survival~\cite{Ayala2014AutomaticParameters,Eftestl2020AArrest}.
Although these metrics are central to effective feedback~\cite{Cheng2015VariabilityInstitutions,Webber2019PaediatricLifeguards,Cheng2015PerceptionRole}, human feedback is often biased~\cite{Jones2015VisualMatter}.
Recent work has explored skill assessment from egocentric video for general activities~\cite{grauman2024ego,huang2024egoexolearn}, but CPR quality has largely been measured with accelerometer equipped CPRcards~\cite{Cheng2015ImprovingTrial,LaerdalCPRcard}, defibrillator pads~\cite{Gonzalez-Otero2015ChestImpedance}, or smartwatch IMU models sensitive to surface conditions~\cite{Lu2018ASmartwatch,jeong2015smartwatch}. Vision based methods using depth cameras~\cite{loconsole2016relive,di2019detecting} require costly sensors in controlled, single-responder settings. In contrast, we introduce the first benchmark for quantitative, online CPR quality estimation that fuses egocentric video and smartwatch IMU to robustly predict compression rate and depth and enable real-time feedback in realistic, emergency scenarios.

% ========== END OF RELATED WORKS ============== %

% ========== START OF METHOD ============== %

\section{EgoEMS Dataset}
\label{sec:dataset}

% bullet list
This section outlines the development of the EgoEMS dataset, including EMS taxonomy, simulation experiments, participants, data collection system and annotations. We refer the reader to Appendices A-C for more details.

\subsection{EMS Taxonomy}
\label{sec:taxonomy}
\begin{figure}[!h]
    \centering
                \includegraphics[width=0.6\linewidth]{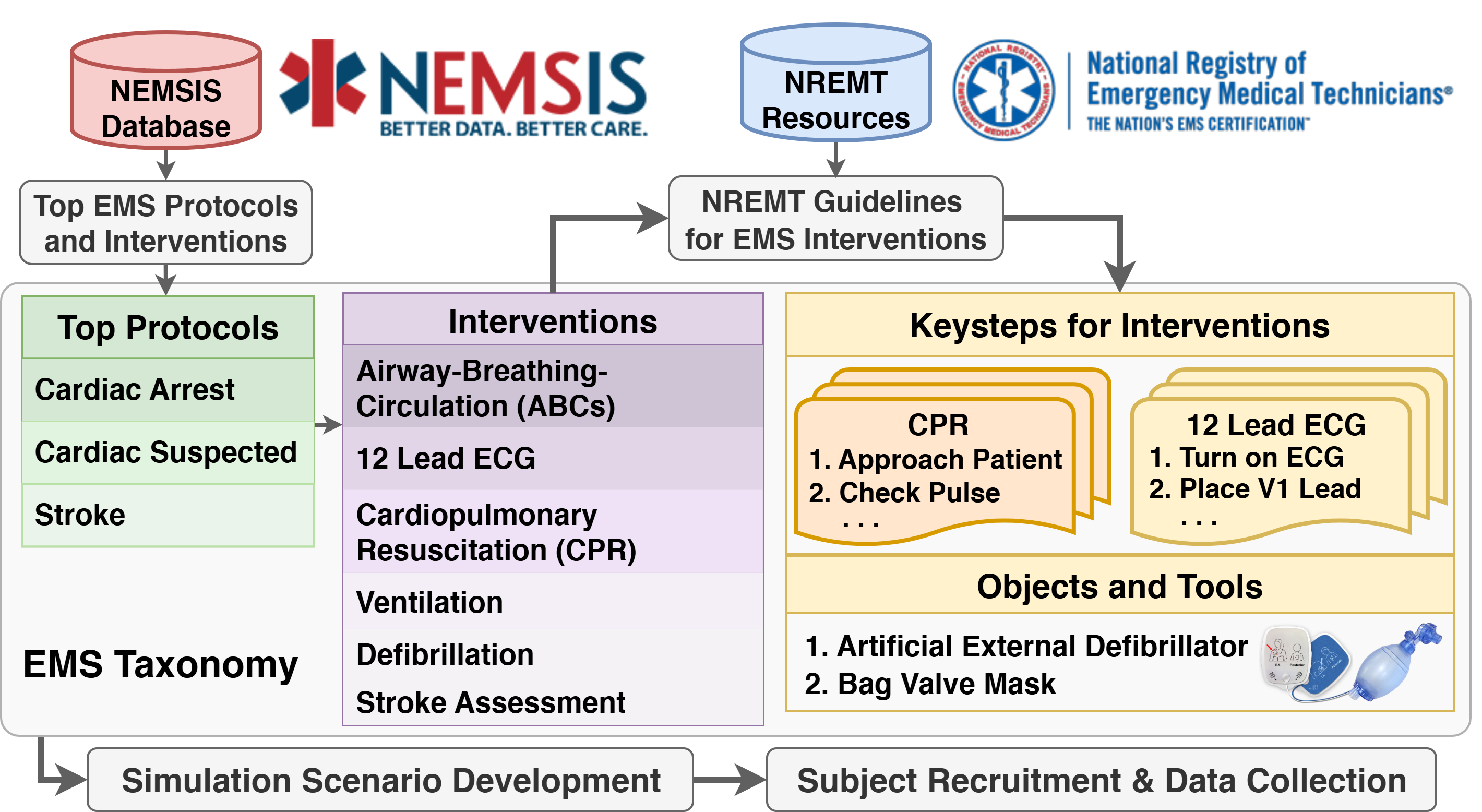}
    \caption{Methodology for creating the EMS taxonomy and EgoEMS simulation scenarios.}
    \label{fig:taxonomy}
\end{figure}

To design a set of realistic EMS scenarios for data collection and adapt activity recognition models to EMS domain, we constructed a taxonomy of hierarchical EMS procedures, capturing high-level EMS protocols and their associated interventions and fine-grained keysteps (see Figure~\ref{fig:taxonomy}).
First, we analyzed the NEMSIS database in consultation with EMS professionals to prioritize protocols that are high-frequency, time-sensitive and critical to patient survival. We focused on cardiac and stroke emergencies and further examined the distribution of interventions within those protocols to isolate most frequently performed interventions. 

Specifically, we focus on scenarios involving the 
``Cardiac Arrest'', ``Chest-Pain Cardiac Suspected'', ``Stroke'' protocols \cite{odemsa} and 9 critical interventions associated with these protocols, including ABCs, 12-lead ECG, CPR, Ventilation, Defibrillation, Stroke Assessment, Patient History, Vital Signs Assessment and Transport. In collaboration with EMS experts, we define key procedural steps (keysteps) for selected interventions based on psychomotor examination guidelines from the NREMT, resulting in a total of 67 detailed keysteps representing the interventions. These keysteps are essential for detecting responders' actions within a protocol and evaluating proper procedural execution to provide continuous feedback. Finally, we identify the key EMS objects/tools used in Cardiac Arrest protocol as a part of the taxonomy, emphasizing the responder-equipment interactions that can improve EMS activity recognition. More details are in Appendix A.

\subsection{Simulation Experiments}

We conducted 233 simulated EMS trials spanning approximately 20 hours, encompassing a range of 
cardiac and stroke related emergency scenarios.
The simulations were divided into two primary types: (i) high-fidelity scenarios, where procedures were performed on human actors portraying patients, and (ii) cardiac arrest scenarios, which used manikins for performing CPR interventions.  High-fidelity simulations were designed to emphasize realism in settings and workflows (e.g., ambulance transfers, in-vehicle recordings with equipment noise), human dynamics (e.g., interfering bystanders), and patient variability, incorporating diverse medical histories and demographics to enhance data diversity. %High-fidelity simulations featured patients with diverse medical histories and demographics to promote data diversity and generalizability.
Each simulation captured end-to-end EMS procedures performed by a team of 2–3 responders, 
performing the critical interventions shown in Table~\ref{annotation_stats}.
In addition, several scenarios were designed to reflect complex, realistic cases in which the initial chief complaint and presenting neurological symptoms mimicked a stroke but were ultimately attributable to alternative causes (e.g., hypoglycemia). These cases required responders to accurately assess, differentiate, and respond using appropriate protocol guided decision making. In cardiac arrest trials, EMS responders typically operated in pairs designated as \textit{primary} and \textit{secondary} responders carrying out critical interventions such as CPR, ventilation, and defibrillation on a manikin. In contrast, subjects from the general public conducted the trials individually and performed only CPR intervention on a manikin due to lack of medical training. To further enhance realism, additional volunteers served as bystanders, particularly in scenarios where the patient was unresponsive, contributing information such as patient history and situational context. 

\begin{table}[t!]
\centering
\setlength{\tabcolsep}{1mm}
\small
\resizebox{\columnwidth}{!}{%
\begin{tabular}{@{}lclll|c|cccc@{}}
\toprule
\multirow{2}{*}{\textbf{Source}} & \multirow{2}{*}{\textbf{Subjects}} & \multirow{2}{*}{\textbf{Scenario}} & \multirow{2}{*}{\textbf{Interventions}} & \multirow{2}{*}{\textbf{Trials (Minutes)}} & \multicolumn{1}{c|}{\textbf{Man. Ann.}} & \multicolumn{4}{c}{\textbf{Semi-auto Ann.}} \\
\cmidrule(lr){6-6} \cmidrule(l){7-10}
 &  &  &  &  & \textbf{Keysteps} & \textbf{BB} & \textbf{SM} & \textbf{TT} & \textbf{RD} \\ 
\midrule
\multirow{3}{*}{EMS Responders} 
    & \multirow{3}{*}{46} & Cardiac Arrest     & CPR, Ventilation, Defibrillation & 76 (183) $\bigstar$
 & \cmark & \cmark & \cmark & \cmark & \cmark \\
    &  & Cardiac Suspected & ABCs, 12-lead ECG                & 23 (173) & \cmark & \xmark & \xmark & \cmark & \(\Diamond\) \\
    &  & Stroke            & ABCs, Stroke Assessment          & 41 (735) & \cmark & \xmark & \xmark & \cmark & \(\Diamond\) \\
General Public 
    & 16 & Cardiac Arrest    & CPR                              & 93 (116) $\bigstar$
 & \cmark & \(\Diamond\) & \(\Diamond\) & \(\Diamond\) & \cmark \\
\midrule
Total 
    & 62 &                   &                                   & 233 (1207) & 2694 & 13.7k & 12k & 140 & 169 \\ 
\bottomrule
\end{tabular}
}
\vspace{1em}
\caption{Simulation scenarios, participants, and annotations in the dataset. \(\Diamond\): Not applicable for the activity. \xmark: Not provided due to limited visibility of the objects of interest. BB: Bounding boxes, SM: Segmentation masks, TT: Timestamped audio transcripts, RD: CPR compression rate and depth. $\bigstar$: Trials that used manikins due to the nature of interventions and safety. The rest were high-fidelity with human patient actors.
}
\label{annotation_stats}
\end{table}

\subsubsection{Participants}

A total of 62 participants were recruited, comprising 46 EMS professionals from 4 rescue squads and 16 individuals from the general public affiliated with an academic institution. EMS responders represented a broad range of experience levels and certifications, including members with basic CPR training, Emergency Medical Responders (EMRs), Emergency Medical Technicians (EMTs), and Paramedics. Years of experience ranged from under one year to over 30. 
The complexity of these simulations and the time required to perform them made large‑scale data collection logistically challenging, as participating EMS agencies relied on volunteer personnel, many of whom were often called away for real emergency dispatches. Despite these challenges, the resulting dataset captures a wide range of realistic responder behaviors and skill levels. See Appendix B for more details.

\subsubsection{Privacy and Ethics}  
We obtained IRB approval prior to data collection and adhered to ethical and privacy protection guidelines for human subjects research in all aspects of recruitment, experiments, and data management.
All identifying information, including participants’ names, faces, ID cards, and vehicles' license-plates were removed from video and audio before data release. 
More details on the IRB protocol, de-identification methods, and real-world ethical considerations are in Appendix B and the Ethical Statement.

\subsection{Data Collection System}

To capture EMS procedures, we used a remotely-controlled chest-mounted GoPro HERO camera to record the responder’s egocentric view along with audio.
The responder’s dominant hand motions were tracked using a Samsung Galaxy Watch 5, recording 3-axis accelerometer data. Additionally, in cardiac arrest scenarios, chest compression ground truth metrics were measured using a VL6180X Time-of-Flight (ToF) sensor mounted on the manikin (see Figure~\ref{fig:main}).
Synchronization was done based on Unix timestamps, with multiperson views synchronized along with all modalities, downsampled to align with GoPro’s frame rate, resulting in a fully synchronized multimodal dataset. 
All data collection tools are available as open-source code, allowing others to replicate the system with low-cost and readily available hardware (see Appendix C).

\subsection{Annotations}
We employed manual and semi‑automatic approaches, with some annotations purely manual and others using zero‑shot models with manual verification.
Table \ref{annotation_stats} shows a summary of annotations in the dataset. See Appendix A for details. \\

\begin{figure}[!h]
    \centering
    \includegraphics[width=0.6\columnwidth]{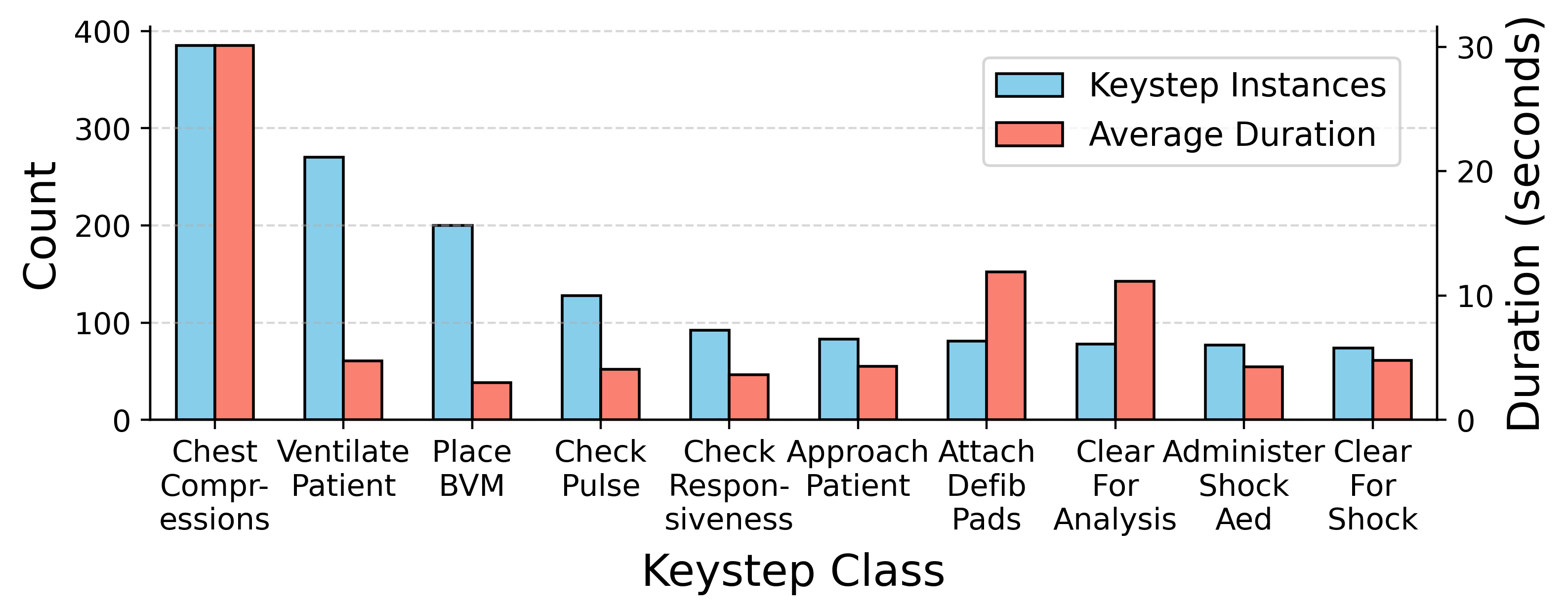}
    \caption{Top-10 keystep distribution with average duration.  
    }
    \label{fig:multi_keysteps_and_keystep_distribution}
\end{figure}

\noindent\textbf{Fine-grained keystep annotation.}  
EgoEMS is manually annotated for 67 keysteps belonging to 9 interventions (see Appendix A.2 and Figure \ref{fig:multi_keysteps_and_keystep_distribution}). The dataset captures multiperson activity typical of real-world EMS procedures, where responders perform concurrent actions during the same time interval (e.g., AED activation during chest compressions; see Appendix A). 
While the dataset includes multiperson annotations, our benchmarks focus on the primary responder and their egocentric viewpoint. We do not leverage secondary responder annotations and multiview data for model training or evaluation, but include them to support future research on multiperson activity understanding.

\noindent \textbf{Timestamped speaker diarization and transcription.} We designed an automated LLM based pipeline to generate speaker-diarized timestamped transcripts for each trial’s audio, which were then manually verified (see Appendix A.4.2). 
The general public participants struggled to narrate while performing CPR due to lack of advanced training, so they were instructed to focus solely on performing the task.

\noindent \textbf{Bounding box and segmentation mask annotations.} We generated bounding box and segmentation mask annotations for medical objects involved in cardiac arrest interventions  (see Figure~\ref{fig:main}) using a semi-automatic pipeline. The nature of these interventions involves frequent and sustained interactions with critical medical equipment, making object localization particularly relevant in this context, serving as a promising candidate for improving the activity recognition capability of an ICA. This pipeline leverages a SOTA object detection model fine-tuned 
based on our EMS taxonomy combined with a zero-shot segmentation method. Our manual verification of 10\% of the bounding box annotations against human annotations shows this method \textit{saves over 60 hours of annotation time at a slight loss of precision.} We refer the reader to Appendix A.4.3 for more detailed analysis.

\begin{figure}[!h] 
    \centering
    \vspace{-1em}
    \includegraphics[width=0.6\linewidth]{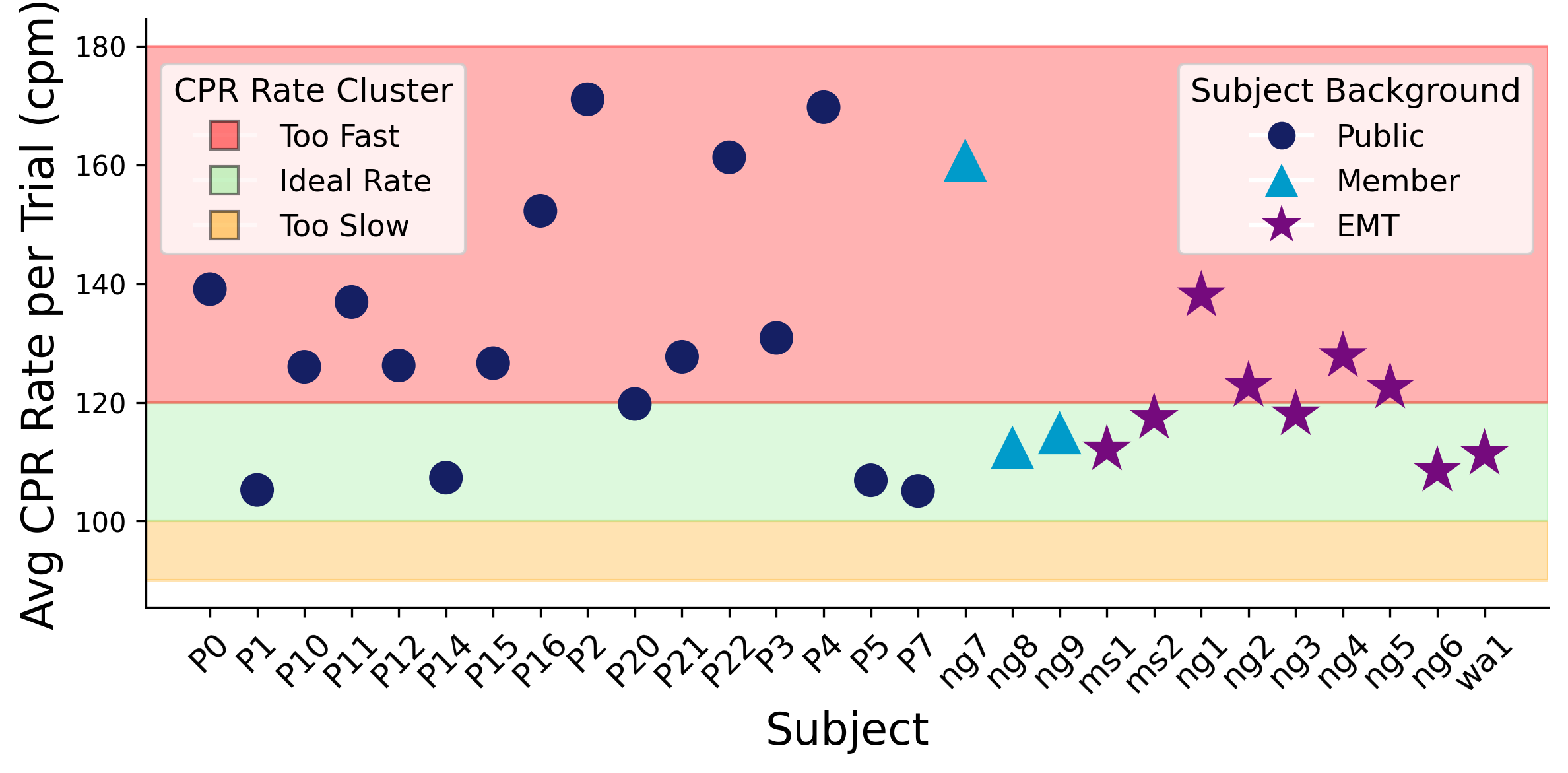}
    \caption{Ground-truth average CPR compression rate per participant across three skill levels (Public, Member, EMT).
    % \homa{Change Y-axis to average CPR rate /min for a trial}
    }
    \label{fig:CPR_actual_averages}
\end{figure}

\noindent \textbf{Compression depth and rate annotations.} The ground-truth CPR depth and rate metrics were automatically generated by recording the compression depth using a ToF sensor integrated with a microcontroller embedded in the manikin (see Figure \ref{fig:main}). 
Figure \ref{fig:CPR_actual_averages} shows the CPR rate distributions across skill levels where EMS professionals maintained steady CPR performance, while novice public participants showed much greater variability.

% ========== END OF METHOD ============== %

\begin{figure*}[t!]
    \centering
    \includegraphics[width=\textwidth]{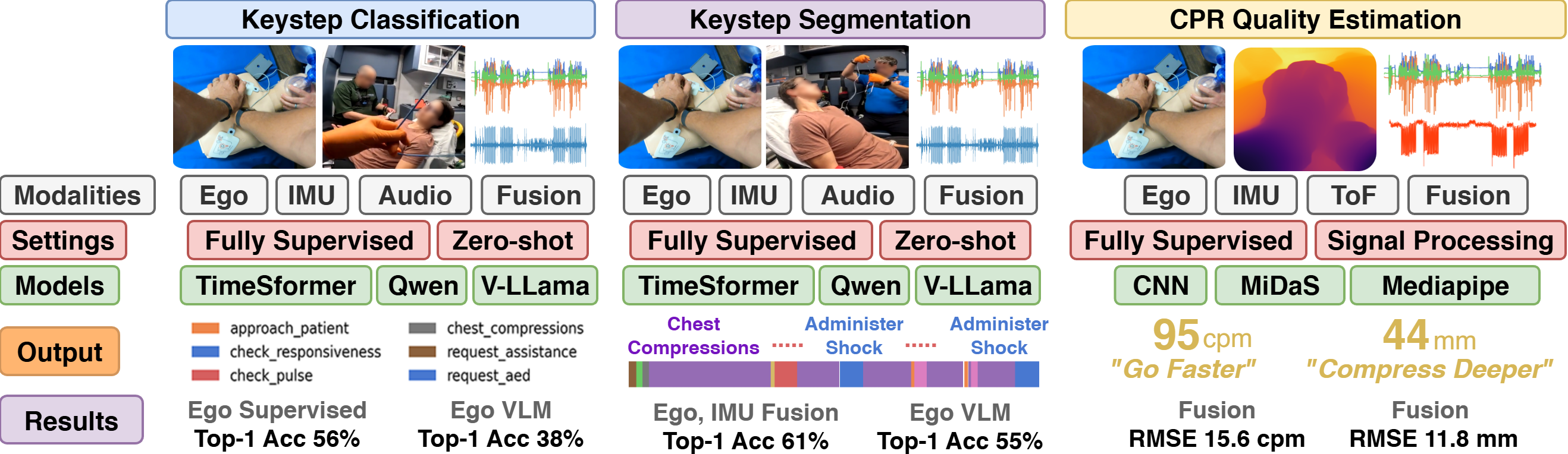}
    \caption{Overview of benchmark tasks for keystep classification, segmentation, and CPR quality estimation using EgoEMS data. Example baseline models and representative results highlight the diverse evaluation scenarios enabled by the dataset.
}
    \label{fig:benchmarks}
\end{figure*}

% ========== START OF BENCHMARKS ============== %

\section{EgoEMS Benchmarks}
\label{sec:benchmarks}

We present three benchmark tasks designed to evaluate the core capabilities of an ICA for EMS: (1) \textit{Keystep Classification} and (2) \textit{Keystep Segmentation}, which together form the broader task of keystep recognition, a foundational capability for guiding responders through complex protocols while monitoring their actions and (3) \textit{Action Quality Estimation}, which enables continuous feedback to improve execution quality. As illustrated in Figure~\ref{fig:benchmarks}, these tasks leverage synchronized multimodal data including egocentric video, audio, and smartwatch IMU to support \textit{real-time inference} of procedural context and responder performance. 

Below, we provide an overview of each benchmark task and results from a representative set of SOTA models. We select strong, representative baselines that cover complementary approaches, including transformer architectures \cite{gberta_2021_ICML,weerasinghe2024multimodal} capable of modeling long temporal dependencies and multimodal fusion, a convolutional TSN model \cite{wang2016temporal} as a widely used CNN baseline, few-shot cross-domain models such as MM-CDFSL \cite{Hatano2024MMCDFSL} and zero-shot vision–language models such as Qwen-2.5 \cite{bai2025qwen2} and VideoLLaMA-3.3 \cite{zhang2025videollama} to assess the potential of large pretrained models. For audio-only recognition, we also include a custom zero-shot pipeline using WhisperTimestamped \cite{lintoai2023whispertimestamped} and GPT-4o \cite{achiam2023gpt}. More detailed results and discussions are in Appendix D.

\subsection {Benchmark 1: Keystep Classification}
\noindent\textbf{Motivation.}  
Real-time keystep classification is a core capability for ICAs to guide responders through complex EMS protocols and monitor procedural adherence in real time. The intricacies of EMS interventions including rapid interactions with medical tools and parallel actions by multiple responders pose significant challenges for fine-grained action recognition. Egocentric video may suffer from occlusions or a limited field-of-view, while audio cues such as verbal requests for equipment can provide complementary information. Thus, drawing on insights from prior work~\cite{yadav2021review,sun_multimodal_activity_review}, leveraging multimodal data is essential for accurate activity classification.

\noindent\textbf{Problem setting.} 
We frame this as a multimodal action classification problem that aims to associate a data segment \( D_{seg} \) with a specific keystep in the set of keysteps in our EMS taxonomy. The trimmed data segments of a single modality or synchronized segments of multiple modalities, along with their associated keystep labels, are used during both training and testing. 
Given the scarcity of multimodal data in this domain, we also evaluate zero-shot methods, including LLMs, as baselines.

\noindent\textbf{Summary results.}
We observed the highest top-1 accuracy of 62.3\% using a supervised transformer model \cite{weerasinghe2024multimodal} with egocentric video features extracted from a ResNet50 backbone~\cite{he2016deep}, closely followed by 62.2\% when smartwatch IMU and video data were fused together. While the fusion of these complementary modalities was expected to provide an improvement, the early fusion strategy we used did not yield additional gains, suggesting that more advanced fusion methods are needed to fully leverage long-range temporal and modality-specific cues. Notably, a zero-shot Qwen-2.5 \cite{bai2025qwen2} model achieves 38.3\%, highlighting the potential of recent LLMs for activity recognition.
See Appendix D.1 for detailed results and additional baselines.

\subsection{Benchmark 2: Keystep Segmentation}
\noindent\textbf{Motivation.} 
While classification assigns keysteps to fixed segments, ICA systems must also operate in online settings where actions unfold continuously. Keystep segmentation enables real-time tracking of procedural progress and timely intervention with a limited amount of context. However, achieving fine-grained segmentation is particularly challenging due to frequent view changes, variable execution speeds, and limited data per window. Multimodal sensing is essential for reliable performance in these dynamic settings.

\noindent\textbf{Problem setting.}  
We approach this as an online action segmentation task, aiming to identify and track specific keysteps performed by the primary responder throughout an EMS trial. Each trial is divided into 5-second streaming data segments, 
and each frame or sample within these windows is analyzed to classify the keystep occurring at that moment. 
Similar to keystep classification, zero-shot methods, including LLMs, are used as baselines.

\noindent\textbf{Summary results.}
We achieved the best segmentation accuracy of 61\% using a supervised transformer model \cite{weerasinghe2024multimodal} 
with the fusion of egocentric video features and smartwatch IMU data.
Unlike classification, this multimodal fusion provides a notable 6\% improvement, likely attributable to the short temporal windows where
complementary modalities help disambiguate subtle, fine-grained activities by better leveraging temporal cues.
Interestingly, the zero-shot Qwen-2.5 \cite{bai2025qwen2} model achieved 55.5\% accuracy, highlighting the potential of modern LLMs. 
WhisperTimestamped combined with GPT‑4o \cite{achiam2023gpt,lintoai2023whispertimestamped}, using only audio, resulted in a lower accuracy of 38\%, likely due to responders not consistently verbalizing their actions during critical interventions. See Appendix D.2 for a more detailed analysis.

\subsection{Benchmark 3: CPR Quality Estimation}
\noindent\textbf{Motivation.}
%Beyond action recognition, 
An effective EMS ICA must be able to assess intervention quality to provide timely feedback. In CPR, compression rate and depth are critical quality metrics. The American Red Cross recommends 100--120 compressions per minute and a depth of at least 50~mm for adults ~\cite{AmericanRedCross2022Steps}. Deviations from these guidelines can compromise patient safety and reduce resuscitation success.
We address this task by estimating compression rate and depth using egocentric video and smartwatch IMU data. The egocentric view captures close-up procedural context, while the repetitive hand motions in chest compressions make wrist IMU signals well-suited for dynamic estimation. By combining visual and inertial cues, an ICA can robustly infer CPR quality metrics in real time.
We also propose a rule-based feedback generation framework that produces continuous, actionable insights to guide responder performance.

\noindent{\textbf{Problem setting.}} 
We formulate this task as an online recognition problem, where the model processes a 5-second sliding window of data,
from either the chest-mounted GoPro or smartwatch IMU to estimate CPR quality metrics in real time. For each window, the model outputs the compression rate \( r \) (compressions per minute) and compression depth \( d \) (mm). The depth \( d \) is first computed for each individual compression within the window and then averaged to generate stable and actionable feedback. Ground-truth values from the ToF sensor are used for model supervision. Additionally, the rule-based feedback framework provides continuous feedback per window, which is also used to evaluate the model's performance (see Appendix D.3).

\noindent\textbf{Summary results.}
Fusion of video and smartwatch IMU achieves the best overall CPR feedback performance, with an F1 score of 0.52 for compression rate and 0.83 for compression depth. While smartwatch IMU alone yields the lowest RMSE for compression rate, outperforming video and fusion, multimodal fusion provides the lowest RMSE for compression depth. See Appendix D.3 for more details.

% ========== END OF BENCHMARKS ============== %

% ========== START OF CONCLUSION ============== %

\section{Conclusion}
\label{sec:conclusion}

EgoEMS is the first end-to-end high-fidelity egocentric, multimodal, multiperson dataset, created to drive the development of AI systems that can support EMS first responders in the field. 
Developed in close collaboration with EMS professionals and aligned with national standards, it provides over 20 hours of synchronized multimodal data from IRB-approved simulated emergencies with real responders, including 9 interventions and 67 keysteps across 233 trials performed by 62 participants from multiple EMS agencies and the general public.
EgoEMS's low-cost and easily replicable data collection system, comprehensive taxonomy, fine-grained annotations, and open-source annotation and de-identification tools establish a foundation for developing and expanding realistic and high-fidelity medical datasets and multimodal cognitive assistants.  

% ========== END OF CONCLUSION ============== %

% ========== START OF ACKNOWLEDGEMENTS ============== %

\section*{Acknowledgments}
% The acknowledgments section, if included, appears right before the references and is headed ``Acknowledgments". It must not be numbered even if other sections are (use \texttt{\textbackslash section*\{Acknowledgements\}} in \LaTeX{}). This section includes acknowledgments of help from associates and colleagues, credits to sponsoring agencies, financial support, and permission to publish. Please acknowledge other contributors, grant support, and so forth, in this section. Do not put acknowledgments in a footnote on the first page. If your grant agency requires acknowledgment of the grant on page 1, limit the footnote to the required statement, and put the remaining acknowledgments at the back. Please try to limit acknowledgments to no more than three sentences.
% This work was supported in part by the award 70NANB21H029 from the U.S. Department of Commerce, National Institute of Standards and Technology (NIST) and a research grant from the Commonwealth Cyber Initiative (CCI). We are grateful for the support and participation of several volunteer EMS first responders at rescue squads and fire agencies in the Charlottesville and Virginia Beach areas, specifically \textit{North Garden Volunteer Fire Company}, \textit{Western Albemarle Rescue Squad}, \textit{Charlottesville Albemarle Rescue Squad} and \textit{Ocean Park Volunteer Rescue Squad}, as well as George Stephens, James Fitz-Gerald, Margaret Sande, Jon Howard, and David Keeler, who helped us with organizing the simulation experiments. 
This work was supported in part by the award 70NANB21H029 from the U.S. Department of Commerce, National Institute of Standards and Technology (NIST), and a research grant from the Commonwealth Cyber Initiative (CCI). We are grateful for the support and participation of several volunteer EMS first responders at rescue squads and fire agencies in the Charlottesville and Virginia Beach areas, specifically \textit{North Garden Volunteer Fire Company}, \textit{Western Albemarle Rescue Squad}, \textit{Charlottesville Albemarle Rescue Squad} and \textit{Ocean Park Volunteer Rescue Squad}.

% ========== END OF ACKNOWLEDGEMENTS ============== %

% ========== START OF ETHICAL STATEMENT ============== %

\section*{Ethical Statement}

This research was reviewed and approved by the Institutional Review Board for the Social and Behavioral Sciences (IRB-SBS) at the University of Virginia, which established protocols for participant recruitment, informed consent, experimental procedures, and data management. All participants were fully informed of the study’s purpose, procedures, potential risks and benefits, and future data usage through consent and materials release forms. Only data from participants who provided explicit consent for de-identified publication are included in the dataset. In accordance with the IRB protocol, all identifying information was removed prior to release. Each participant was assigned a unique identifier, faces were blurred semi-automatically in egocentric video, and sensitive textual or audio identifiers such as names, license plates, and ID cards were manually obscured. All data underwent manual verification to ensure privacy preservation and compliance with ethical standards.

EgoEMS was developed using simulated emergency scenarios with trained responders under IRB oversight, enabling high-fidelity modeling of real-world conditions while maintaining participant safety and privacy. However, the extension of such systems to real-world EMS introduces substantial ethical and privacy challenges. Obtaining informed consent during emergencies is often infeasible when patients are unconscious or in critical conditions, and incidentally captured bystanders or minors may not have provided consent. Real scenes also encompass sensitive situations such as domestic violence, substance use, or mental health crises, which require enhanced safeguards, institutional oversight, and alignment with frameworks like HIPAA. Furthermore, ensuring robust de-identification and responsible data governance in unconstrained environments remains an open technical and ethical challenge.

Beyond data collection, the deployment of ICAs in EMS contexts raises broader societal considerations regarding surveillance, bias, and trust. Models trained on simulated data may underperform in the field due to shifts in environmental and behavioral distributions, necessitating rigorous validation, transparency, and accountability measures before clinical use. Our work therefore positions EgoEMS as a privacy conscious proof-of-concept to advance ICAs responsibly, by offering an open-source data collection framework, EMS-specific ontology, annotation tools, and de-identification pipeline to support gradual, ethically governed expansion to real-world settings.

We advocate for future efforts to establish consent and governance frameworks in collaboration with IRBs, EMS agencies, and legal experts; to develop policy-level standards inspired by body-worn camera protocols; and to implement on-device, HIPAA-compliant data processing with built-in de-identification. At the algorithmic level, we emphasize the importance of multimodal learning methods that prioritize privacy-preserving modalities (e.g., IMU or depth sensors) when visual data is restricted. We also propose using transfer learning methods and exploiting multimodal redundancy to reduce the sim-to-real gap and improve models robustness and generalizability. Together, these principles aim to ensure that cognitive assistance technologies for emergency response evolve safely, fairly, and in service of public trust and societal benefit.

% ========== END OF ETHICAL STATEMENT ============== %

\clearpage

% ========== START OF SUPPLEMENTARY ============== %

% right before the section that should be the first entry in the partial TOC
\startcontents[after]           % begin collecting TOC entries into a bucket named "after"
\setcounter{tocdepth}{2}        % include up to subsections (adjust if needed)

% \title{EgoEMS: A High-Fidelity Multimodal Egocentric Dataset for \\ Cognitive Assistance in Emergency Medical Services \\ \Large Technical Appendix}
\clearpage
\appendixformat

\begin{center}
    \LARGE \textbf{Technical Appendix} \\[1em]
    \large EgoEMS: A High-Fidelity Multimodal Egocentric Dataset for Cognitive Assistance in EMS
\end{center}
\vspace{1em}

\printcontents[after]{}{1}{\section*{Table of Contents}}

\section{Taxonomy \& Annotations}
\label{sec:taxonomy_additional}
This section describes the methodology for creating the EMS taxonomy, defining protocols and interventions of interest, and associating them with relevant keysteps based on EMS guidelines and expert input. Additionally, it provides a detailed explanation of the annotation process.

\subsection{NEMSIS}
\begin{figure}[h!]
    \centering
    \includegraphics[width=0.6\linewidth]{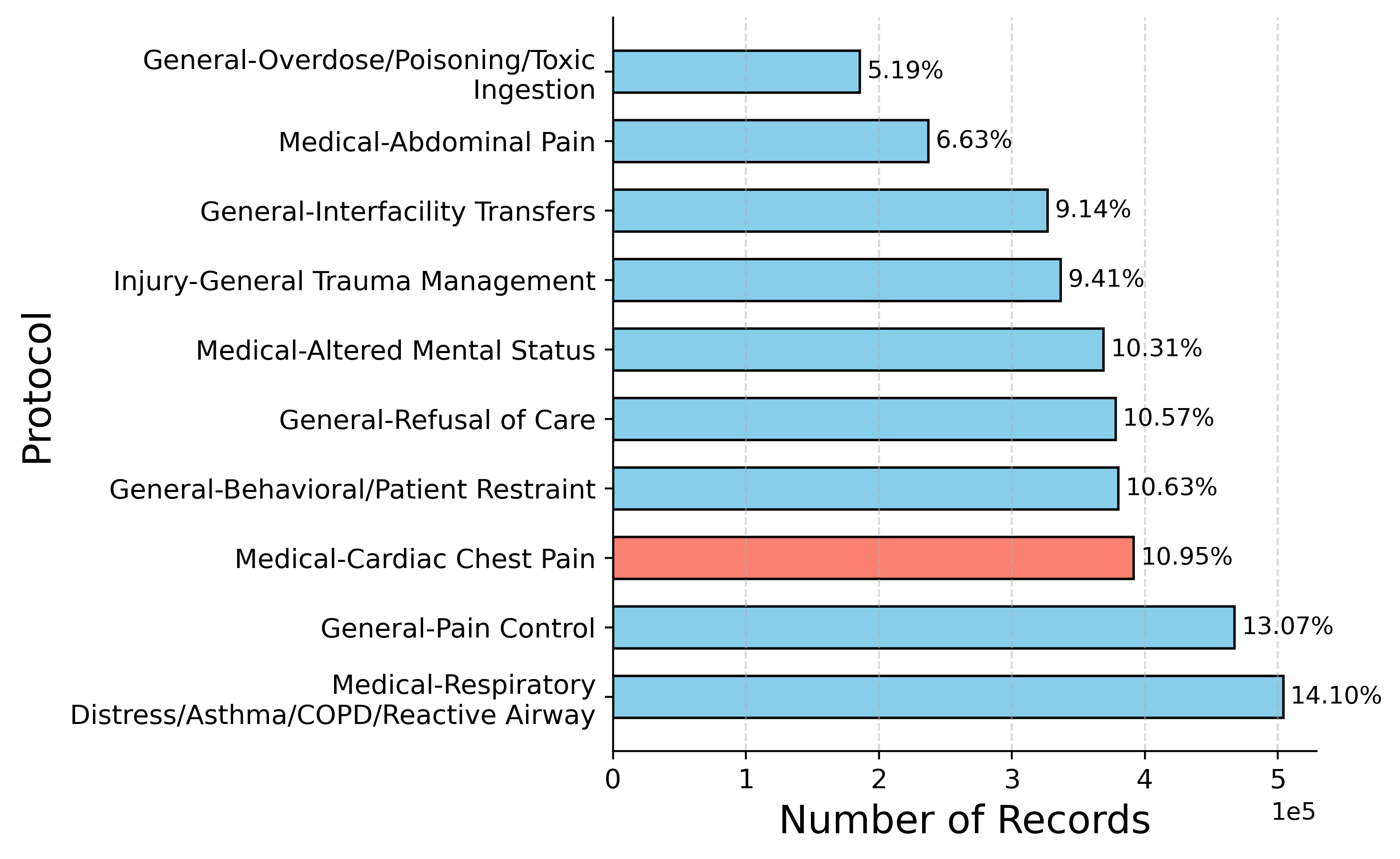}
    \caption{Most frequently executed EMS protocols in 2021.}
    \label{fig:protocols}
\end{figure}

We examined the 2021 dataset
from the National EMS Information System (NEMSIS)
consisting of 15,221,143 records from EMS agencies across the U.S., detailing emergency types, protocols followed, and procedures performed. We first analyzed the overall dataset’s distribution 
to identify the most frequently executed protocols and interventions at the national level. As shown in Figure \ref{fig:protocols}, we observe that protocols related to  cardiac emergencies (``Medical-Cardiac Chest Pain") rank among the top three with over 391,000 executions. Further, we included stroke emergencies due to the time-sensitive nature of the emergency and prompt management is critical for patient survival with minimum neurological damage \cite{saver2006time, odemsa}.
In consultation with EMS experts, we identified and selected 9 critical interventions from EMS protocols for inclusion in our dataset:
Airway-Breathing-Circulation (ABCs), Patient History, Vital Signs Assessment, Stroke Assessment, 12-lead Electrocardiogram (ECG), CPR, Ventilation, Defibrillation, and Transport (see Table~\ref{tab:interventions}).

\begin{table}[!h]
\centering
\small  % or \footnotesize or \scriptsize
\setlength{\tabcolsep}{1mm}
\begin{tabular}{@{}ll@{}}
\toprule
\textbf{Protocol} & \textbf{Interventions} \\ \midrule
\multirow{3}{*}{Cardiac Arrest} & ABCs \\
 & CPR \\
 & Ventilation \\
 & Defibrillation \\
 & Transport \\ \midrule
\multirow{5}{*}{Cardiac Suspected} & ABCs \\
 & Patient History \\
 & Vital Sign Assessment \\
 & 12-Lead ECG \\
 & Transport \\ \midrule
\multirow{6}{*}{Stroke} & ABCs \\
 & Patient History \\
 & Vital Sign Assessment \\
 & Stroke Assessment \\
 & 12-Lead ECG \\
 & Transport \\ \bottomrule
\end{tabular}
\vspace{1em}
\caption{Interventions associated with the protocols for basic life support.}
\label{tab:interventions}
\end{table}

% ============ tables ================ %
\begin{table*}[!t]
\centering
\small
\setlength{\tabcolsep}{1mm}
\begin{tabular}{@{}lll@{}}
\toprule
\textbf{Intervention} & \textbf{Keystep} & \textbf{Description} \\
\midrule

\multirow{4}{*}{ABCs}
 & approach\_patient & Approach patient and assess scene safety. \\
 & check\_responsiveness & Check for response to verbal or physical stimuli. \\
 & check\_pulse & Assess for carotid or radial pulse. \\
 & check\_breathing & Look, listen, and feel for breathing. \\
\midrule

\multirow{9}{*}{Stroke Assessment}
 & check\_grip\_strength & Check equal grip strength bilaterally. \\
 & face\_droop\_check & Ask patient to smile to check symmetry. \\
 & arm\_drift\_check & Ask patient to raise arms and observe drift. \\
 & speech\_abnormality\_check & Have patient repeat phrase to assess speech. \\
 & assess\_balance\_and\_coordination & Evaluate ability to balance and move. \\
 & document\_lkw\_time & Note last known well time. \\
 & check\_vision\_deficits & Ask about vision changes or loss. \\
 & evaluate\_aphasia & Assess understanding and speech. \\
 & assess\_neglect\_signs & Check for inattention to one side. \\
\midrule

\multirow{4}{*}{Patient History}
 & review\_medications & Review current medications. \\
 & inquire\_medication\_anticoagulants & Ask about anticoagulant use. \\
 & inquire\_hpi\_and\_pmh & Ask about illness and past history. \\
 & inquire\_substance\_use & Ask about alcohol or drug use. \\
\midrule

\multirow{9}{*}{{Vital Sign Assessment}}
 & check\_blood\_pressure & Measure blood pressure. \\
 & check\_heart\_rate & Check pulse rate. \\
 & check\_oxygen\_saturation & Measure oxygen saturation. \\
 & check\_respiratory\_rate & Count breaths per minute. \\
 & prepare\_glucometer\_and\_strip & Prepare glucometer and strip. \\
 & read\_and\_record\_glucose\_level & Record blood glucose reading. \\
 & check\_perrl & Check pupils for light reactivity. \\
 & check\_skin\_condition & Assess skin color, temp, moisture. \\
 & check\_a\&o & Check alertness and orientation. \\
\bottomrule
\end{tabular}
\caption{Keystep taxonomy (Part 1): Assessment and history-related interventions.}
\label{tab:taxonomy_assessment}
\end{table*}

% Table: Treatment interventions (Part A)
\begin{table*}[!t]
\centering
\small
\setlength{\tabcolsep}{1mm}
\begin{tabular}{@{}lll@{}}
\toprule
\textbf{Intervention} & \textbf{Keystep} & \textbf{Description} \\
\midrule

\multirow{5}{*}{CPR}
 & check\_responsiveness & Check for response to verbal or physical stimuli. \\
 & check\_pulse & Assess for carotid or radial pulse. \\
 & check\_breathing & Look, listen, and feel for breathing. \\
 & chest\_compressions & Perform compressions at correct rate/depth. \\
 & assess\_patient & Reassess patient condition. \\
\midrule

\multirow{7}{*}{Defibrillation}
 & request\_aed & Request AED early in resuscitation. \\
 & request\_assistance & Call for bystander or team assistance. \\
 & turn\_on\_aed & Power on AED and prepare for use. \\
 & attach\_defib\_pads & Attach pads to bare chest correctly. \\
 & clear\_for\_analysis & Ensure no contact during rhythm analysis. \\
 & clear\_for\_shock & Confirm everyone is clear before shock. \\
 & administer\_shock\_aed & Deliver shock if advised by AED. \\
\midrule

\multirow{5}{*}{Ventilation}
 & open\_airway & Use airway technique (e.g., head-tilt). \\
 & place\_bvm & Position and seal BVM mask. \\
 & ventilate\_patient & Deliver ventilations appropriately. \\
 & suction\_airway & Use suction to clear airway. \\
 & inset\_NPA & Insert NPA to maintain airway. \\
\bottomrule
\end{tabular}
\caption{Keystep taxonomy (Part 2B): Treatment-related interventions (CPR, defibrillation, ventilation).}
\label{tab:taxonomy_treatment_a}
\end{table*}

% Table: Treatment interventions (Part B)
\begin{table*}[!t]
\centering
\small
\setlength{\tabcolsep}{1mm}
\begin{tabular}{@{}lll@{}}
\toprule
\textbf{Intervention} & \textbf{Keystep} & \textbf{Description} \\
\midrule

\multirow{18}{*}{12-Lead ECG}
 & explain\_procedure & Explain ECG to patient. \\
 & shave\_patient & Shave skin for lead contact. \\
 & place\_left\_arm\_lead & Attach left arm electrode. \\
 & place\_right\_arm\_lead & Attach right arm electrode. \\
 & place\_left\_leg\_lead & Attach left leg electrode. \\
 & place\_right\_leg\_lead & Attach right leg electrode. \\
 & place\_v1\_lead & Place V1 at 4th ICS, right of sternum. \\
 & place\_v2\_lead & Place V2 at 4th ICS, left of sternum. \\
 & place\_v3\_lead & Place V3 between V2 and V4. \\
 & place\_v4\_lead & Place V4 at 5th ICS, midclavicular. \\
 & place\_v5\_lead & Place V5 at anterior axillary line. \\
 & place\_v6\_lead & Place V6 at midaxillary line. \\
 & request\_patient\_to\_not\_move & Ask patient to stay still during ECG. \\
 & turn\_on\_ecg & Power on ECG device. \\
 & connect\_leads\_to\_ecg & Connect leads to electrodes and ECG. \\
 & ask\_patient\_age\_sex & Confirm age and sex. \\
 & obtain\_ecg\_recording & Record ECG tracing. \\
 & interpret\_and\_report & Review and report ECG results. \\
\midrule

\multirow{6}{*}{Transport}
 & transport & Move patient to ambulance. \\
 & load\_patient\_to\_stretcher & Load onto stretcher. \\
 & secure\_patient\_on\_stretcher & Secure with straps. \\
 & notify\_hospital & Alert receiving hospital. \\
 & notify\_hospital\_of\_stroke\_alert & Activate stroke alert if needed. \\
 & handoff\_patient\_to\_hospital & Give report during handoff. \\
\midrule

Miscellaneous
 & no\_action & No specific action taken. \\
\bottomrule
\end{tabular}
\caption{Keystep taxonomy (Part 2A): Treatment-related interventions (12-lead ECG, transport, miscellaneous).}
\label{tab:taxonomy_treatment_b}
\end{table*}

\subsection{Keysteps}
Following selecting the protocols and interventions, we consulted the National Registry of Emergency Medical Technicians (NREMT) to review the keysteps and guidelines necessary for accurately executing each intervention. To ensure the robustness of our taxonomy, we collaborate closely with EMS experts, who validate the defined keysteps for each procedure. Specifically, we reference the Emergency Medical Responder Psychomotor Examination criteria \cite{nremt_psychomotor_exams}
% \homa{add citation} 
to inform the development of this keystep taxonomy, aligning it with established standards for practical skill assessment. Tables \ref{tab:taxonomy_assessment},\ref{tab:taxonomy_treatment_a},\ref{tab:taxonomy_treatment_b} present the taxonomy of EMS interventions and keysteps defined for our dataset.\\

\subsection{Scenarios}

\begin{figure}[!h]
    \centering
    \includegraphics[width=0.5\linewidth]{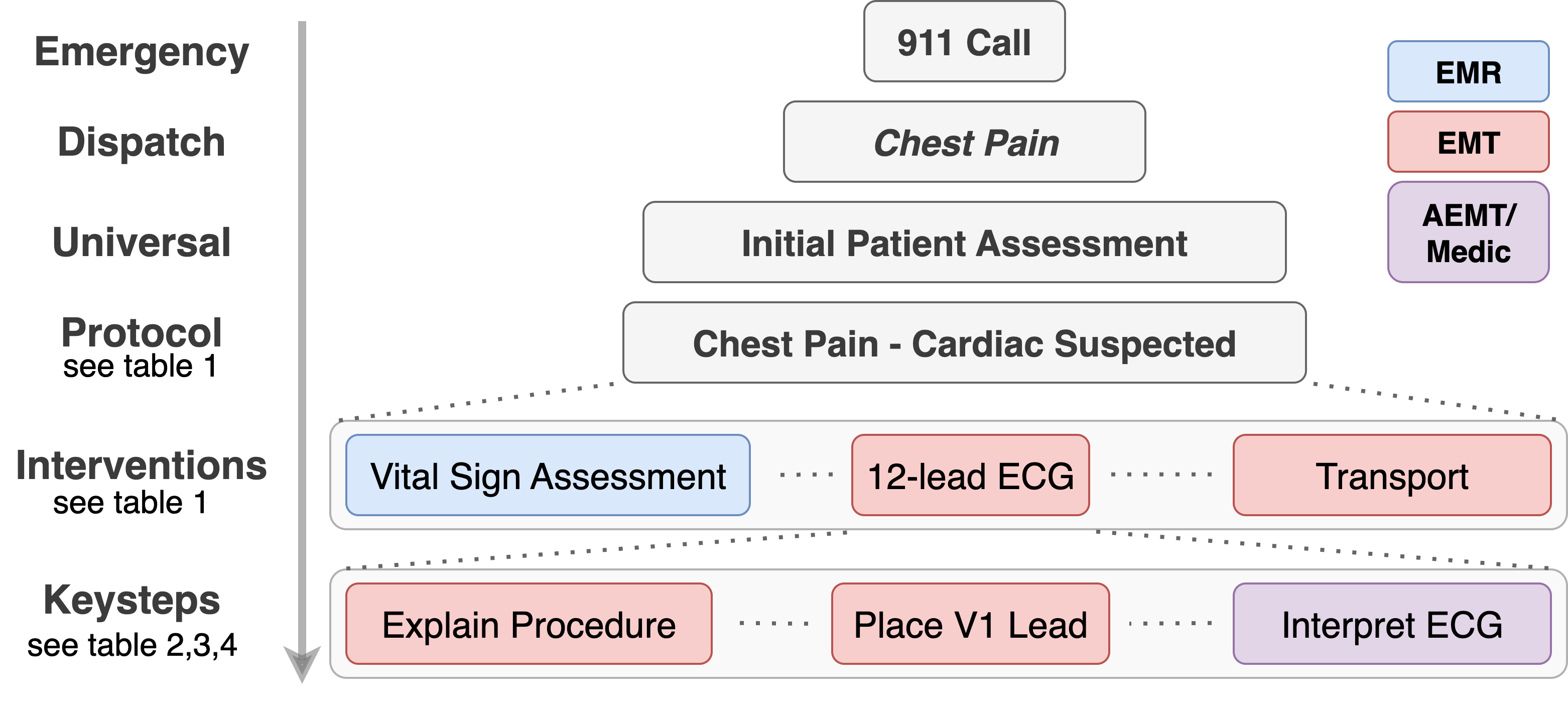}
    \caption{EMS emergency scenario flow of events.}
    \label{fig:ems-flow-hierarchy}
\end{figure}

In collaboration with EMS experts, we designed a diverse set of realistic simulation scenarios that reflect the urgency and complexity of actual emergency medical calls. These scenarios include a wide range of conditions such as cardiac arrest, suspected cardiac events, and stroke presentations. To increase clinical realism, the stroke scenarios also incorporate common stroke mimics such as hypoglycemia, as well as cases involving unresponsive patients, requiring responders to engage in differential diagnosis and protocol driven decision making. Figure \ref{fig:ems-flow-hierarchy} illustrates the hierarchical structure of a typical EMS response, using a chest pain emergency as an example. The workflow begins with a 911 call reporting symptoms (e.g., “my husband can’t breathe”), triggering dispatch to the nearest EMS agency. Upon arrival, responders initiate the \textit{Universal Patient Assessment}, a protocol driven sequence applicable across emergencies. Based on clinical observations, responders may reclassify the initial complaint and proceed with a tailored protocol such as \textit{Chest Pain – Cardiac Suspected} guided by the patient's evolving signs and symptoms.

This structure cascades into specific \textit{interventions} (e.g., Vital Sign Assessment, 12-lead ECG, Transport), each decomposed into granular \textit{keysteps} (e.g., explain procedure, place V1 lead, interpret ECG). To support model generalization and realism, we designed diverse scenarios featuring patients with varying comorbidities and sociocultural backgrounds. Scenario briefs were provided to participants before each simulation to standardize objectives while allowing natural variations and reactions in responder-patient interactions. For example, in the context of cardiac arrest emergency, we created multiple scenarios (see Figure~\ref{fig:example_scenario}) featuring patients with varied backgrounds and comorbidities, thereby enhancing the diversity and ecological validity of the dataset. 

% These scenario outlines were shared with participants prior to each simulation to guide the general flow of actions and conversations, while still allowing for responders to dynamically react to patient condition.

\begin{figure}[!h]
\centering
\begin{tcolorbox}[
  colback=blue!5!white,
  colframe=blue!75!black,
  title=Example Scenario: Cardiac Arrest,
  breakable=false
]
\textbf{En route}\\
\textbf{Dispatch:} Medic 427 is responding to a possible cardiac arrest involving an 86-year-old female found unconscious at home.\\

\textbf{On scene}\\
\textbf{EMT and Paramedic:} [Enter through the unlocked front door] Hello, we’re here with the rescue squad!\\
\textbf{Caretaker:} Please help me in here!\\

[The EMT and Paramedic enter the living room, finding the patient unconscious on the floor while the caretaker administers CPR.]\\

\textbf{EMT:} [Approaches the patient] Ma’am, can you move out of the way?\\
\textbf{EMT:} [Checks for pulse and breathing] No pulse, and she isn’t breathing. Starting CPR.\\
\textbf{Paramedic:} [Over the radio] Medic 427 confirms the code. The patient has no pulse and is not breathing. Starting CPR now.\\

\textbf{EMT 1:} [Begins chest compressions.]\\
\textbf{Paramedic 1:} [Attaches the AED to the patient.]\\

[Another medic unit and a fire truck arrive.]\\

\textbf{EMR 1:} [Begins ventilating between sets of compressions] I’ll get the BVM.\\
\textbf{Paramedic 1:} [Intubates the patient; Firefighter 1 takes over ventilation] I’m going to intubate the patient.\\
\textbf{Paramedic 2:} [Establishes IV access] I’ll try to start an IV.\\

[AED advises a shock and delivers it. The patient regains a pulse, remains unconscious, and breathes shallowly. Compressions are stopped, but ventilations continue.]\\

[The crew moves the patient onto the stretcher, into the ambulance, and transports to the hospital where care is transferred.]
\end{tcolorbox}
\caption{Example cardiac arrest simulation scenario.}
\label{fig:example_scenario}
\end{figure}

\begin{figure}[ht]
    \centering
    \includegraphics[width=0.40\linewidth]{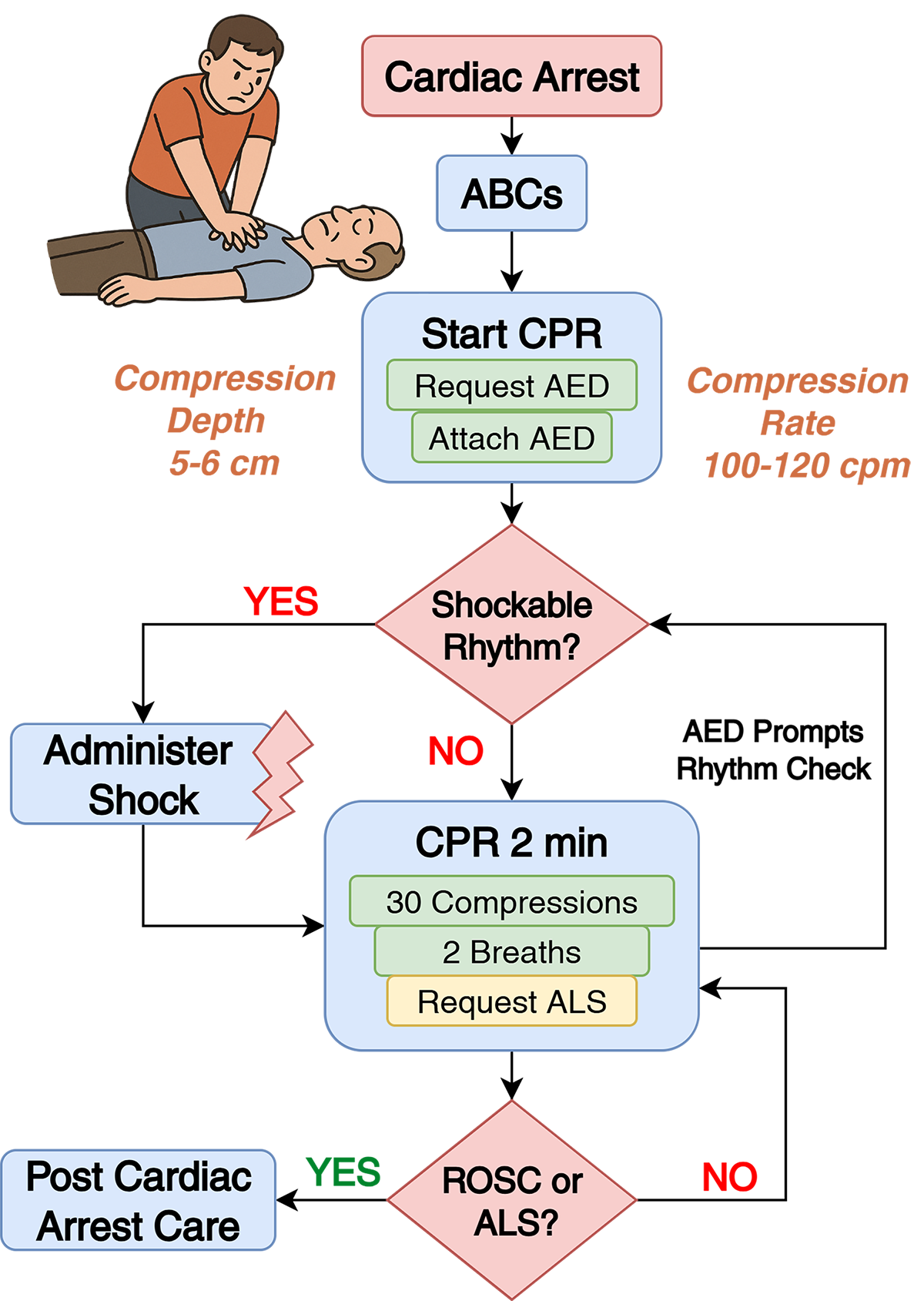}
    \caption{High-level overview of Cardiac Arrest protocol for BLS providers based on ODEMSA \cite{odemsa} and \cite{AmericanRedCross2022Steps} guidelines. ROSC: Return of Spontaneous Circulation, ALS: Advanced Life Support, AED: Automated External Defibrillator. 
}
    \label{fig:cpr_protocol_bls}
\end{figure}

Figure~\ref{fig:cpr_protocol_bls} presents a high-level overview of the Cardiac Arrest protocol for Basic Life Support (BLS) providers~\cite{odemsa, AHA2020_CPR_ECC_Algorithms}. Timely and accurate execution of this protocol is critical to improving patient survival outcomes. This also underscores the potential role of an ICA as a virtual partner that can alleviate cognitive burden by providing real-time feedback on CPR performance, timely reminders (e.g., delivering breaths after 30 compressions), and other protocol-driven cues.

We release these simulation scripts as part of our dataset to support future research on prehospital decision-making, multi-responder coordination, and cognitive assistance in dynamic emergency settings.

\subsection{Annotation}
We annotate the keysteps, conversational audio transcripts, and medical objects/tools in our dataset using manual and semi-automatic methods as described next.

\subsubsection{Keystep Annotations}
\label{keystep_annotation_appendix}

\begin{figure}[!h]
    \centering
    \includegraphics[width=0.7\linewidth]{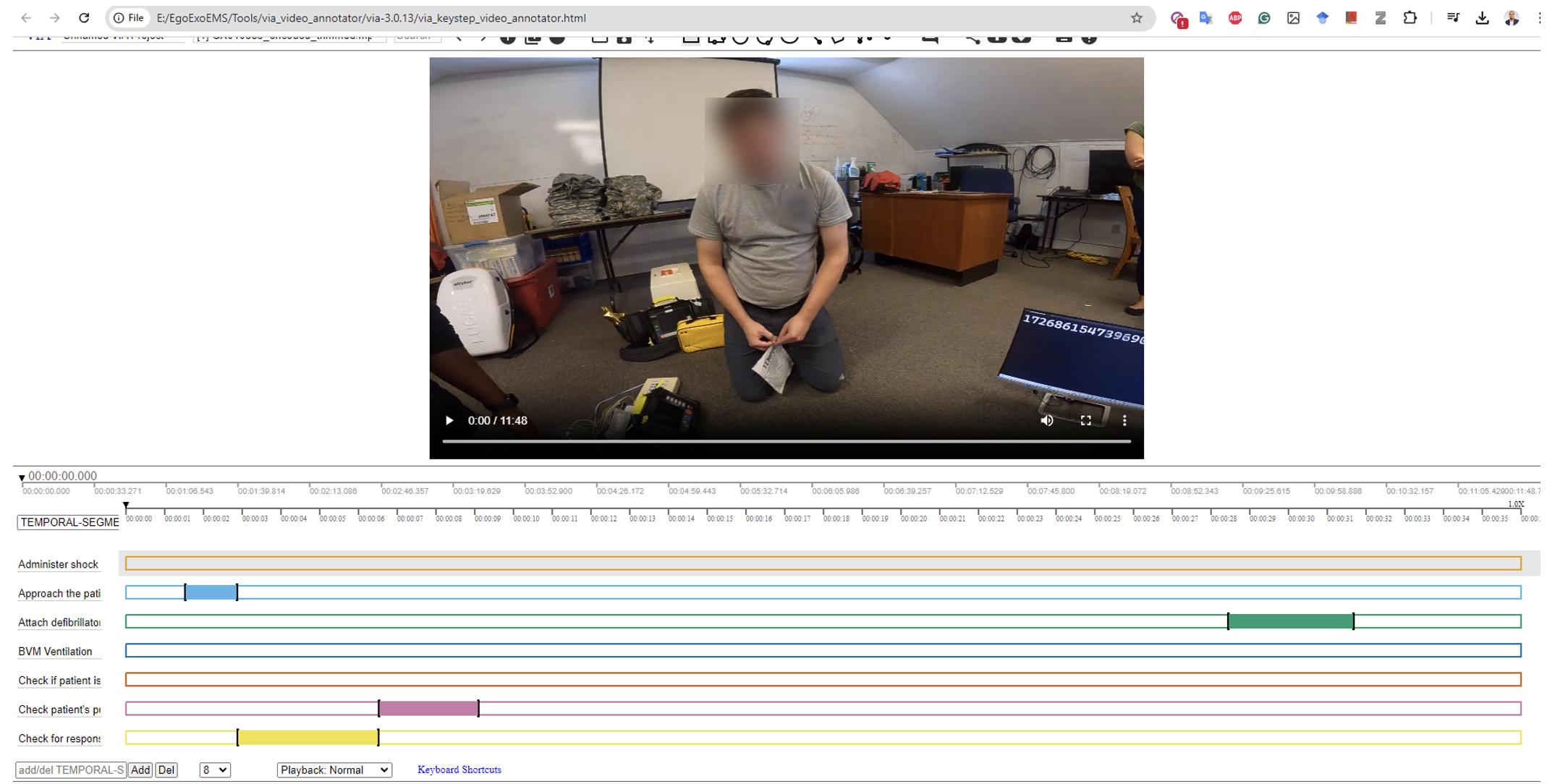}
    \caption{Video annotation tool from Oxford \cite{dutta2019vgg}.}
    \label{fig:video_annotation_tool}
\end{figure}

We utilized the publicly available Video Annotation Tool (VIA) from Oxford \cite{dutta2019vgg}, with custom modifications to manually annotate the sequence of keysteps (fine-grained actions) within each trial (see Figure \ref{fig:video_annotation_tool}). 

Human annotators initially completed all annotations and subsequently verified them with a different annotator to ensure accuracy. We consulted EMS experts for additional validation of complex keysteps, particularly those involved in the 12-lead ECG and Stroke Assessment interventions. 

\begin{figure}[!h]
    \centering
    \includegraphics[width=0.6\linewidth]{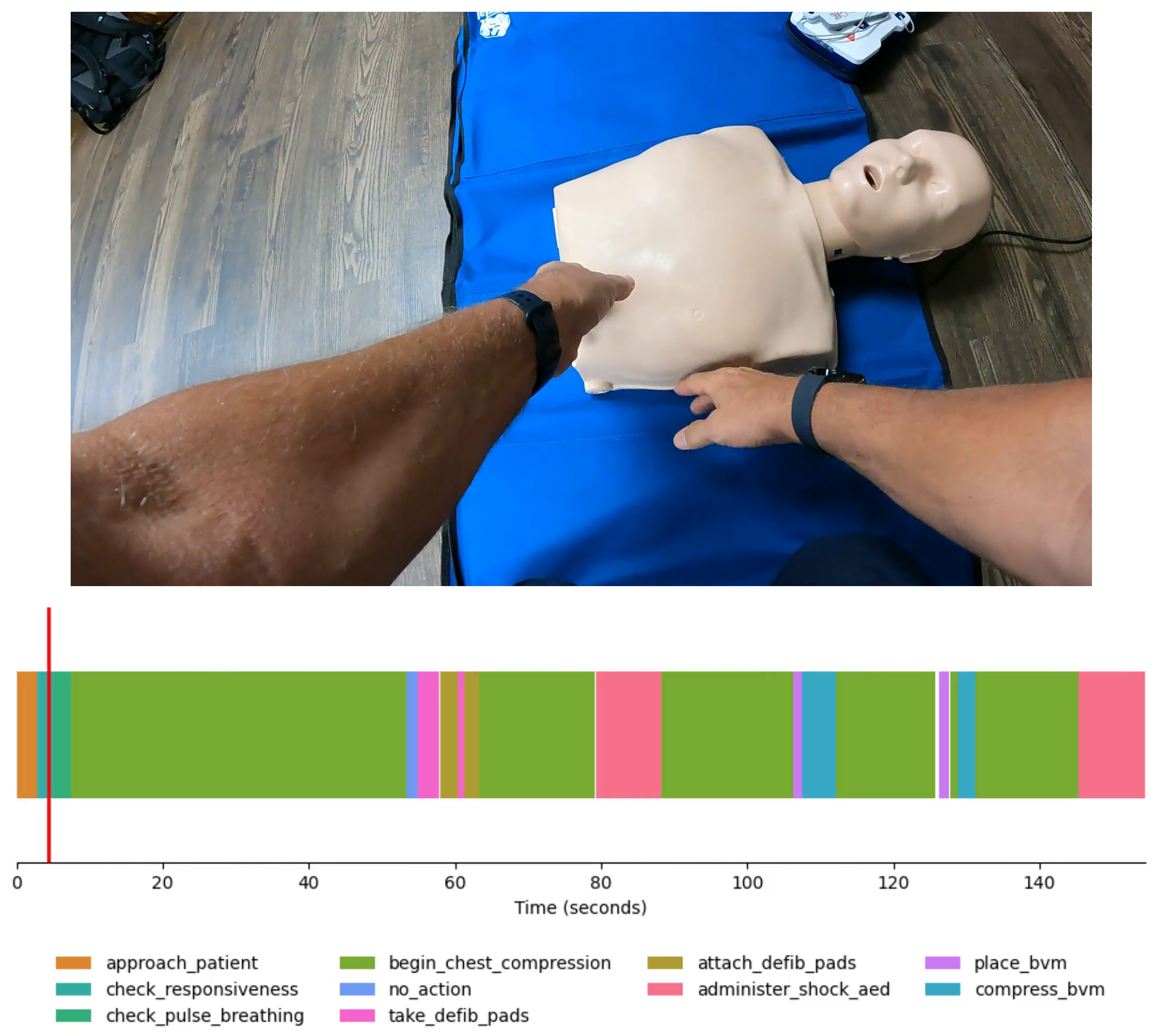}
    \caption{Visual verification tool for keystep annotation.}
    \label{fig:visual_verification_annotation}
\end{figure}

\begin{figure*}[!h]
    \centering
    \includegraphics[width=1\textwidth]{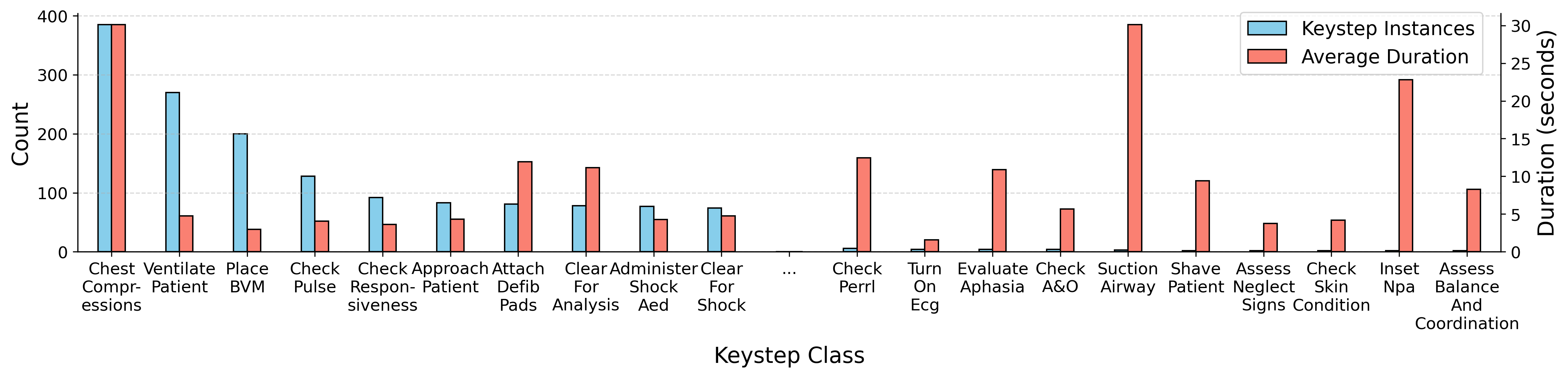}
    \caption{Keystep distribution showing the most frequent (head) and least frequent (tail) classes, with counts and average duration.
}
    \label{fig:head_tail_keystep_distribution}
\end{figure*}

\begin{figure*}[!h]
    \centering
    \includegraphics[width=\textwidth]{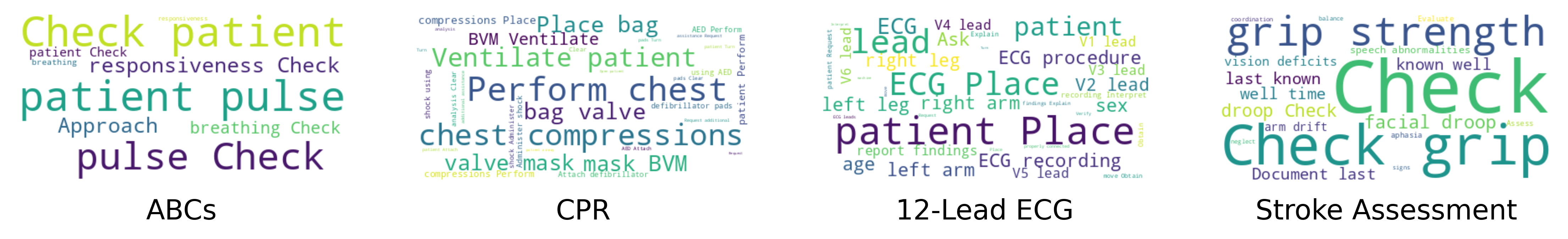}
    \caption{Keystep annotation distribution for certain interventions visualized through word clouds.}
    \label{fig:word_cloud}
\end{figure*}

We developed a tool (see Figure \ref{fig:visual_verification_annotation}) to visualize keysteps and verify annotations. %, enabling a secondary reviewer to verify the annotations. 
Figure~\ref{fig:head_tail_keystep_distribution} illustrates the head–tail distribution of keysteps in the dataset, while Figure~\ref{fig:word_cloud} presents word clouds of keystep annotations across few different interventions.
Finally, we transformed the annotations from the VIA tool into a custom annotation file structure tailored for usability with the dataset’s multiview, multimodal data, and subject-specific details, such as expertise level. 

\subsubsection{Audio Transcript Annotations}
\label{sec:automated_audio_transcript_annotation}
We employ a semi-automatic approach based on AI speech recognition and manual verification to generate transcripts of the EMS conversational audio recordings. %Gemini 1.5~\cite{reid2024gemini} to generate transcripts of the EMS audio recordings. 
We evaluated the performance of four state-of-the-art zero-shot speech recognition models (shown in Table~\ref{tab:speech_recognition_result}) and selected Gemini-2.5~\cite{reid2024gemini} for this task. Speech recognition accuracy of each model was evaluated against the manual transcriptions for a randomly selected 10 samples from the dataset using the word error rate (WER) and Mean Absolute Error (MAE) for timestamp.

Gemini-2.5 achieved the lowest WER among all SOTA models, with a WER of 0.31 and MAE for 0.18s/0.18s for word-level start/end timestamp, indicating high reliability and quality in its transcription output. The reason for superiority of Gemini-2.5 over others is that our prompt provides abundant EMS context and require Gemini-2.5 to filter out the background noise.
Figure~\ref{fig:prompt_for_ASR} shows the prompt template for Gemini-2.5 to do zero-shot speech recognition.

\begin{table}[!h]
\centering
\small
\setlength{\tabcolsep}{1mm}
\begin{tabular}{ccc c}
\toprule
\textbf{Model} & \textbf{WER} & \multicolumn{2}{c}{\textbf{MAE (s)}} \\
\cmidrule(lr){3-4}
& & $T_{\text{start}}$ & $T_{\text{end}}$ \\
\midrule
\shortstack{Google Speech-to-Text~\cite{zhang2023google}} & 0.59 & 5.06 & 5.79\\
\shortstack{Whisper-X~\cite{radford2023robust}} & 0.68 & 5.42 & 5.42\\
\shortstack{Whisper-Timestamped~\cite{lintoai2023whispertimestamped}} & 0.62 & 1.20 & 1.08\\
\shortstack{Gemini-2.5~\cite{reid2024gemini}} &  0.31 & 0.18 & 0.18\\
\bottomrule
\end{tabular}
\vspace{1em}
\caption{Speech recognition result of SOTAs.}
\label{tab:speech_recognition_result}
\end{table}

\begin{figure}[!h]
\centering
\begin{tcolorbox}[breakable=false]
\noindent\textbf{PROMPT:} Given a recording, transcribe it into a timestamped dialogue
between first responders (label them 1, 2, 3), a patient, and bystanders.
Keep only content relevant to the EMS scenario. \\

\noindent\textbf{Step 1:} Perform speech recognition and produce a
role‑tagged dialogue. Include AED machine utterances and discard irrelevant
noise. \\

\noindent\textbf{Step 2:} Correct role labels and remove remaining irrelevant lines. \\

\noindent\textbf{Step 3:} Assign precise start and end timestamps
(\texttt{MM:SS.MS}) for each word and each full utterance. \\

\noindent\textbf{Step 4:} Return the output in this JSON schema:
\begin{lstlisting}
{
"Role": "<role>",
"Utterance": "word_1 word_2 ..",
"Start": "MM:SS.MS",
"End": "MM:SS.MS",
"Words": 
 [
  {
    "Word": "word_1",
    "start": "MM:SS.MS",
    "end":   "MM:SS.MS"
  },
  {
    "Word": "word_2",
    "start": "MM:SS.MS",
    "end":   "MM:SS.MS"
  }
 ]
}
\end{lstlisting}
\end{tcolorbox}
\caption{Prompt template for timestamped EMS transcription.}
\label{fig:prompt_for_ASR}
\end{figure}

\subsubsection{Bounding Box and Segmentation Masks}

\begin{figure}[!h]
    \centering
    \includegraphics[width=0.6\linewidth]{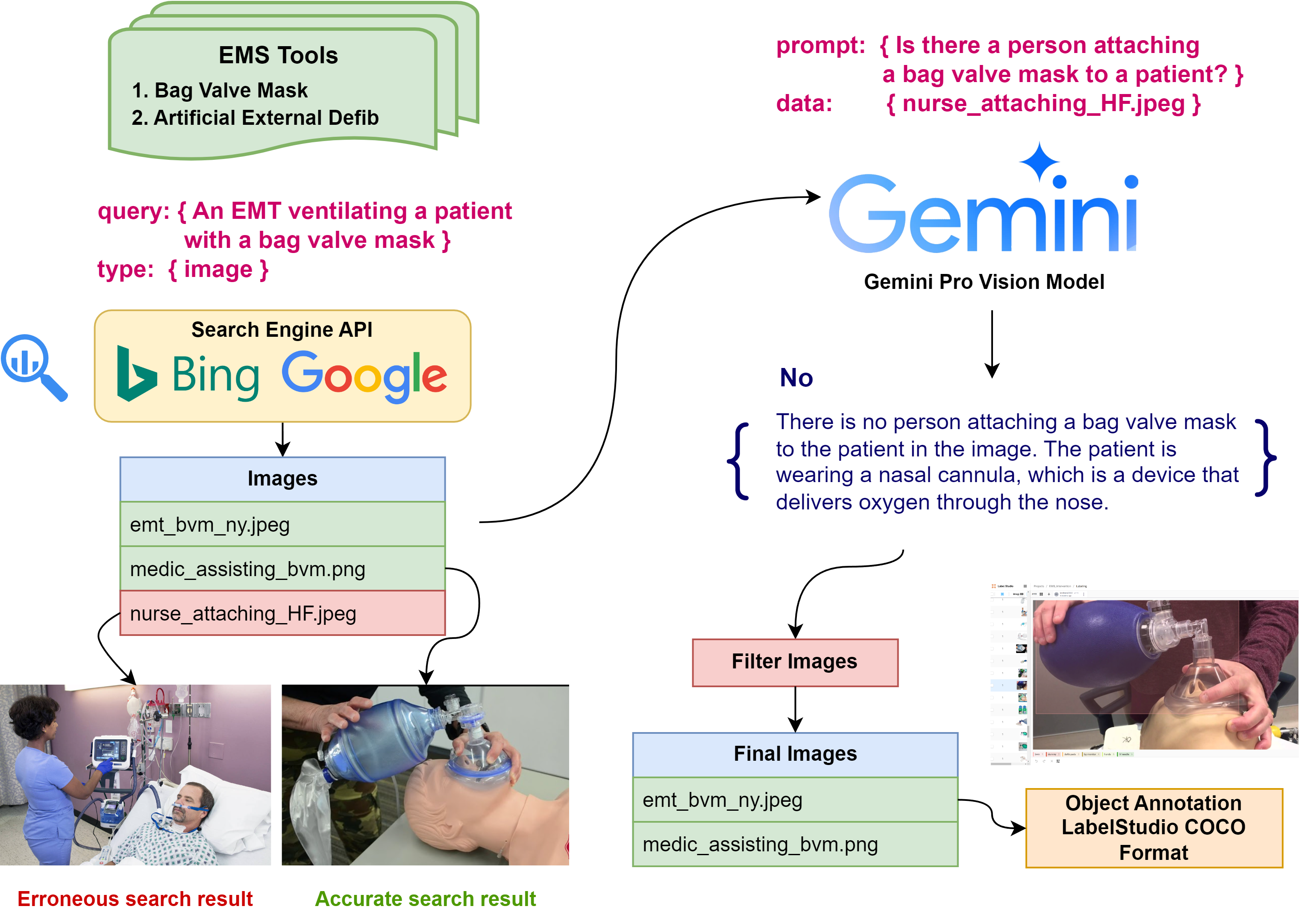}
    \caption{Automated object image dataset creation pipeline.}
    \label{fig:object_dataset}
\end{figure}

To provide bounding boxes and segmentation masks for EMS-related medical objects,
we developed a semi-supervised approach to generate bounding boxes and segmentation masks efficiently. This approach involves fine-tuning a state-of-the-art object detector, DETR \cite{carion2020end}, and an unsupervised segmentation mask generation model, SAM2 \cite{ravi2024sam2}.
Our method begins with a small, labeled dataset of EMS objects, which we term the “seed dataset” used to fine-tune the object detector. We created an automated pipeline to source this seed dataset by leveraging APIs from various search engines such as Google and Bing, followed by filtering with Gemini \cite{reid2024gemini} to remove inaccurate results, as illustrated in Figure \ref{fig:object_dataset}.

After annotating the seed dataset of EMS objects, we fine-tune the DETR model by replacing its classification head and training it for 30 epochs to predict bounding boxes for EMS objects of interest. We then apply Segment Anything 2 (SAM2) \cite{ravi2024sam2} to generate segmentation masks for these objects automatically. Due to SAM2’s design, it does not directly generate masks specific to the objects of interest. To address this, we implement a pipeline that leverages the bounding boxes predicted by DETR as regions of interest, enabling SAM2 to generate segmentation masks within those areas. Since the segmentation masks rely on bounding boxes generated by DETR, any error in the bounding boxes can propagate to the masks. We advise users to consider this potential error when utilizing these bounding boxes and segmentation masks in their applications.

\begin{figure}[!h]
    \centering
    \includegraphics[width=0.6\linewidth]{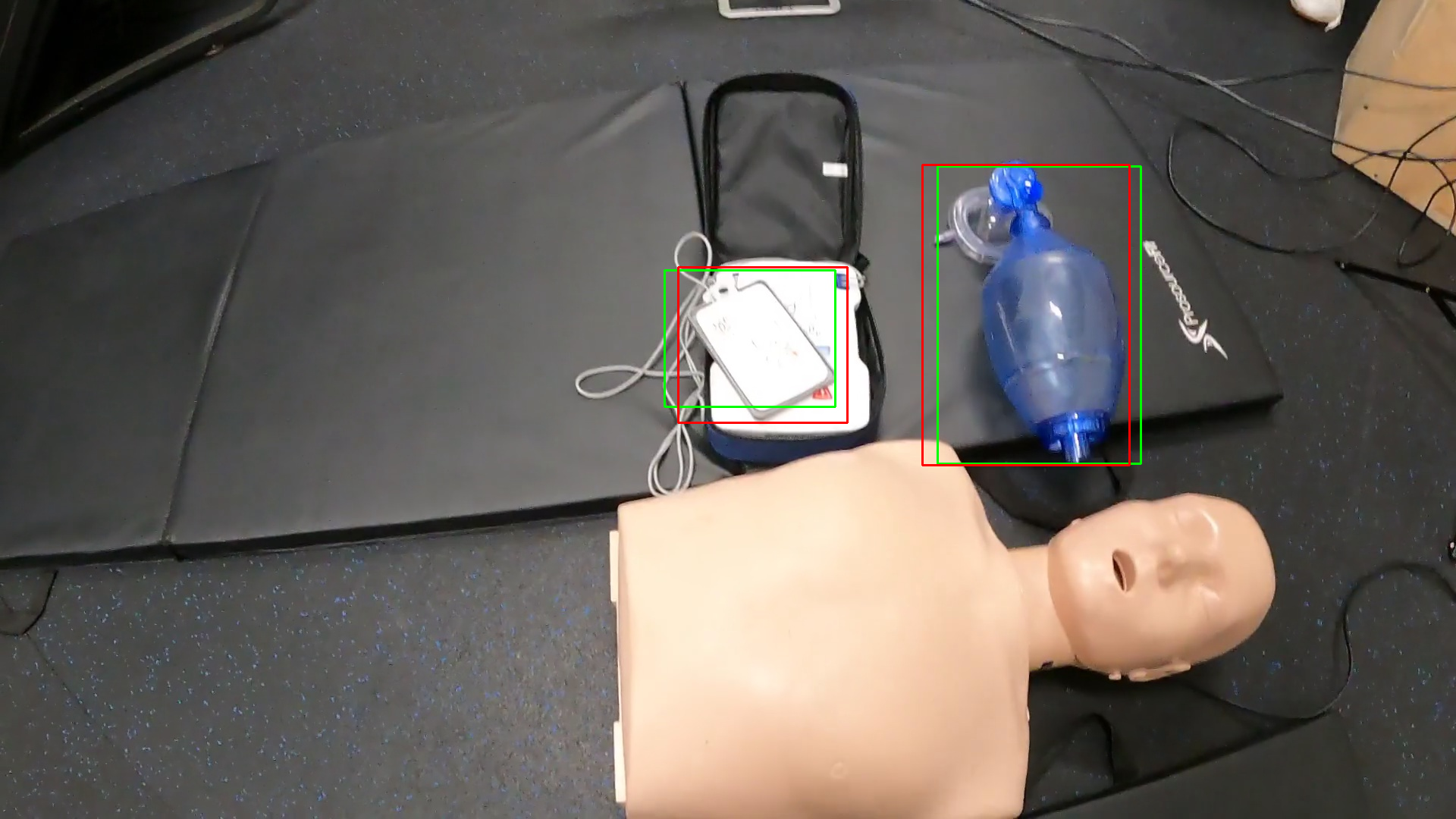}
    \caption{Comparison between human annotated bounding boxes and semi-supervised approach. Green boxes represent human annotation, and red boxes represent machine prediction.}
    \label{fig:iou_comparison}
\end{figure}

This semi-automatic approach significantly reduces the labeling cost while slightly affecting accuracy, as shown in Table \ref{table:bbox_comparison}. 
We compare the resource requirements for fully manual annotation of medical objects across the entire dataset with those of our semi-supervised approach. To get an estimate of the time needed for manual annotation and error introduced by the semi-automatic approach, we select a subset of 10\% randomly selected trials from the dataset and manually label the bounding boxes of EMS objects (see Figure \ref{fig:iou_comparison}). We then calculate the IOU, center, width, and height differences between manual and semi-supervised approaches for these trials as shown in Table \ref{table:bbox_comparison}.

\begin{table}[!h]
\centering
\small
\setlength{\tabcolsep}{1mm}
\begin{tabular}{@{}lcc@{}}
\toprule
\textbf{Metric} & \textbf{Fully manual} & \textbf{Semi-supervised} \\
\midrule
Time (hrs) & 66* & 1 \\
IoU (\(\uparrow\)) & \(\Diamond\) & 0.76 \\
Center difference \% (\(\downarrow\)) &  \(\Diamond\) & 11.67 \\
Width difference \% (\(\downarrow\)) &  \(\Diamond\) & 2.18 \\
Height difference \% (\(\downarrow\)) &  \(\Diamond\) & 5.24 \\
\bottomrule
\end{tabular}%
\vspace{1em}
\caption{
Comparison between fully manual and semi-supervised approach for bounding box annotation.
\(\Diamond\):  indicates metrics not applicable to the fully manual method, as it serves as the ground truth. *: Estimated based on the average time required per frame consisting of 3 objects.}
\label{table:bbox_comparison}
\end{table}

We observe that the bounding boxes generated using the semi-supervised approach achieve an average IoU of 0.76 when compared to human-labeled bounding boxes (validated on 10\% of the dataset). Additionally, the differences in bounding box dimensions are ~11\% for the center, ~2\% for the width, and ~5\% for the height. Most of the observed errors stem from false negatives, where certain objects were not detected. Despite these limitations, this approach offers a significant trade-off in terms of cost-efficiency,\textit{ achieving a 98.5\% reduction in labeling time} compared to fully manual annotation. 
These accuracy levels are typically sufficient for downstream tasks. For instance, large-scale benchmarks such as Open Images V4 \cite{kuznetsova2020open} treat automatically generated boxes with IoU $\geq$ 0.7 as high quality after human verification. Accordingly, our semi-automatic pipeline provides a practical, scalable annotation strategy for large datasets.

% \begin{figure*}[t!]
%     \centering
%     \includegraphics[width=\textwidth]{ArxivVersion/assets/wordclouds.png}
%     \caption{Keystep annotation distribution for certain interventions visualized through word clouds.}
%     \label{fig:word_cloud}
% \end{figure*}

\subsection{Dataloaders}
% \homa{Does this section belong here? Maybe better in Appendix A?}
The final annotation files follow the structure shown in Figure~\ref{fig:json_format}, providing a standardized, easy‑to‑use format that eliminates the need for complex custom parsing. To further improve the useability of this dataset, we also provide PyTorch implementations of a dataset class designed to load data across all modalities, either at the segment level or window level. This functionality is released as a Python package, enabling seamless installation and integration for machine learning development.

\begin{figure}[!h]
\centering
\begin{minipage}{0.7\linewidth}
\begin{tcolorbox}[
  colback=blue!5!white, 
  colframe=blue!75!black,
  title=JSON Structure,
  breakable=false
]
\begin{lstlisting}
{
"subjects": [
{
  "subject_id": "...",
  "expertise_level": "...",
  "scenarios":[
    {
    "scenario_id":"",
      "trials": [
        {
          "trial_id": "...",
          "scenario": "...",
          "streams": {
            "smartwatch_imu": { "..." },
            "vl6180_ToF_depth": { "..." },
            "egocam_rgb_audio": { "..." }
          },
          "keysteps": [
            {
              "keystep_id": "...",
              "start_t": "...",
              "end_t": "...",
              "label": "...",
              "class_id": "..."
            }
          ]
 .........
}
\end{lstlisting}
\end{tcolorbox}
\caption{Organized annotation structure improves the ease of use of the dataset.}
\label{fig:json_format}
\end{minipage}
\end{figure}

% ========================================================= %

\section{Participants, Privacy \& Ethics}
\label{sec:participants_additional}

This section provides an overview of participant recruitment, their backgrounds, contributions to the dataset, and the privacy-preserving measures implemented in compliance with the IRB guidelines.

\subsection{Participants}

We recruited a total of 62 participants. 46 participants were affiliated with EMS agencies where 42 were certified EMS professionals (EMR, EMT, or Paramedic) with experience ranging from a few years to over 30 years, and 4 were members of an EMS agency who did not hold formal EMS certifications. The remaining 16 participants were volunteers from the general public with no EMS background.
Certified participants at the EMT level or higher from ALS agencies performed complex procedures like the 12-lead ECG, Stroke Assessments while participants from the general public performed CPR (chest compressions only) due to their lack of EMS training. Basic instructions on CPR were provided to these volunteers before data collection. Table \ref{participant_background_stats} summarizes the dataset contributions from participants of various backgrounds.
Data was primarily collected from EMS agencies across [Anonymous], USA, and volunteers from a University, resulting in a total of 233 trials ($\sim$20 hours) and 2,694 keysteps of procedural recordings.\\

\begin{table}[!h]
\centering
\small
\begin{tabular}{@{}lcccc@{}}
\toprule
\textbf{Background} & \textbf{Subjects} & \textbf{CA} & \textbf{CS} & \textbf{ST} \\
\midrule
EMT, Paramedic & 43 & 62 (155) & 23 (173) & 41 (735) \\
EMS Member $\bigstar$     & 3  & 14 (28)  & \(\Diamond\) & \(\Diamond\) \\
General Public  & 16 & 93 (116) & \(\Diamond\) & \(\Diamond\) \\
\midrule
\textbf{Total}  & \textbf{62} & \textbf{169 (299)} & \textbf{23 (173)} & \textbf{41 (735)} \\
\bottomrule
\end{tabular}
\vspace{1em}
\caption{Dataset composition by participant background and scenario. CA: Cardiac Arrest; CS: Cardiac Suspected; ST: Stroke. Values represent trial counts, with total video duration in minutes shown in parentheses. \(\Diamond\): Not certified to perform the intervention. $\bigstar$: EMS members are not individually certified.}
\label{participant_background_stats}
\end{table}

% \subsection{IRB}

\subsection{Privacy \& Ethics}
\label{privacy_appendix}

This research study was reviewed and approved by the Institutional Review Board for the Social and Behavioral Sciences%(IRB-SBS)
, which established protocols for participant recruitment, informed consent, experiment procedures, and data management. All participants received a detailed informed consent agreement and materials release forms outlining the study’s purpose, procedures, risks, benefits, and future data usage. We retain all IRB documents and signed forms, available upon request for verification. 

Only data from participants who consented to de-identified publication is included in the dataset. In accordance with the guidelines outlined in the IRB package, we ensure the removal of all identifying information from our dataset prior to release. For de-identification, each subject is assigned a unique ID, and we eliminate identifying information such as names and places from the audio and transcripts and blur faces semi-automatically in egocentric video. Further, identifying information such as vehicle license plates from EMS ambulances and ID cards that were in the field-of-view of the camera were manually blurred. Finally, all data go through a manual verification and correction to ensure that privacy is preserved. 

% new discussion on real-world aspects
While EgoEMS was developed using simulated emergency scenarios with trained responders under IRB approval, transitioning toward real-world data collection and deployment of cognitive assistance systems introduces additional privacy and ethical challenges. In real emergencies, obtaining informed consent may be infeasible when patients are unconscious, in critical condition, or when bystanders and minors are incidentally recorded. These environments often include sensitive contexts such as domestic violence, substance use, or mental health crises that require protections and oversight beyond conventional IRB frameworks. Moreover, maintaining compliance with regulations like HIPAA and ensuring that de-identification tools remain effective in unconstrained, dynamic field environments are major open challenges.

EgoEMS therefore serves as a foundational, privacy conscious testbed for advancing ICAs in Emergency Medical Services. Beyond the dataset itself, our open-source data collection framework, EMS specific ontology, semi-automatic annotation tools, and de-identification pipeline provide the infrastructure for gradual and ethically governed expansion to real-world data. Future efforts should focus on developing consent and data governance frameworks in collaboration with IRBs, EMS agencies, and legal advisors; defining policy-level standards inspired by existing precedents such as body-worn camera protocols; and enabling on-device, HIPAA compliant data processing with built-in de-identification.
At the algorithmic level, advancing multimodal learning that prioritizes privacy preserving modalities such as IMU or depth data when video is restricted will be critical. Together, these efforts will ensure that cognitive assistance systems for EMS evolve responsibly from controlled simulation toward safe, privacy-respecting real-world deployment.

\subsubsection{Video De-identification}

\begin{figure}[!h]
    \centering
    \includegraphics[width=0.6\linewidth]{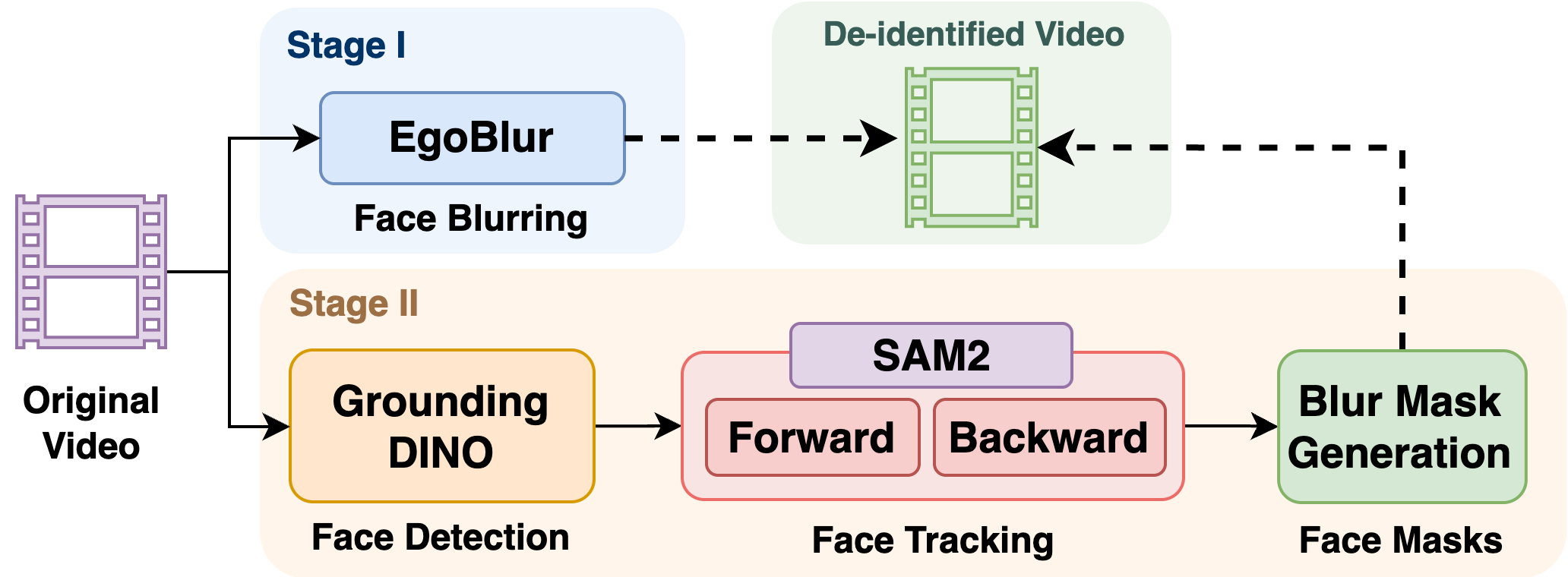}
    \caption{Automated video de-identification pipeline for maximum efficacy.}
    \label{fig:video-deidentification-pipeline}
\end{figure}

To preserve participant privacy in compliance with IRB requirements, we implement a two-stage video de-identification pipeline (see Fig.~\ref{fig:video-deidentification-pipeline}). In Stage I, we apply EgoBlur \cite{raina2023egoblur}, a model specifically developed for egocentric video de-identification, to detect and obscure identifiable regions such as faces and license plates. This stage performs fast, frame-wise blurring and serves as a strong baseline for privacy protection.

To address failure cases where EgoBlur may miss partially occluded or side-profile faces, we introduce Stage II, a more robust spatiotemporal de-identification method. We first perform text-guided face detection using Grounding DINO \cite{liu2023grounding}, which provides high-recall bounding boxes based on the textual prompt “face.” These detections are then passed to SAM2 \cite{ravi2024sam2}, which performs bidirectional tracking (forward and backward) across frames, producing temporally consistent masks.
From the tracked sequences, we generate pixel-accurate blur masks using SAM2 segmentation outputs. These masks are applied to the de-identified video from Stage I video, ensuring consistent, robust face blurring even under motion, occlusion, or camera shifts.

Finally, all videos undergo manual verification to correct any missed detections or false positives, ensuring strict adherence to ethical and legal data-sharing requirements. This hybrid pipeline allows us to achieve reliable and scalable visual de-identification for egocentric datasets.

\subsubsection{Audio and Transcript De-identification}

\begin{figure}[!h]
    \centering
    \includegraphics[width=0.6\linewidth]{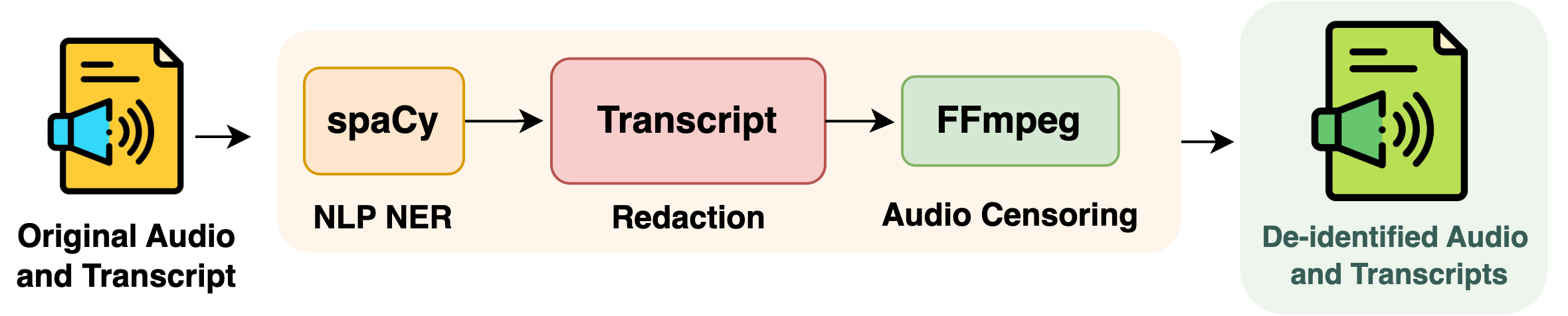}
    \caption{Automated audio de-identification pipeline for maximum efficacy.}
    \label{fig:audio-deidentification-pipeline}
\end{figure}

To remove personally identifiable information (PII) from speech data, we implement a natural language processing (NLP)-driven audio de-identification pipeline (Fig.~\ref{fig:audio-deidentification-pipeline}). The process begins with the timestamped transcript generated via automatic speech recognition, which is then processed using the spaCy NLP library \cite{spacy2020} to perform Named Entity Recognition (NER). This identifies sensitive entities such as names and location that may compromise subject privacy.

Detected entities are passed to the Redaction module, where they are masked in the transcript. The corresponding audio timestamps of these redacted words are then used to censor the original audio using FFmpeg \cite{ffmpeg}, applying beeps or silence overlays to ensure spoken identifiers are inaudible in the final version.

The result is a synchronized pair of de-identified audio and transcripts, which can be safely used for downstream analysis and machine learning model training while remaining compliant with institutional review board (IRB) requirements.

\subsubsection{Manual Verification and Correction}

\begin{figure}[!h]
    \centering
    \includegraphics[width=1\linewidth]{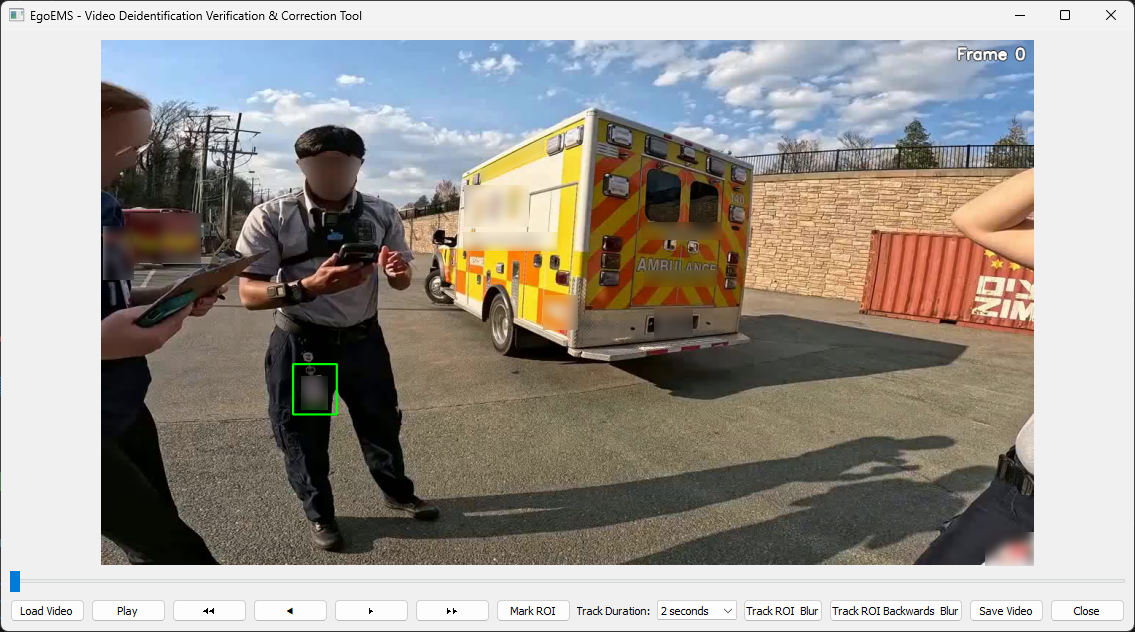}
    \caption{Interactive graphical application for manual verification and correction of PII elements in videos.}
    \label{fig:video-deid-correction}
\end{figure}

To ensure the completeness and accuracy of the de-identification process, we developed custom tools to facilitate human verification and correction of both video and audio data. The first tool is an interactive graphical application (see Figure \ref{fig:video-deid-correction}) that enables annotators to play, seek, and inspect videos frame by frame. Annotators can manually draw regions of interest (ROIs) to mark faces, identification cards, or other objects revealing PII. Once marked, these regions are automatically tracked and blurred throughout the video. Each video was meticulously reviewed and corrected by human annotators, as fully automated methods cannot guarantee perfect accuracy.

In addition, we designed a complementary graphical tool for reviewing and editing de-identified transcripts alongside the corresponding audio. Annotators verified that PII was correctly removed from transcripts and ensured that the corresponding audio segments were properly censored. The use of time-stamped transcripts allowed precise alignment for automated audio censoring, further strengthening privacy safeguards.

% ========================================================= %
\section{Data Collection System (DCS)}
\label{sec:dcs_additional}

\begin{figure}[!h]
    \centering
    \includegraphics[width=0.6\linewidth]{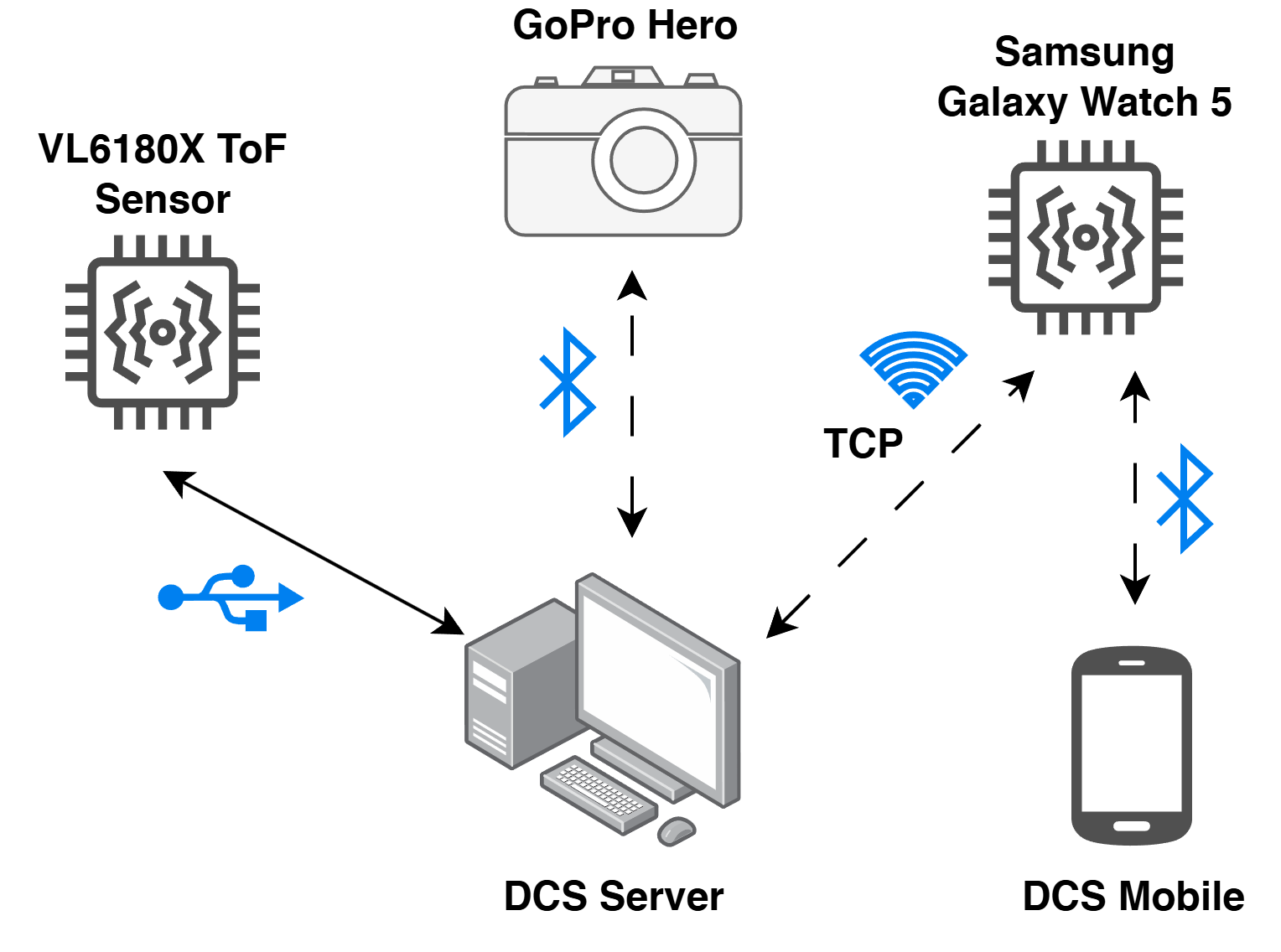}
    \caption{Data collection system architecture.
    }
    \label{fig:dcs_arch}
\end{figure}

Our data collection system integrates multiple sensing devices to capture egocentric video, track EMS responder hand movements via a smartwatch, and record compression rate and depth using a Time-of-Flight (ToF) sensor embedded in a CPR training manikin. The system supports multi-responder simulations through a centralized controller, enabling synchronized recording across multiple participants. To facilitate seamless and efficient data collection, we developed a unified architecture for managing heterogeneous communication protocols across all devices, coordinated through a central local server (see Figure~\ref{fig:dcs_arch}).

\subsection{Cameras}

To effectively capture the complexity of EMS procedures, we employ body-worn cameras during data collection while ensuring it does not interfere with responders activities. The responder is equipped with a chest-mounted GoPro HERO camera, providing an egocentric view that is ideal for capturing detailed hand-object interactions during medical interventions.

In high-fidelity scenarios, each responder was equipped with a GoPro and a smartwatch capturing their individual activities.
The GoPro captures video and audio data with associated timestamps and metadata, synchronized to satellite time. These timestamps are used to compute temporal offsets across modalities, enabling precise alignment and establishing a unified time reference for multimodal synchronization.

\subsection{Sensors}  
\begin{figure}[!h]
    \centering
    \includegraphics[width=0.6\linewidth]{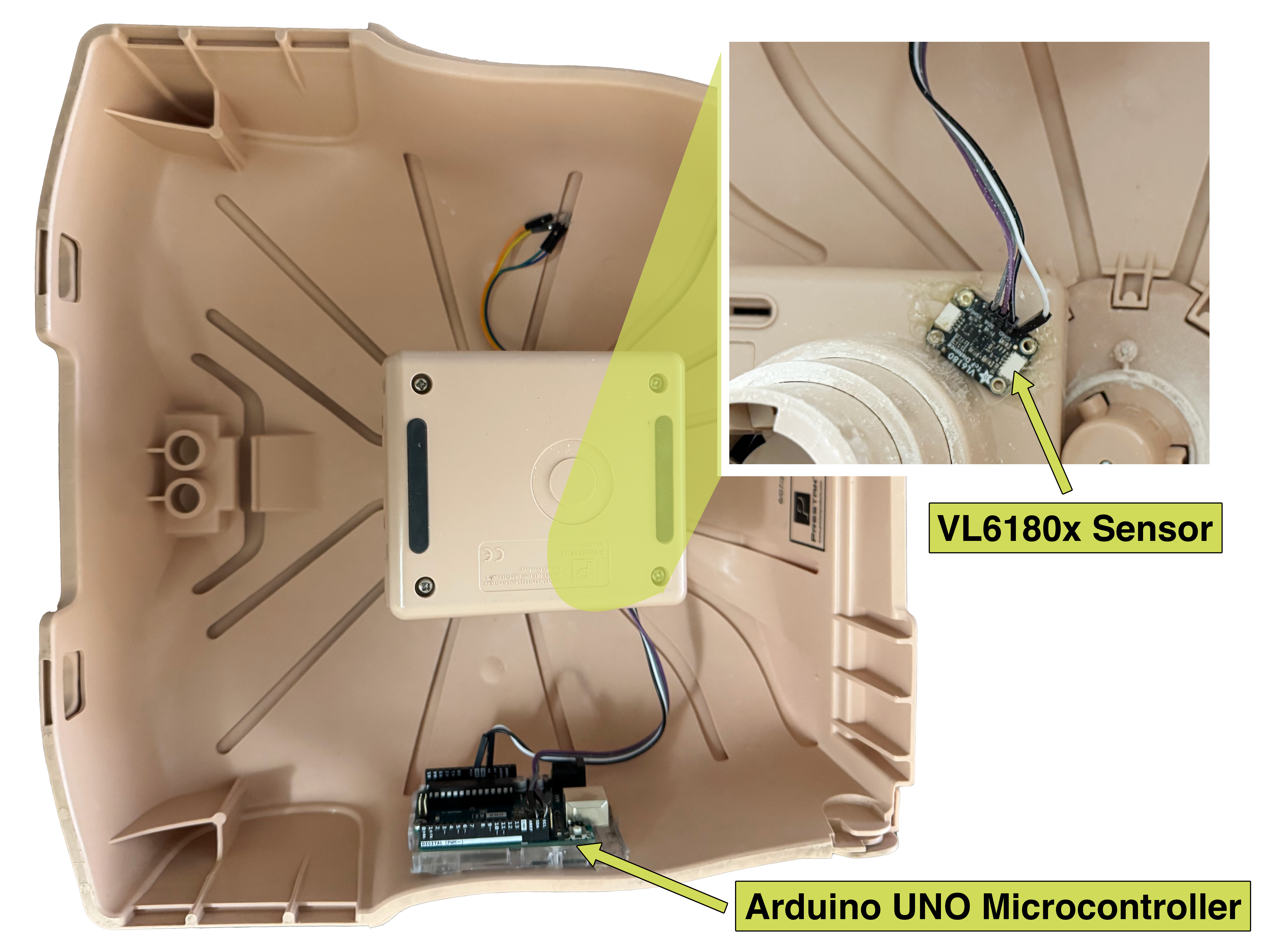}
    \caption{VL6180X sensor placement in the manikin enables accurate capture of compression depth.}
    \label{fig:vl6180x_manikin}
\end{figure}

We employ a Samsung Galaxy Watch 5 and a VL6180X ToF sensor to measure the primary responder's hand movements for activity recognition and quality evaluation. The smartwatch worn on the dominant hand of the responder 
records 3-axis acceleration data via its integrated IMU sensors to precisely track the hand movements during each intervention.
The ToF sensor, embedded in the CPR manikin (see Figure \ref{fig:vl6180x_manikin}), measures the depth of chest compressions performed by the responder during CPR procedures.
The VL6180X sensor employs laser-based ToF measurement for accurate distance detection within a 0–10 cm range along its line of sight. 
We utilize an Arduino Uno microcontroller to communicate with the sensor using the I2C protocol at 100Hz. 

\subsection{Software}

\begin{figure}[!h]
    \centering
    \includegraphics[width=1\linewidth]{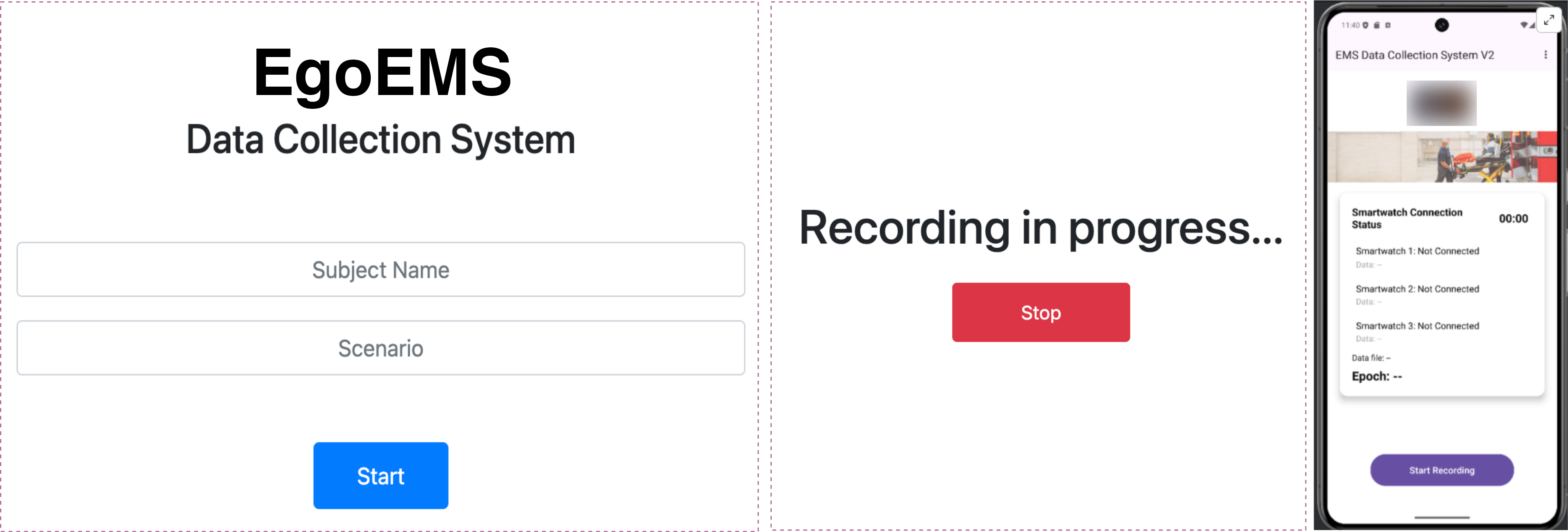}
    \caption{User-friendly simple GUI improves the usability of the data collection system.}
    \label{fig:dcs_software}
\end{figure}

To enable simultaneous recording from multiple data sources, we developed an open-source software suite with a user-friendly GUI (see Figure \ref{fig:dcs_software}) to streamline the data collection process. The software is designed to support low cost, off-the-shelf devices, making the entire system affordable and accessible. Its modular architecture also allows for easy integration of additional devices, ensuring flexibility and scalability for diverse research applications. We also developed an Android smartphone application capable of controlling and recording data from up to three responders’ smartwatches via Bluetooth, enabling practical and mobile use in high-fidelity scenarios where responders move between vehicles or operate in areas without feasible direct Wi-Fi connections to the DCS server.

\subsubsection{Smartphone Application}

We developed an Android-based smartphone application that uses the Bluetooth Low Energy (BLE) protocol to connect with up to three smartwatches simultaneously, each worn by a different responder. The application can be operated by a single user to start and stop recordings, without requiring a central server or Wi-Fi network. This lightweight, decentralized setup enabled high-fidelity data collection in mobile scenarios, including simulations where responders drive an actual ambulance and transport a patient to a destination.

\subsubsection{Smartwatch Application}

We developed an Android WearOS application for the smartwatch that employs a TCP protocol with a handshake mechanism to synchronize timestamps with the server or the smartphone application, ensuring high temporal accuracy. This application enables continuous recording across multiple trials without user intervention, addressing volunteers’ preference for minimal interaction with the device. The smartwatch records data at a frequency of approximately 50 Hz, buffering readings for one second before transmitting them to the server via TCP for reliable data transfer.

\subsubsection{VL6180X Firmware}

We developed a simple C++ program deployed in an Arduino Uno, which records sensor data at 100Hz and transmits the data through serial communication to the server.
The server records the data along with the timestamp at the point of receipt.

\subsection{Data Synchronization}
We developed a toolchain to synchronize multimodal data, including multiperson egocentric video recordings, smartwatch IMU data, and ToF sensor (VL6180X) measurements. The process begins by determining the timestamp offset between the GoPro and the DCS server for each trial due to the lack of a common clock between GoPro cameras and other devices. 
Once the offset is identified, the tool integrates with the main synchronization pipeline to produce synchronized and clipped outputs for the GoPro, smartwatch, and depth sensor data automatically. Additionally, the tool generates a preview video of the synchronized and clipped GoPro recordings, ensuring seamless data alignment across all perspectives.

\subsection{Post Processing}

Upon achieving synchronization, several post-processing steps are conducted. First, the GoPro video is re-encoded using the H.264 codec at 720p resolution, selecting specific video and audio streams to optimize file size and compatibility. We then perform de-identification for GoPro (egocentric) recordings where the faces are blurred using pipeline developed using 
% \homa{mention the name of the tool, e.g., EgoBlur from Facebook} 
EgoBlur \cite{raina2023egoblur}, Grounding DINO \cite{liu2023grounding} and SAM2 \cite{ravi2024sam2} for maximum accuracy in de-identification.
Next, we extract video features, including optical flow, using a pre-trained I3D model \cite{carreira2017quo}, Resnet50 features \cite{he2016deep} utilizing a toolchain provided by \cite{videofeatures2020} and make these features available alongside the original data. Additionally, bounding boxes and segmentation masks are generated using a semi-automatic approach, details of which are provided in a subsequent section.

\subsection{Multimodal Visualization}

We developed a suite of open-source tools to visualize synchronized multimodal data, facilitating a deeper understanding of the dataset and how different views and modalities complement each other in various scenarios. These tools include visualization of hand tracking data and manikin depth sensor data alongside video recordings. We also provide tools for adding role-based captions to videos using timestamped audio transcripts. While these tools are tailored for this dataset, some are versatile and can support multimodal data collection and analysis in other applications. Furthermore, these tools will be released as open-source resources to facilitate reproducibility and enable other research groups to expand and extend the dataset.

% ======================================================== %

% ======================================================== %
\section{Benchmarks}
\label{sec:benchmark_additional}
This section outlines the benchmark tasks, their definitions, various settings, and annotations, followed by a discussion of baseline results and the insights gained. For each task, we explicitly define the inputs and outputs to ensure clarity and consistency for future research utilizing our dataset. The primary objective of these benchmarks is to illustrate the broad applicability of the dataset for research in emergency medical services, rather than to propose new methodological innovations.  Nonetheless, the presented benchmarks offer meaningful performance analyses and insights into the challenges of multimodal, real-time ICA tasks. Current benchmarks use within-view supervision only. Future work targets multiperson activity recognition, inter-responder coordination and cross-view reasoning.

\subsection{Keystep Classification}

\noindent\textbf{Task details.} 
This task is framed as an action classification problem, where the objective is to predict the keystep class associated with each trimmed clip. Given an input clip \( x_{clip} \in \mathbb{R}^{T \times D} \), where \( T \) represents the variable temporal dimension (reflecting different keystep durations) and \( D \) is the feature dimension (varying by modality), the model aims to output a single keystep label \( y \in \mathcal{K} = \{k_1, k_2, \ldots, k_{67}\} \). The model has access to multimodal inputs, denoted by \( x_{clip} = f(x_{\text{ego}}, x_{\text{imu}}, m_1, \ldots, m_n) \), where \( x_{\text{ego}} \) and \( x_{\text{imu}} \) represent the egocentric and smartwatch IMU data, respectively, and \( m_i \) represents other modalities.

The task allows flexibility in data fusion, permitting models to learn from one or more modalities and viewpoints while not limited to a supervised setting.
We evaluate classification performance using class-wise top-k accuracy, F1 score, precision, and recall. 
% To assess sequential accuracy across an entire trial, we combine individual classifications to form a complete trial and measure the Levenshtein edit distance \cite{levenshtein1965binary}. Additionally, for improved generalization, we report the normalized edit distance, referred to as the edit score. 

\noindent\textbf{Annotations.}
For this task, there are 67 keysteps to predict, and the train/validation/test splits are based on trials. The provided dataloader is capable of providing trimmed clips, including multimodal data associated with the keystep annotation. \\

\subsubsection{Fully Supervised Setting} 
In this setting, we define the task as a fully supervised multi-class classification problem. Given an input data clip \(x_{clip}\), the goal is to predict the keystep label \( y \in \mathcal{K} \).

The training objective is to minimize the multi-class cross-entropy loss:
\[
\mathcal{L}_{\text{CE}} = - \frac{1}{N} \sum_{i=1}^{N} \sum_{j=1}^{67} \mathbf{1}_{\{y_i = k_j\}} \log p(y_i = k_j \mid x_i),
\]
where \(N\) is the number of training samples, \(y_i\) is the true keystep label for clip \(i\), and \(p(y_i = k_j \mid x_i)\) is the predicted probability of keystep \(k_j\) for the given clip \(x_i\). \\

\noindent\textbf{Baseline models.} 
We implement MTRSAP \cite{weerasinghe2024multimodal}, a transformer encoder model for action recognition that supports multimodal inputs. To adapt it for our dataset, we adjust key parameters: the embedding dimension (\(d_{\text{model}}\)), the number of layers in the transformer stack, and the number of attention heads to maximize the performance on our dataset.

Our dataset exhibits a long-tailed distribution inherent to the frequency of different keysteps within an intervention. To address this imbalance, we apply Class-Balanced (CB) loss \cite{cui2019class}, which is designed to mitigate the effects of long-tailed distribution.

\begin{equation}
\text{CB Loss} = \frac{1 - \beta}{1 - \beta^{n}} \cdot \mathcal{L}_{\text{CE}}
\end{equation}

Where \( \beta \) is a hyperparameter close to 1, such as 0.99, which controls the degree of balance adjustment. \( n \) is the number of samples for each class. For this task, we set  \( \beta \) to 0.99. We then train MTRSAP with randomly initialized weights for 30 epochs, starting with a learning rate of \(1 \times 10^{-5}\), and applying a learning rate drop after 20 epochs to enhance training stability.

Furthermore, we implement two video classification baselines: TSN~\cite{wang2016temporal}, pretrained on ImageNet~\cite{deng2009imagenet}, and TimeSformer~\cite{gberta_2021_ICML}, a transformer-based architecture initialized with Kinetics-600~\cite{kay2017kinetics} pretraining, and previously used in egocentric video benchmarks~\cite{grauman2024ego}. For this task, we fine-tune both models with our data to assess the performance.

\begin{table*}[!ht]
\centering
\small
\setlength{\tabcolsep}{1mm}
\resizebox{\columnwidth}{!}{%
\begin{tabular}{@{}cclccccc@{}} % 8 columns
\toprule
\multirow{2}{*}{\textbf{Learning Type}} &
\multirow{2}{*}{\textbf{Model (Architecture)}} &
\multirow{2}{*}{\textbf{Modality (Feature Extraction)}} &
\multicolumn{5}{c}{\textbf{Classification Metric}} \\
\cmidrule(l){4-8}
 &  &  & \textbf{Precision} & \textbf{Recall} & \textbf{F1-score} & \textbf{Top-1 Acc.} & \textbf{Top-5 Acc.} \\
\midrule
% --- Supervised group (merged) ---
\multirow{12}{*}{\textbf{Supervised}}
& \multirow{7}{*}{\shortstack{MTRSAP (Transformer) \\ \cite{weerasinghe2024multimodal}}}
& Ego Video (Resnet50) & 0.55 & 0.58 & 0.55 & 62.27 & 77.01 \\
& & Ego Video (I3D) & 0.56 & 0.57 & 0.56 & 60.41 & 72.50 \\
& & Ego Video, IMU (Resnet50, Norm.) & 0.55 & 0.58 & 0.56 & 62.19 & 76.14 \\
& & IMU (Norm.) & 0.42 & 0.45 & 0.42 & 48.34 & 67.57 \\
& & Ego Video, Audio (Resnet50, Mel-Spec) & 0.19 & 0.23 & 0.20 & 25.67 & 44.38 \\
& & Ego Video, Audio (Resnet50, Wav2Vec2) & 0.34 & 0.39 & 0.35 & 44.71 & 66.16 \\
& & Audio (Wav2Vec2) & 0.26 & 0.27 & 0.26 & 33.25 & 52.14 \\
\cmidrule(l){2-8}
& \shortstack{TSN (ConvNet) \\ \cite{wang2016temporal}} 
  & Ego Video (End-to-End) & 0.32 & 0.30 & 0.29 & 57.53 & 76.53 \\
  \cmidrule(l){2-8}
& \shortstack{TimeSformer (Transformer) \\ \cite{gberta_2021_ICML}} 
  & Ego Video (End-to-End) & 0.26 & 0.25 & 0.24 & 50.74 & 73.58 \\
  \cmidrule(l){2-8}
& \shortstack{Audiovisual SlowFast (ConvNet) \\ \cite{xiao2020audiovisual}} 
  & Audio (End-to-End) & 0.00 & 0.00 & 0.00 & 9.18 & 27.39 \\
\midrule
% --- Few-Shot (single row, unmerged) ---
\multirow{1}{*}{\textbf{Few-Shot}}
& \shortstack{MM-CDFSL (Transformer) \\ \cite{Hatano2024MMCDFSL}} 
  & Ego Video (Multimodal Distilled) $\dagger$ & - & - & - & 67.70 & - \\
\midrule
% --- Zero-shot group (merged) ---
\multirow{3}{*}{\textbf{Zero-Shot}}
& \shortstack{VideoLlama-3.3 (VLM) \\ \cite{zhang2025videollama}} 
  & Ego Video (End-to-End) & 0.16 & 0.17 & 0.16 & 18.51 & - \\
& \shortstack{Qwen-2.5 (VLM) \\ \cite{bai2025qwen2}} 
  & Ego Video (End-to-End) & 0.33 & 0.35 & 0.34 & 38.30 & - \\
& \shortstack{GPT-4o (LLM) \\ \cite{achiam2023gpt}} 
  & Audio (End-to-End)      & 0.17 & 0.19 & 0.18 & 23.89 & - \\

\bottomrule
\end{tabular}%
}
\caption{Performance of different models with various modalities for the keystep classification task. Norm: Normalization, Mel-Spec: Mel-spectrogram., MM Distilled: Multimodal Distilled Student Video Encoder. $\dagger$ \textit{Evaluated in a 5-way,5-shot setting. Accuracies are not directly comparable to the fully supervised results, as the few-shot evaluation considers only $N = 5$ classes per episode, yielding a higher apparent top-1 score}.}
\label{tab:classification_results}
\end{table*}

\begin{figure}[!h]
    \centering
    \includegraphics[width=0.8\linewidth]{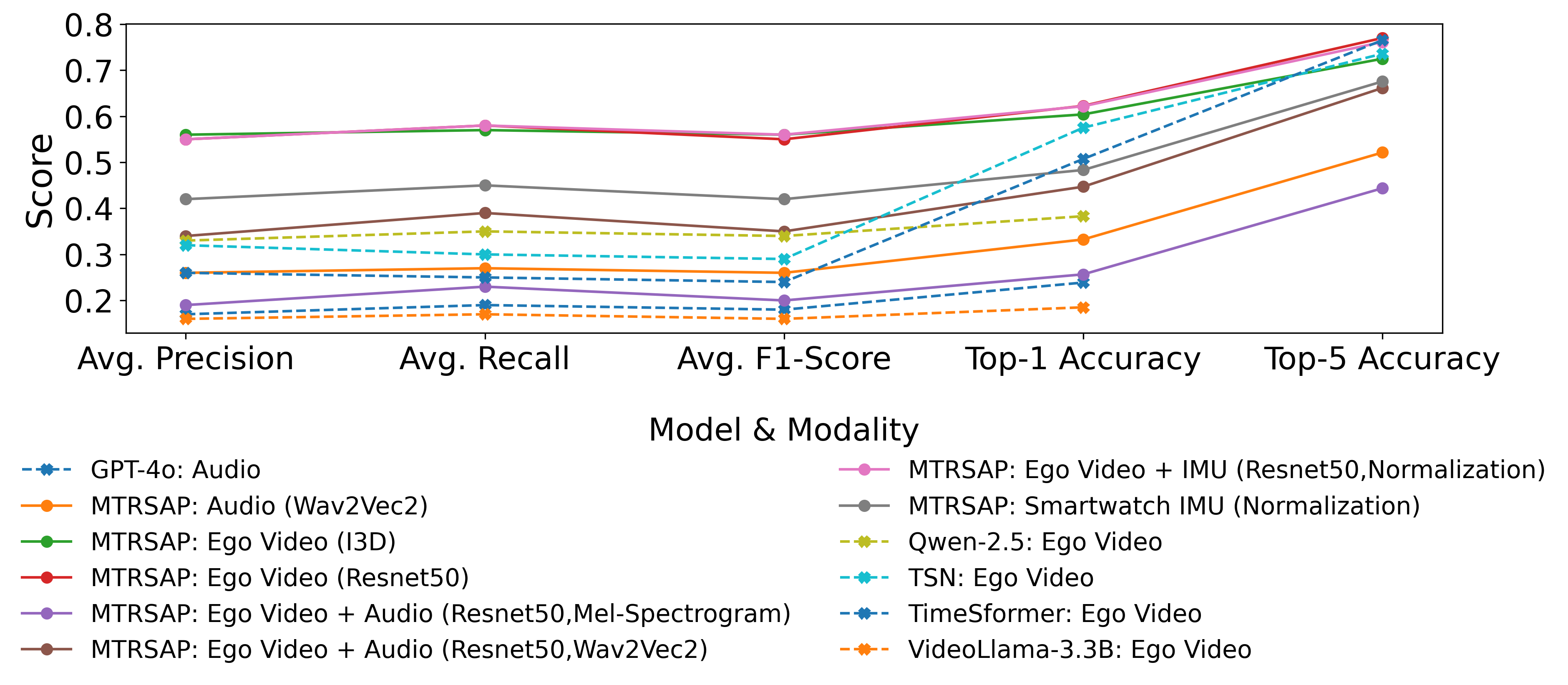}
\caption{Comparison of keystep classification results across different modalities across all settings.}
    \label{fig:keystep_classification}
\end{figure}

\begin{figure*}[!h]
    \centering
    \includegraphics[width=1\textwidth]{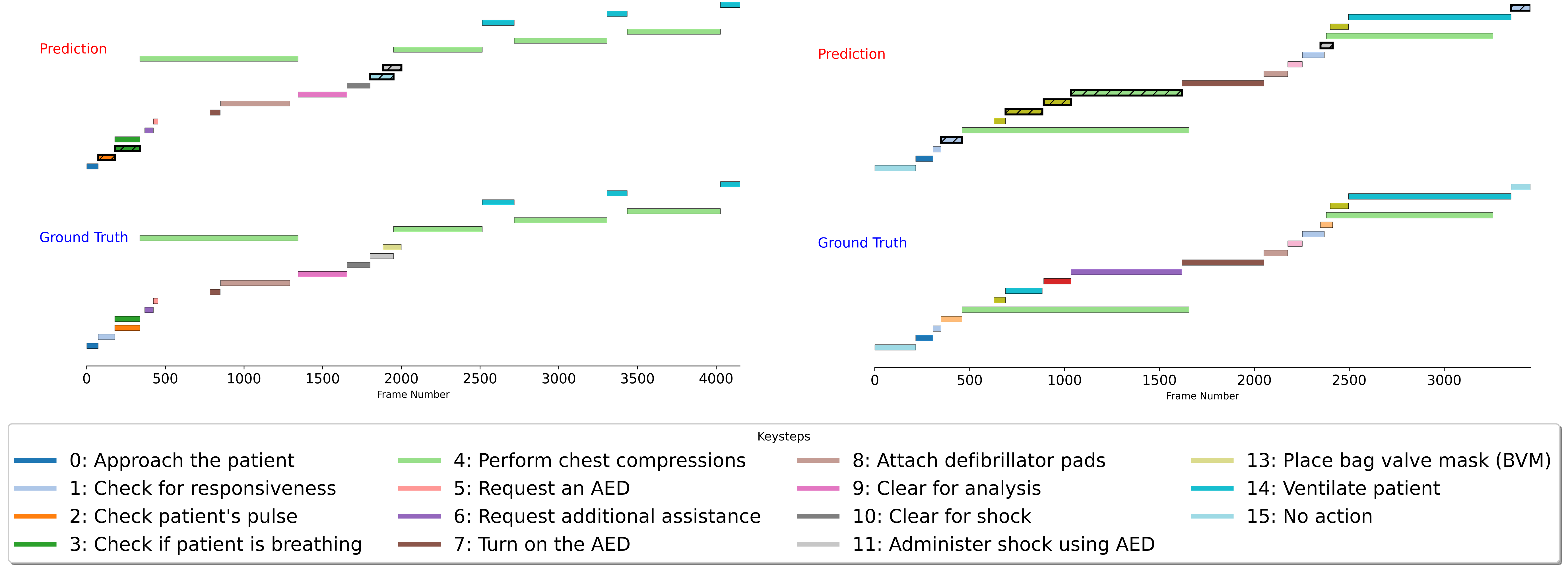}
    \caption{Comparison of ground truth and predicted keysteps for two subjects during keystep classification task.}
    \label{fig:comparison_keystep_classification_i3d}
\end{figure*}

\noindent\textbf{Results.}
Table \ref{tab:classification_results} and Figure \ref{fig:keystep_classification} presents the results under both fully supervised, few-shot and zero-shot settings. MTRSAP \cite{weerasinghe2024multimodal} demonstrates higher performance over both TSN \cite{wang2016temporal} and TimeSformer \cite{gberta_2021_ICML} in the egocentric video only setting, with the best results achieved using features extracted from an Resnet50 \cite{he2016deep} backbone. This configuration also yields the highest top-1 accuracy, resulting a score of \textbf{62.27\%}. Figure \ref{fig:comparison_keystep_classification_i3d} visualizes the predicted keysteps compared to ground truth. 

Fusion of egocentric video with smartwatch IMU features yields the second-highest performance within the MTRSAP framework. While this fusion was expected to improve results by leveraging complementary modalities, it did not surpass the egocentric only configuration. We attribute this to the early fusion strategy employed, which may not fully exploit the temporal and modality specific cues captured by MTRSAP \cite{weerasinghe2024multimodal}. 

\begin{figure}[!h]
    \centering
    \includegraphics[width=0.8\linewidth]{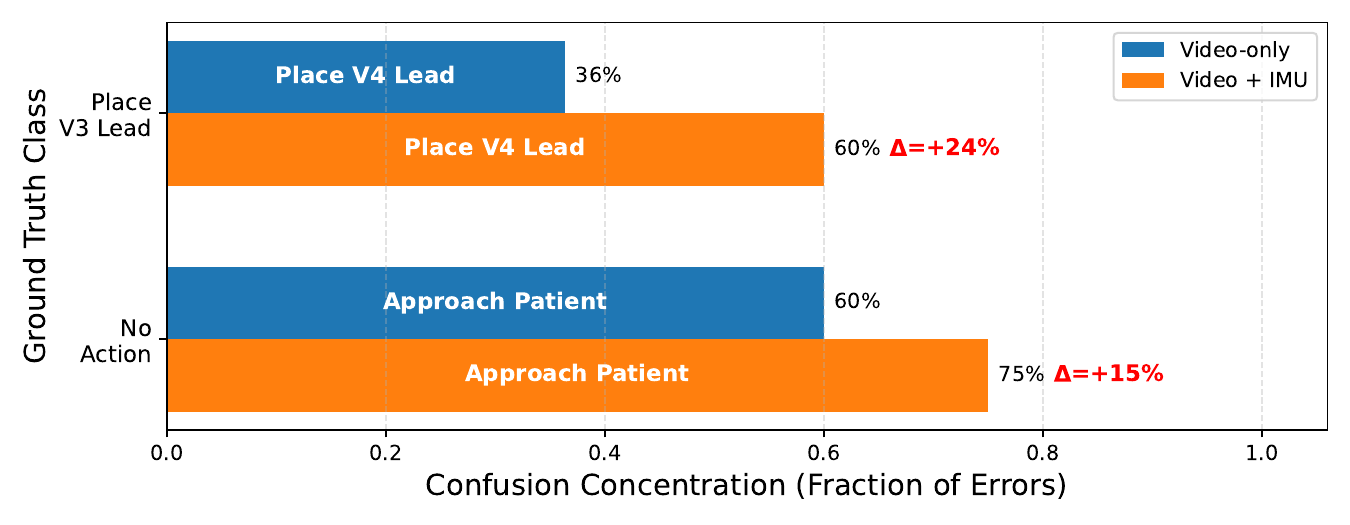}
\caption{Class confusion concentration for selected keysteps exhibiting poor performance.}
    \label{fig:imu_confusion}
\end{figure}

To further investigate this, we analyze the top five confusion pairs when smartwatch IMU features are early–fused with video features. Within these, we focus on two representative keysteps: ``Place V3 Lead'' and ``No Action''. Figure~\ref{fig:imu_confusion} shows that early fusion can \emph{increase} error concentration for both classes. For ``No Action'', the video-only model already confuses it with ``Approach Patient'' about $60\%$ of the time; adding IMU raises this to $75\%$ ($\Delta=+15\%$). For ``Place V3 Lead'', the top confusion partner is ``Place V4 Lead'': IMU fusion increases the concentration from $36\%$ (video-only) to $60\%$ ($\Delta=+24\%$). In both cases, wrist motion captured by the watch is either weak/ambiguous (idle segments resembling an initial approach) or highly similar across adjacent lead placements, so simple early fusion nudges the classifier toward the most kinematically similar neighbor even when the video signal alone is more balanced.

% \begin{figure}[!h]
%     \centering
%     \includegraphics[width=\linewidth]{ArxivVersion/assets/cars_11_0_ks63_7320-7382.png}
% \caption{imu checkao investigate.}
%     \label{fig:imu_checkao}
% \end{figure}

\begin{table}[!h]
\centering

\small
\begin{tabular}{@{}lccc@{}}
\toprule
\textbf{Group} & \textbf{\#Pairs} & \textbf{Cos.} & \textbf{DTW Dist.} \\
\midrule
Within KS25 & 156 & 0.25 $\pm$ 0.09 & 0.055 $\pm$ 0.02 \\
Within KS26 & 132 & 0.24 $\pm$ 0.07 & 0.048 $\pm$ 0.02 \\
Cross KS25$\leftrightarrow$KS26 & 156 & 0.26 $\pm$ 0.08 & 0.048 $\pm$ 0.02 \\
\bottomrule
\end{tabular}
\vspace{1em}
\caption{Pairwise similarity of IMU acceleration segments for \textit{place V3 lead} (KS25) and \textit{place V4 lead} (KS26). Cos.: Centered Cosine, Higher centered cosine and lower DTW indicate greater similarity.}
\label{tab:imu_similarity_ks25_ks26}
\end{table}

\begin{figure}[t]
  \centering
  \begin{minipage}[t]{0.8\linewidth}
    \centering
    \includegraphics[width=\linewidth]{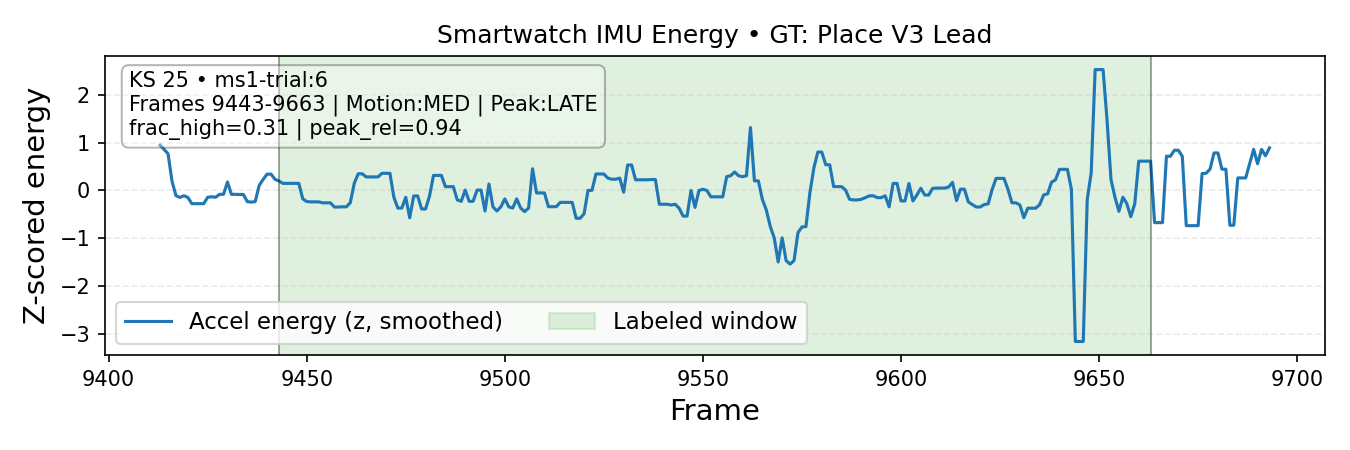}
    \par\smallskip
    {\small (a) IMU energy (accelerometer magnitude) for KS25: \emph{Place V3 Lead}.}
  \end{minipage}

  \medskip

  \begin{minipage}[t]{0.8\linewidth}
    \centering
    \includegraphics[width=\linewidth]{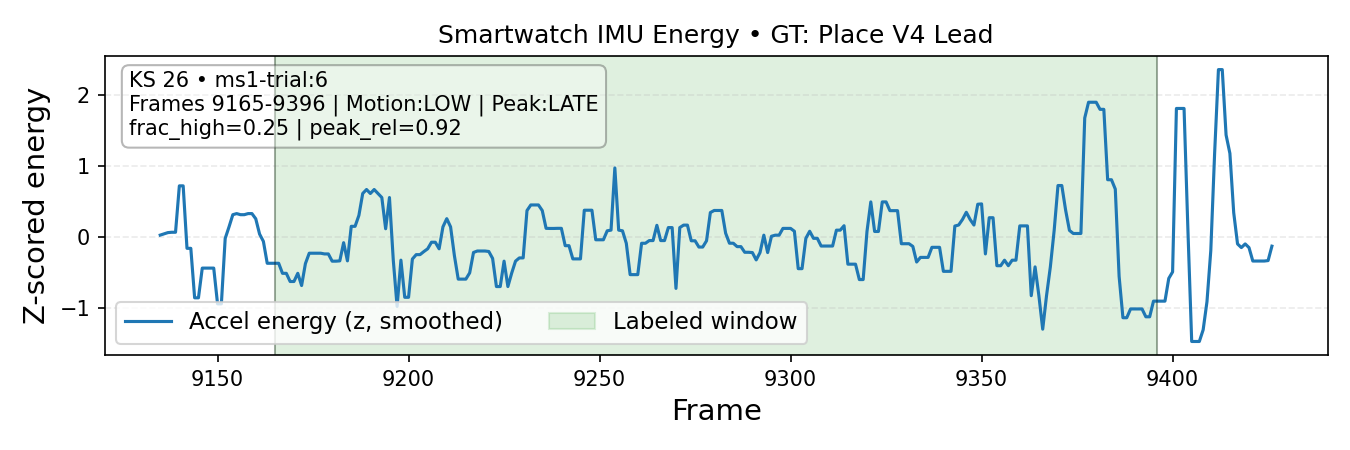}
    \par\smallskip
    {\small (b) IMU energy (accelerometer magnitude) for KS26: \emph{Place V4 Lead}.}
  \end{minipage}

  \caption{Smartwatch IMU energy traces for two adjacent keysteps. Both panels show comparable amplitude and temporal profiles within the labeled windows, indicating limited discriminability in the IMU modality and aligning with the similarity statistics in Table~\ref{tab:imu_similarity_ks25_ks26}. Higher values denote increased wrist motion.}
  \label{fig:smartwatch_imu_comparison}
\end{figure}

To assess whether this ambiguity is systematic, we compared IMU segments across KS25 (Place V3 Lead) and KS26 (Place V4 Lead) over all subjects (Table~\ref{tab:imu_similarity_ks25_ks26}). The distributions are nearly indistinguishable: centered cosine $\approx 0.25\pm0.09$ (within KS25), $0.24\pm0.07$ (within KS26), and $0.26\pm0.08$ (cross); DTW $\approx 0.055\pm0.02$, $0.048\pm0.02$, and $0.048\pm0.02$, respectively. This indicates that smartwatch motion patterns for placing V3 lead and V4 lead are as similar \emph{across} classes as they are \emph{within} class, providing little discriminative signal and occasionally sharpening the wrong attractor during fusion. This effect is further illustrated in Fig.~\ref{fig:smartwatch_imu_comparison}, which plots the normalized IMU \emph{energy} (magnitude of acceleration) for representative segments of the two keysteps. 

Here, IMU energy is computed as the instantaneous acceleration magnitude,
\begin{equation}
E(t) = \sqrt{x(t)^2 + y(t)^2 + z(t)^2},
\end{equation}
and is z-scored within each segment for visualization. The metric $\textit{frac\_high}$ denotes the fraction of frames exceeding a robust activity threshold defined as the median plus half the interquartile range (IQR), serving as a proxy for the amount of motion within the labeled window. The $\textit{peak\_rel}$ indicates the relative temporal position of the maximum energy (0=start, 1=end), while the qualitative tag “Motion:LOW/MED/HIGH” is derived from $\textit{frac\_high}$ and highlights the overall activity intensity. As seen, both keysteps exhibit medium to low motion with similar peak timing (“LATE”), confirming that their wrist motion signatures are strongly overlapping and therefore difficult to separate using simple early fusion.

These results suggest moving beyond simple early fusion. Practical mitigations include \emph{adaptive/attention-based fusion} that downweights IMU when its discriminability is low, \emph{late or gated fusion} (e.g., confidence/uncertainty-aware weighting across modalities), and \emph{class-aware fusion policies} that learn to rely on IMU only for motion-distinct keysteps while prioritizing visual evidence for low-motion or near-duplicate actions.

Overall, these findings highlight that although the proposed models provide reasonable supervised baselines, advancing multimodal reasoning will require moving beyond simple early fusion toward adaptive, context-aware integration strategies.
Additionally, as shown in Table~\ref{tab:classification_results}, both the audio-only and audio-video fusion settings yield the lowest performance among all modality configurations. We attribute this to two primary factors: (1) the use of a simple early fusion strategy, which may limit effective cross-modal interaction, and (2) inherent misalignment between audio cues and physical actions. In many cases, responders either do not verbalize their actions or do so asynchronously before or after the keystep occurs reducing the utility of audio for fine-grained action recognition. Notably, using Wav2Vec2~\cite{baevski2020wav2vec} for audio feature extraction significantly improves performance over mel-spectrograms, indicating the importance of high-quality temporal representations. Further analysis of the role of audio is provided in our zero-shot evaluation setting.

\subsubsection{Multimodal Few-Shot Setting}
Knowledge transfer and multimodal distillation show promising performance in low-data regimes. To assess cross-domain transfer, we evaluate cross-domain, multimodal few-shot learning with MM-CDFSL~\cite{Hatano2024MMCDFSL}. Following the authors, we use the \emph{distilled} RGB student encoder on a large scale general purpose egocentric dataset (Ego4D \cite{grauman2022ego4d}) (teachers: optical flow and hand pose)  and perform episodic 5-way, 5-shot evaluation on our dataset. For each episode, a linear classifier (logistic regression) is fit on the support set and evaluated on the query set using RGB only; we also apply ensemble masked inference as in the original work. We report the mean top-1 accuracy across 600 episodes.

\noindent\textbf{Results.}
Under the 5-way, 5-shot protocol (600 episodes), MM-CDFSL’s distilled RGB student attains a mean top-1 accuracy of \textbf{67.70\%}. The per-episode variability is \textbf{11.39} percentage points (std), yielding a standard error of \textbf{0.47} points and an approximate 95\% CI of \textbf{[66.8, 68.6]\%}. This indicates that, given only five labeled clips per class, the distilled features support strong within-episode discrimination among five candidate keysteps. (Note: N-way few-shot accuracy is not directly comparable to full multi-class supervision, since each episode evaluates over only five classes.)

\subsubsection{Video-Based Zero-Shot Setting}

Vision Large Language Models (VLMs) shows outstanding performance to any visual classification benchmark by simply providing the names of the visual categories to be recognized. We choose state-of-the-art VLMs including Qwen-2.5-VLM~\cite{bai2025qwen2} and Videollama-3.3~\cite{zhang2025videollama} on the keystep classification in a zero-shot manner. Specifically, given a video clip \( x_{clip} \), we used the prompt on VLMs to derive the prediction \( y\) on \( x_{clip} \). The prompts we used are shown in Figure~\ref{fig:vlm_cls_prompt}.

\begin{figure}[!h]
\centering
\begin{tcolorbox}[breakable=false]
\noindent\textbf{PROMPT: }Here is a EMS video. Identify the keystep (action first responder is taking) in the video. There is only one keystep in the video.

Return your final answer as a numerical value in strict JSON format, like: ["1"]

Here are all keysteps:
\{keystep\}
\end{tcolorbox}
\caption{Keystep Classification Prompt template for VLMs.}
\label{fig:vlm_cls_prompt}
\end{figure}

\noindent\textbf{Results.}
As shown in Table~\ref{tab:classification_results}, the zero-shot Videollama-3.3 model has the worst performance compared with other baselines. Qwen-2.5 (VLM) has the best performance among all zero-shot setting models. However it only achieves 38.30 accuracy which is considerably low for the keystep classification task. One reason why VLMs underperform is that our dataset is an EMS domain-specific task which VLMs may not generalize well for unseen tasks.

\subsubsection{Audio-Based Zero-Shot Setting} 
\begin{figure}
    \centering
    \includegraphics[width=0.6\linewidth]{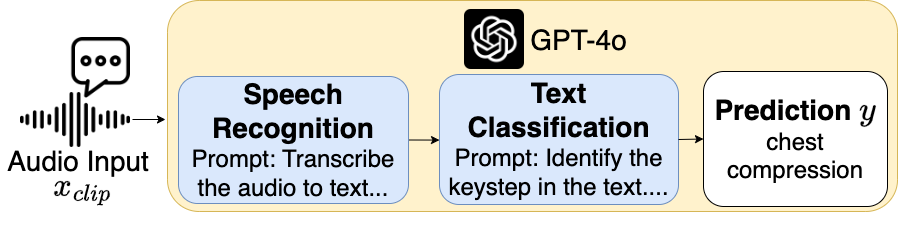}
    \caption{Audio-based keystep classification.}
    \label{fig:audio_keystep_cls}
\end{figure}

Large Language Models (LLMs) have shown excellent abilities at various tasks in zero-shot manners~\cite{achiam2023gpt}. In this task, we investigate keystep classification performance by zero-shot prompting LLMs. Specifically, only audio will be used as the input while the output contains the keystep prediction for the audio.
As shown in Figure~\ref{fig:audio_keystep_cls}, GPT-4o~\cite{achiam2023gpt} is prompted to first do speech recognition by transcribing audio to text and followed by classifying which keystep prediction $y$ is included in the audio. 

\noindent\textbf{Results.}
\label{sec:error_analysis}
In Table~\ref{tab:classification_results}, while GPT-4o does not outperform the top-performing baseline, it demonstrates strong zero-shot keystep classification capabilities compared to several fully-supervised models.

\begin{figure}[!t]
    \centering
    \includegraphics[width=0.6\linewidth]{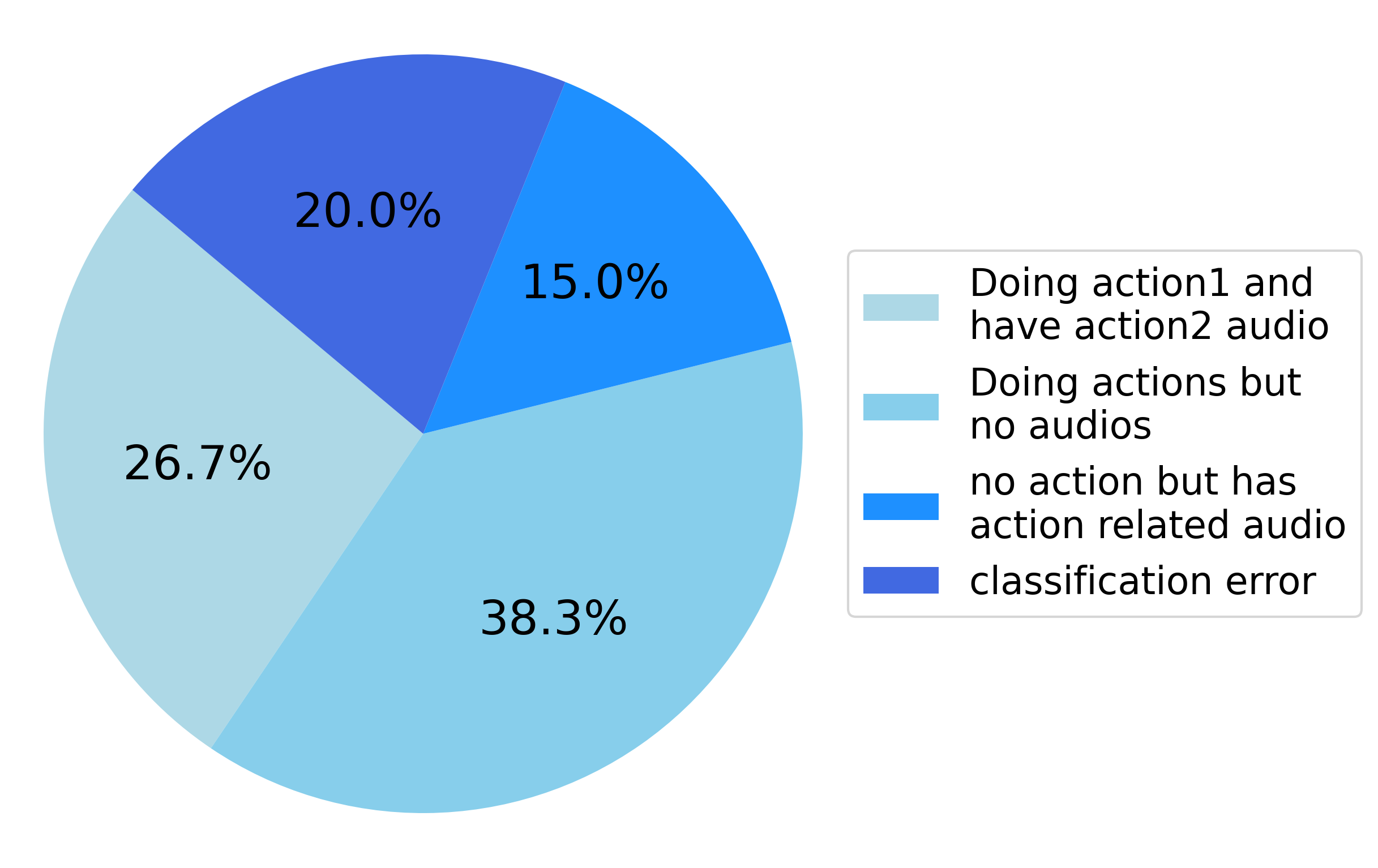}
    \caption{Error analysis for audio-based keystep classification.}
    \label{fig:audio_keystep_cls_error}
\end{figure}

To further explore the limitations and advantages of using audio as an input modality, we conducted an error analysis on GPT-4o's keystep classification. We identified a total of 290 errors, which we reviewed manually to categorize into four primary error types (see Figure~\ref{fig:audio_keystep_cls_error}). The majority of errors (38\%) occur when first responders perform actions without verbalizing the associated keysteps. Additionally, 26.7\% of errors arise when first responders verbalize one action while performing a different one. Another 15.0\% of errors occur when first responders are not performing any actions, yet keystep-relevant audio is still present. Only 20\% of the errors are due to misclassifications by GPT-4o. The results suggest that using audio as the input modality introduces missing or noisy information related to keysteps. It validates why other modalities, such as video, are necessary as additional sources for keystep classification.

\subsection{Keystep Segmentation}

\noindent\textbf{Task details.}
We formulate this as an online video segmentation task, where the objective is to segment an entire video \( V_{t} \) of a trial into specific keysteps. Given the video \( V_{t} \), we define a context window of \(x_{\text{window}} = 4 \) seconds, representing the duration of each segment to be analyzed for keystep recognition. The video \( V_{t} \) is decomposed into non-overlapping windows, resulting in a sequence of segments \( \{V_{t}^{(i)}\}_{i=1}^{N} \), where each window \( V_{t}^{(i)} \) corresponds to a 5-second clip from \( V_{t} \), and \( N = \frac{\text{len}(V_{t})}{x_{\text{window}}} \) represents the total number of segments.

Each segment \( V_{t}^{(i)} \) is analyzed at the frame/sample level to classify the keystep(s) occurring within that time frame, aiming to capture both the current keystep(s) and transitions to subsequent keysteps. The online setting imposes a constraint such that segmentation must rely solely on information within the current window or the cumulative context of the current and past windows up to \( V_{t}^{(i)} \), rather than on the entire video \( V_{t} \). This setup emulates real-time processing conditions, where the model must generate accurate keystep predictions and segment boundaries dynamically as the video progresses.

For each video frame or data sample within the window \( V_{t}^{(i)} \), the model outputs a predicted keystep label \( y_{j}^{(i)} \), where \( j \) indexes the frames/samples within \( V_{t}^{(i)} \). Consequently, the sequence \( \{ y_{j}^{(i)} \}_{j=1}^{M_i} \) (where \( M_i \) is the number of frames in window \( V_{t}^{(i)} \)) provides a fine-grained keystep timeline across \( V_{t} \), detailing keystep predictions for each frame or sample within every context window. 

We then combine the windows to form a complete trial to assess the overall segmentation and recognition accuracy. We employ the segmental F1@k score proposed by \cite{lea2017temporal}, which is well-suited for detection and segmentation evaluation. We evaluate performance using three overlap thresholds \( k = \{0.1, 0.25, 0.5\} \), and additionally report class-wise accuracy. The edit score is also used to assess the sequential accuracy of keystep predictions, providing a measure of alignment with the true keystep order. 

\noindent\textbf{Annotations.}
This task requires detecting 67 keysteps along with their precise temporal boundaries. We provide frame-level and sample-wise keystep labels for each 5-second window. The dataset is organized into train, validation, and test splits at the trial level, with each trial segmented into consecutive 5-second windows. %Due to the nature of this task, 
We automatically fill gaps in keysteps with ``no\_action" class to ensure the continuity of the annotations. Further, for simplicity, these annotations only involve sequential keysteps in contrast to the classification task annotations. The provided data loader generates these windows, incorporating multimodal data aligned with the keystep annotations. \\

\begin{table*}[!ht]
\centering
\small
\setlength{\tabcolsep}{1mm}
\resizebox{\columnwidth}{!}{%
\begin{tabular}{ccccccccc}
\toprule
\multirow{2}{*}{\textbf{Learning Type}} &
\multirow{2}{*}{\textbf{Model (Architecture)}} & 
\multirow{2}{*}{\shortstack{\textbf{Modality} \\ \textbf{(Feature Extraction)}}} & 
\multirow{2}{*}{{\textbf{IOU}}} & 
\multicolumn{5}{c}{\textbf{Evaluation Metric}} \\ \cmidrule{5-9} 
&  &  &  & \textbf{Precision} & \textbf{Recall} & \textbf{F1-score} & \textbf{Accuracy} & \textbf{Edit Score} \\ 
\midrule
% --- Supervised group (merged) ---
\multirow{24}{*}{\textbf{Supervised}} &
\multirow{24}{*}{\shortstack{MTRSAP (Transformer) \\ 
\cite{weerasinghe2024multimodal}}} & \multirow{3}{*}{Ego Video (Resnet50)} & 10\% & 0.54 & 0.53 & 0.53 & \multirow{3}{*}{0.55} & \multirow{3}{*}{0.55} \\
 &   &  & 25\% & 0.51 & 0.50 & 0.50 &  &  \\
&   &  & 50\% & 0.45 & 0.44 & 0.44 &  &  \\ \cmidrule{3-9}
 & & \multirow{3}{*}{Ego Video (I3D)} 
 &  10\% & 0.54 & 0.57 & 0.54 & \multirow{3}{*}{0.56} & \multirow{3}{*}{0.56} \\
 & &  & 25\% & 0.52 & 0.53 & 0.52 &  &  \\
  &  &  & 50\% & 0.49 & 0.50 & 0.49 &  &  \\ \cmidrule{3-9}
  & & \multirow{3}{*}{IMU (Normalization)} & 10\% & 0.56 & 0.44 & 0.40 & \multirow{3}{*}{0.49} & \multirow{3}{*}{0.49} \\
  & &  & 25\% & 0.40 & 0.41 & 0.37 &  &  \\
 &  &  & 50\% & 0.32 & 0.37 & 0.33 &  &  \\ \cmidrule{3-9}
  & & \multirow{3}{*}{\shortstack{Ego Video + IMU \\ (Resnet50, Normalization)}} & 10\% & 0.59 & 0.56 & 0.56 & \multirow{3}{*}{0.61} & \multirow{3}{*}{0.61} \\
 &  &  & 25\% & 0.56 & 0.53 & 0.53 &  &  \\
  & &  & 50\% & 0.49 & 0.47 & 0.47 &  &  \\ \cmidrule{3-9}
 & & \multirow{3}{*}{\shortstack{Ego Video + Audio \\ (Resnet50, Wav2Vec2)}} & 10\% & 0.41 & 0.39 & 0.38 & \multirow{3}{*}{0.43} & \multirow{3}{*}{0.43} \\
 & &  & 25\% & 0.36 & 0.33 & 0.33 &  &  \\
  & &  & 50\% & 0.26 & 0.22 & 0.23 &  &  \\ \cmidrule{3-9}
 & & \multirow{3}{*}{\shortstack{Ego Video + Audio \\ (Resnet50, Mel-spectrogram)}} & 10\% & 0.26  & 0.21 & 0.22  & \multirow{3}{*}{0.29} & \multirow{3}{*}{0.29} \\
 & &  & 25\% & 0.22 & 0.17 & 0.19 &  &  \\
 & &  & 50\% & 0.15 & 0.11  & 0.13 &  &  \\ \cmidrule{3-9}
 & & \multirow{3}{*}{{Audio (Wav2Vec2)}} & 10\% & 0.27 & 0.29 & 0.26 & \multirow{3}{*}{0.30} & \multirow{3}{*}{0.30} \\
& &  & 25\% & 0.24 & 0.23 & 0.22 &  &  \\
 & &  & 50\% & 0.16 & 0.13 & 0.14 &  &  \\ 
 %  & \multirow{3}{*}{{Exo Video (Resnet50)**}} & 10\% & 0.79 & 0.72 & 0.72 & \multirow{3}{*}{0.75} & \multirow{3}{*}{0.75} \\
 % &  & 25\% & 0.77 & 0.70 & 0.75 &  &  \\
 % &  & 50\% & 0.70 & 0.65 & 0.67 &  &  \\ 
\midrule

\multirow{9}{*}{\textbf{Zero-shot}} &
\multirow{3}{*}{\shortstack{VideoLlama-3.3 (VLM) \\ \cite{zhang2025videollama}}} & \multirow{3}{*}{\shortstack{Ego Video \\ (End-to-End)}} & 10\% & 0.07 & 0.51 & 0.12 & \multirow{3}{*}{0.33} & \multirow{3}{*}{0.34} \\
  & & & 25\% & 0.06 & 0.33 & 0.09 & & \\
  & & & 50\% & 0.03 & 0.11 & 0.04 & & \\ 
\cmidrule{2-9}
& \multirow{3}{*}{\shortstack{Qwen-2.5 (VLM) \\ \cite{bai2025qwen2}}} & \multirow{3}{*}{\shortstack{Ego Video \\ (End-to-End)}} & 10\% & 0.26 & 0.64 & 0.33 & \multirow{3}{*}{0.55} & \multirow{3}{*}{0.56} \\
& & & 25\% & 0.25 & 0.61 & 0.31 & & \\
& & & 50\% & 0.20 & 0.40 & 0.24 & & \\ 
\cmidrule{2-9}

& \multirow{3}{*}{\shortstack{WhisperTimestamp + GPT-4o (LLM) \\ ~\cite{achiam2023gpt}}} & \multirow{3}{*}{\shortstack{Audio \\ (End-to-End)}} 
& 10\% & 0.10 & 0.61 & 0.17 & \multirow{3}{*}{0.38} & \multirow{3}{*}{0.39} \\
& & & 25\% & 0.07 & 0.45 & 0.09 & & \\
& & & 50\% & 0.03 & 0.21 & 0.05 & & \\

\bottomrule
\end{tabular}%
}
\caption{Performance of different models for the keystep segmentation task using a 5-second window.}
\label{tab:segmentation_results}
\end{table*}

\begin{figure*}[!ht]
    \centering
    \includegraphics[width=\textwidth]{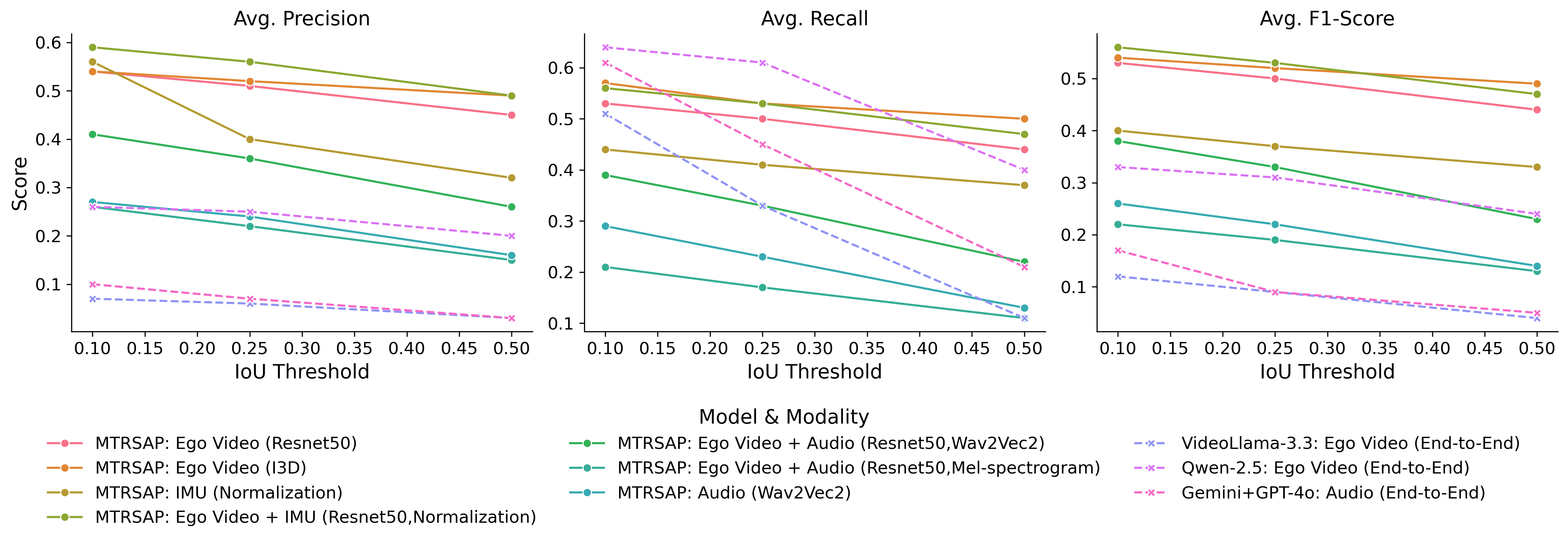}
\caption{Comparison of keystep segmentation results across different modalities in the supervised setting.}
    \label{fig:results_supervised_keystep_segmentation}
\end{figure*}

\begin{figure}[!h]
  \centering
  \begin{tcolorbox}[breakable=false,width=\columnwidth,]
\textbf{PROMPT:} You will receive a 5‑s EMS video clip and a list of all possible keysteps. Your task is to identify every keystep (i.e., the specific action the first responder performs) and mark its start and end times in the video.

\begin{itemize}
  \item Time format: seconds with two decimal places (e.g., \texttt{1.23})
  \item Output format: a JSON array of entries, each entry an object with three fields:
    \begin{itemize}
      \item \texttt{"keystep\_id"}: the ID (as a string) from the provided list
      \item \texttt{"start\_time"}: time the action begins, must $\ge0.00$
      \item \texttt{"end\_time"}: time the action ends, must $\le5.00$
    \end{itemize}
\end{itemize}

Return only valid JSON—no extra text. For example:
\begin{lstlisting}
[
 {
  "keystep id": "1",
  "start time": "0.00",
  "end time": "1.20"
 },
 {
  "keystep id": "4",
  "start time": "1.21",
  "end time": "2.75"
 }
]
\end{lstlisting}
% [
% \par
% \quad\{\par
% \quad\quad"keystep\_id": "1",\par
% \quad\quad"start\_time": "0.00",\par
% \quad\quad"end\_time": "1.20"\par
% \quad\},\par
% \quad\{\par
% \quad\quad"keystep\_id": "4",\par
% \quad\quad"start\_time": "1.21",\par
% \quad\quad"end\_time": "2.75"\par
% \quad\}\par
% ]\par
Here are the keysteps:
\begin{verbatim}
{keystep}
\end{verbatim}
  \end{tcolorbox}
  \caption{Keystep segmentation prompt pemplate for VLMs.}
  \label{fig:vlm_seg_prompt}
\end{figure}

\subsubsection{Fully Supervised Setting} 
In contrast to the keystep classification task, where a single label is predicted for a trimmed clip, in keystep segmentation, the model is tasked with predicting the keystep label for each frame within a window of data \( x_{\text{window}} \). Given an input window \( x_{\text{window}} \in \mathbb{R}^{T \times D} \), where \( T \) is the temporal length of the window and \( D \) is the feature dimension, the objective is to output a sequence of frame-level predictions \( \{y_t\}_{t=1}^{T} \), where \( y_t \in \mathcal{K} = \{k_1, k_2, \ldots, k_{67}\} \) represents the keystep label for each frame \( t \). 

\noindent\textbf{Baseline models.} 
Similar to the keystep classification task, we implement MTRSAP \cite{weerasinghe2024multimodal} with the same parameters as in the classification task and the class-balanced loss. Due to the nature of this task, we were unable to utilize TimeSformer \cite{gberta_2021_ICML} and TSN \cite{wang2016temporal} as a supervised baseline.

\noindent\textbf{Results.} 
Table~\ref{tab:segmentation_results} includes the results under the fully supervised setting. MTRSAP~\cite{weerasinghe2024multimodal} achieves the best performance when fusing egocentric video features (extracted using a ResNet50~\cite{he2016deep} backbone) with smartwatch IMU data. We attribute this improvement to the complementary nature of wrist motion patterns, which provide fine-grained temporal cues within short context windows. These cues help the model disambiguate and temporally localize keysteps more effectively than using egocentric video alone.

Additionally, as expected, segmentation accuracy declines with increasing overlap thresholds due to the model's reduced boundary precision (see Figure~\ref{fig:results_supervised_keystep_segmentation}). Overall, sequence-level accuracy also peaks when the egocentric and IMU modalities are fused.

Notably, the video and audio fusion seems to have a comparable accuracy compared to the classification task. We attribute this to the model consuming a smaller 5-second context window. We believe that the shorter window size allows the model to capture immediate audio cues that align more closely with responder actions, enhancing the utility of audio features in segmentation. This suggests a shorter context is beneficial in scenarios where real-time audio aligns with visual cues, improving segmentation accuracy for certain keysteps, especially in an online recognition setting.

\subsubsection{Video-Based Zero-Shot Setting}
We employ VLMs like Videollama-3.3 and Qwen-2.5 in a zero-shot setting to predict a keystep label \( \{y_t\}_{t=1}^{T} \) and its start and end timestamp \( t_{start}, t_{end} \) within \( x_{\text{window}} \). We use the prompt shown in Figure~\ref{fig:vlm_seg_prompt}.

\noindent\textbf{Results.}
As shown in Table~\ref{tab:segmentation_results}, the Qwen-2.5 VLM model has the best performance compared with other zero-shot baselines. Notably, all zero-shot VLMs perform worse in segmentation than classification. Since the model need to predict both time boundaries and the keystep within this time boundary, there are more noisy predictions in $x_{window}$, which leads to worse performance.

\subsubsection{Audio-Based Zero-Shot Setting}

In this task, we investigate keystep segmentation performance by zero-shot prompting LLMs. Specifically, only the context window of audio \(x_{\text{window}}\) will be used as the input. The output is a sequence of time-level predictions \( \{y_t\}_{t=1}^{T} \) and its corresponding start and end timestamp.\newline

\begin{figure}[!h]
    \centering
    \includegraphics[width=0.6\linewidth]{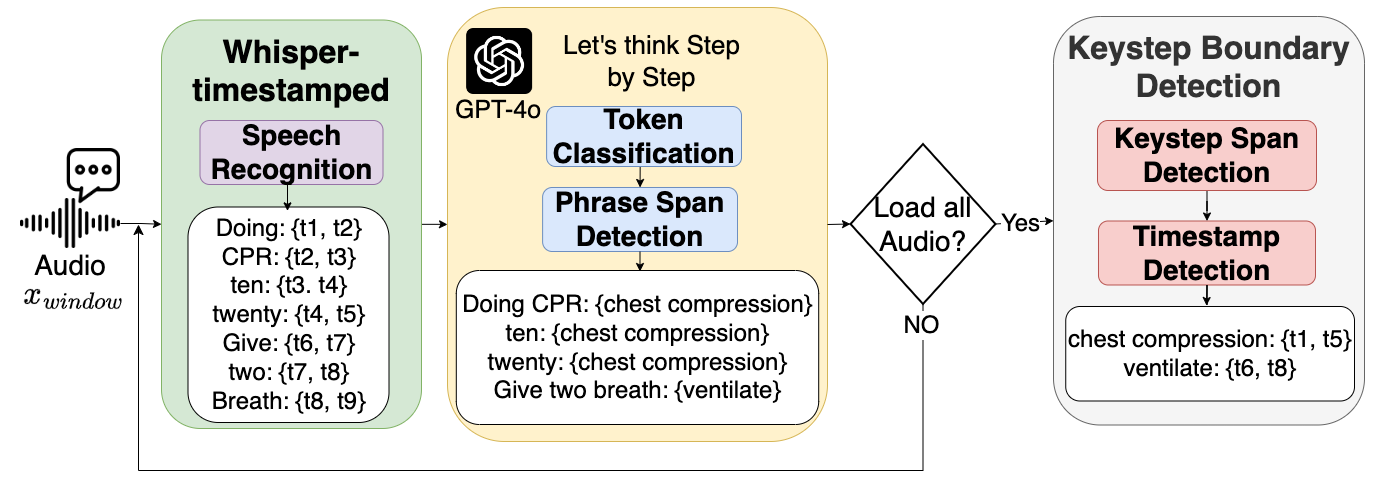}
    \caption{Audio-based keystep segmentation.}
    \label{fig:audio_keystep_recognition}
\end{figure}

\noindent\textbf{Method.} As shown in the Figure~\ref{fig:audio_keystep_recognition}, our method contains three main modules: i) Whisper-timestamped~\cite{lintoai2023whispertimestamped} takes audio \(x_{\text{window}}\) as input and conducts automatic speech recognition with word-level timestamps; ii) GPT-4o~\cite{achiam2023gpt} is prompted in Chain-of-Thought~\cite{wei2022chain} to decompose keystep detection task into two sub-tasks~\cite{ge-etal-2024-dkec}: token classification and phrase span detection and ask LLMs to think step by step. The input for GPT-4o is the transcripts generated by Whisper-timestamped, and the output is the phrases from the transcripts, each labeled with keystep type. The detailed prompt is shown in Figure~\ref{fig:prompt_for_audioseg}. iii) Keystep Boundary Detection further applies keystep span detection by grouping phrases with the same keystep label together and updating the start and end timestamp for each grouped keystep. The predictions for all segments \(x_{\text{window}}\) are concatenated to form a complete trial, which is then passed to the final Keystep Boundary Detection module.

\begin{tcolorbox}[breakable]
\noindent\textbf{PROMPT: }Here is a conversation about EMS. Identify the keystep (action the first responder is taking) and its corresponding words.

Here is one example: \par
Conversation: 
Assessing patient. Hello can you hear me? No pulse, no breaths, Doing CPR, Call 911, Applying defib, Adult patient If the patient is a child, press the child button. Press pads firmly on skin. Press the pads as shown in the picture. Do not touch the patient, everyone clear, analyzing heart rhythm, shock advised. Do not touch the patient, everyone clear. Press the flashing shocking button, shock delivered. Begin CPR 10 20 30. give two breaths. Resume CPR, ten, twenty, thirty.

Let's think step by step,\par
Step 1: Token classification: classify every token in the conversation in following classes, \{classes\}. If you think the token contains multiple keysteps, you should assign all keysteps to that token.\par
Assessing: approach\_patient\par
patient: approach\_patient\par
Hello: check\_responsiveness\par
can: check\_responsiveness\par
...\par
ten: chest\_compressions\par
twenty: chest\_compressions\par
thirty: chest\_compressions 

Step 2: Span Detection: concatenate tokens with the same labels into utterances, and identify all keysteps for the utterance. Every Keystep must has its corresponding words. Return the results in the defined json format as follows, in the json file EVERY Keystep must have a list with corresponding words.\par

\begin{lstlisting}
[
 {
  "Utterance": "",
  "Keystep": 
   {
    "Keystep1": [word1, word2, ...],
    "Keystep2": [word1, word2, ...],
    "Keystep3": [word3, word4, ...]
   }
  }
]

[
 {
  "Utterance": "Assessing patient",
  "Keystep": 
   {
    "approach patient": ["Assessing", "patient"]
   }
 },
...
 {
  "Utterance": "ten twenty thirty",
  "Keystep": 
   {
    "chest compressions": ["ten", "twenty", "thirty"]
   }
 }
]

\end{lstlisting}

Here is the real conversation
Conversation: [...]\par

\end{tcolorbox}
\noindent\begin{minipage}{\linewidth}
\captionof{figure}{Prompt for audio-based keystep segmentation.}\label{fig:prompt_for_audioseg}
\end{minipage}

\noindent\textbf{Results.}
In Table~\ref{tab:segmentation_results}, while GPT-4o does not outperform the top-performing baseline, it demonstrates strong zero-shot keystep classification capabilities compared to several fully supervised models. Several factors contribute to the lower performance of the Whisper-timestamped + GPT-4o approach. First, current speech recognition models (e.g., Whisper-timestamped, Whisper-1) that provide transcripts with word-level timestamps often lack accuracy, introducing transcription and timing errors that propagate into GPT-4o’s input. Additionally, as discussed in the error analysis~\ref{sec:error_analysis}, audio input may lack relevant information or contain noise, which further impacts keystep segmentation accuracy.

\subsection{CPR quality estimation}
\label{subsec:cpr_quality_estimation}

\noindent\textbf{Task details.}
We frame this task as an online recognition task where, given a window of data limited to a context duration of 5-seconds in a sliding window approach replicating real life a procedure, the models should predict two main metrics. First, the CPR compression rate \(r\) per minute and compression depth \(d\) in mm computed per compression and then averaged over the window. Given an input window \( x_{\text{window}} \in \mathbb{R}^{T \times D} \), where \( T \) is the temporal length of the window and \( D \) is the feature dimension, the objective is to output the \(r\) and  \(d\). It should be noted that, in this task, T is limited to 5-seconds. 

\begin{figure}[!h]
\centering
    \includegraphics[width=0.6\linewidth]{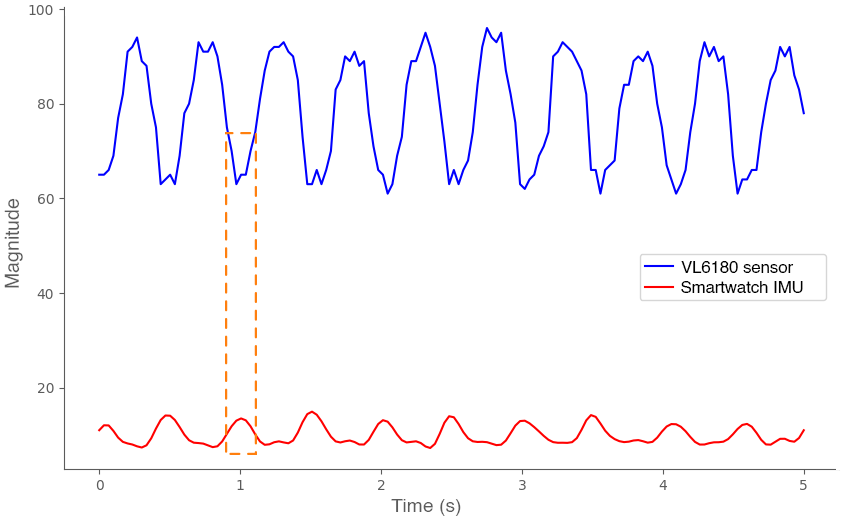}
\caption{Hand movement recorded through the smartwatch closely resembles the compression data from the sensor embedded in the manikin.}
\label{fig:data_peak_detection}
\end{figure}

\noindent\textbf{Annotations.}
The VL6180X sensor embedded in the manikin serves as the ground truth source for measuring CPR compression depth and rate. Peaks and valleys in the sensor data are detected using the Python SciPy library \cite{Virtanen2020SciPyPython}, as illustrated in Figure~\ref{fig:data_peak_detection}. The average number of compressions over a given time period is determined by counting the detected peaks and valleys, with the compression rate calculated as the number of compressions divided by the elapsed time. For average ground truth depth, we calculate the difference between consecutive peak and valley depths and then compute the mean.

Figure \ref{fig:data_peak_detection} also highlights the alignment between the responder’s hand movements and the CPR procedure, demonstrating the precise time synchronization of our multimodal data. Each valley detected in the VL6180X depth data corresponds to a chest compression on the manikin, while the responder’s hand exhibits a significant acceleration peak in the smartwatch accelerometer data, confirming synchronized multimodal recording.

To ensure fair evaluation for all settings, we utilized the test split of the dataset for both supervised and unsupervised approaches.

\noindent\textbf{Evaluation metrics and rule-based feedback framework.}
To evaluate the performance of our CPR quality assessment system, we adopt a combined quantitative and qualitative approach. Specifically, we calculate the root mean square error (RMSE) for the predicted compression rate and depth to assess the assistant’s numerical accuracy. In addition, we evaluate feedback accuracy using explicitly defined rules for compression rate and depth, derived from established clinical guidelines and refined through expert input from EMS professionals which provides more interpretable evaluation of the assistants performance.

\noindent\textit{Compression rate feedback.}
Let $r \in \mathbb{R}^+$ denote the compression rate in compressions per minute (cpm). We define the optimal range as $[100, 120]$ \cite{AmericanRedCross2022Steps}. The feedback function for rate, $f_r(r)$, is given by:
\[
f_r(r) = 
\begin{cases}
\text{``Go faster''}, & \text{if } r < 100, \\
\text{``Doing great''}, & \text{if } 100 \leq r \leq 120, \\
\text{``Go slower''}, & \text{if } r > 120.
\end{cases}
\]

\noindent\textit{Compression depth feedback.}
Let $d \in \mathbb{R}^+$ denote the compression depth in centimeters (cm), with the optimal range defined as $[5, 6]$ \cite{AmericanRedCross2022Steps}. The feedback function for depth, $f_d(d)$, is defined as:
\[
f_d(d) = 
\begin{cases}
\text{``Compress deeper''}, & \text{if } d < 5, \\
\text{``Doing great''}, & \text{if } 5 \leq d \leq 6, \\
\text{``Too deep''}, & \text{if } d > 6.
\end{cases}
\]

\noindent\textit{Feedback classification logic.}
We evaluate feedback correctness using standard classification terminology. Let $g$ denote the ground-truth feedback and $a$ the assistant-generated feedback. We define:
\[
\begin{aligned}
\text{True Positive (TP):} & \quad g \neq \text{``Doing great''} \land a = g. \\
\text{True Negative (TN):} & \quad g = \text{``Doing great''} \land a = g. \\
\text{False Positive (FP):} & \quad g = \text{``Doing great''} \land a \neq g. \\
\text{False Negative (FN):} & \quad g \neq \text{``Doing great''} \land a \neq g.
\end{aligned}
\]

Based on these definitions, we compute standard performance metrics such as Precision, Recall, F1-score and feedback accuracy. This formulation ensures rigorous, interpretable evaluation of CPR guidance feedback, aligning with established clinical standards.

\subsubsection{Egocentric Setting}
In this setting, the model inputs are egocentric video data from a body-worn GoPro camera, comprising a sequence of RGB frames. Let the RGB input during the 5-second window \( x_{\text{window}} \) be denoted as \( \mathbf{Ego}^{\text{RGB}}_x = \{I_t\}_{t=1}^{T} \), where \( I_t \) is the RGB frame at time step \( t \), and \( T \) is the number of frames in the window.

To detect hand movements, we extract bounding boxes of the most prominent hand from the frames in \( \mathbf{Ego}_{x} \) using Grounding DINO~\cite{liu2023grounding}. This cropped hand region is then processed with Google Mediapipe \cite{lugaresi2019mediapipe} to detect hand joint locations. The wrist position, \( \text{Wrist}_{xy}^t \), is extracted from the joint positions in each frame \( t \). We apply a Butterworth low-pass filter with a 2Hz cut-off frequency and an order of 4 is applied to smooth the wrist trajectory and remove high-frequency noise. Compression rate \( r \) is then calculated by performing peak detection on the \( y \)-coordinate (representing vertical wrist movement in 2D) and scaling to a per-minute rate.

For depth estimation, we present both a supervised and unsupervised approach for indepth exploration of this task. 

\noindent\textbf{Unsupervised depth estimation.}  
In this approach, we calculate the distance between the detected peaks and valleys of the wrist coordinates within the window and then scale that to real-world dimensions utilizing a hyper-parameter that needs to be tuned using trial and error. This parameter depends on several factors such as camera resolution, mounted location of the body. 

\begin{figure}[!h]
    \centering
    \includegraphics[width=0.6\linewidth]{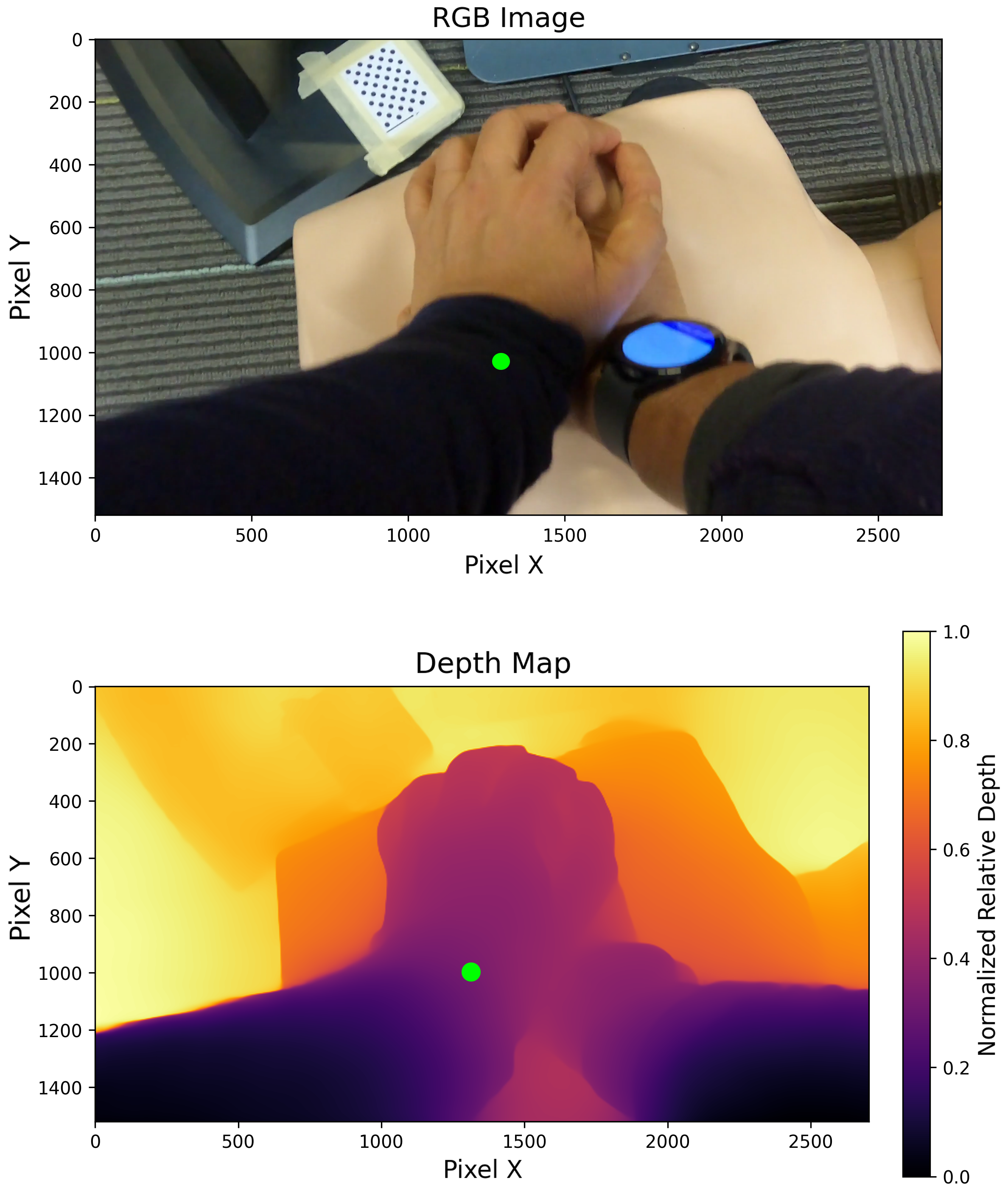}
    \caption{Comparison of the RGB frame with the detected wrist keypoint (green marker) and the estimated normalized depth map using the MiDas\cite{lasinger2019towards} model for the egocentric view during CPR procedure.}
    \label{fig:midas_depth_estimation}
\end{figure}

\noindent\textbf{Supervised depth estimation.}  
For this approach, we employ a supervised depth estimation model to estimate the depth of RGB frames (see Figure \ref{fig:midas_depth_estimation}). Specifically, we utilize a \textit{monocular depth estimation method} \cite{lasinger2019towards} to obtain a relative depth map, which we scale to absolute depth using a reference object with known real-world depth. Further, we evaluate the large and small variants of the MiDas model introduced in \cite{lasinger2019towards}.

Given an absolute depth map \(Z(u, v)\) at pixel coordinates \((u, v)\), we recover the 3D coordinates in the camera coordinate system using the camera intrinsic matrix:

\begin{equation}
X = \frac{(u - c_x) Z}{f_x}, \quad Y = \frac{(v - c_y) Z}{f_y}, \quad Z = Z(u, v)
\end{equation}

where \((u, v)\) are pixel coordinates in the image plane, \((X, Y, Z)\) are real-world 3D coordinates in the camera coordinate system, \(f_x\) and \(f_y\) are the focal lengths in pixels and \((c_x, c_y)\) is the principal point of the camera.

Once the \emph{3D coordinates} of the detected \emph{peak} and \emph{valley} positions are reconstructed, the \emph{compression depth} for each peak--valley pair is first computed as the standard \emph{Euclidean distance}:

\begin{equation}
d_{\text{3D}} = \left\| \mathbf{P} - \mathbf{V} \right\|_2 
\label{eq:depth_calculation_1}
\end{equation}
\begin{equation}
d_{\text{3D}} = \sqrt{(X_{p} - X_{v})^2 + (Y_{p} - Y_{v})^2 + (Z_{p} - Z_{v})^2},
\label{eq:depth_calculation_3d}
\end{equation}

where $\mathbf{P} = \bigl(X_{p},\, Y_{p},\, Z_{p}\bigr)$ and 
$\mathbf{V} = \bigl(X_{v},\, Y_{v},\, Z_{v}\bigr)$ are the estimated 3D positions of the \emph{peak} and \emph{valley}, respectively.

Finally, we compute the \textit{mean compression depth} $D$ over the entire analysis window for both measures. In the 3D case:
\begin{equation}
D_{\text{3D}} = \frac{1}{N} \sum_{i=1}^{N} d_{\text{3D},\,i},
\end{equation}

where $N$ is the number of detected compression cycles within the window. Comparing $D_{\text{3D}}$ and $D_{Z}$ illustrates the effect of lateral ($X,Y$) variation on the measured compression depth.

\begin{table*}[!h]
\centering
\small
\setlength{\tabcolsep}{1mm}
\resizebox{\columnwidth}{!}{%
\begin{tabular}{@{}llcccccc@{}}
\toprule
\multirow{2}{*}{\textbf{Input Configuration}} & \multirow{2}{*}{\textbf{Depth Estimation Method}} & \multicolumn{2}{c}{\textbf{RMSE}~(\(\downarrow\))} & \multicolumn{2}{c}{\textbf{Rate Feedback~(\(\uparrow\))}} & \multicolumn{2}{c}{\textbf{Depth Feedback~(\(\uparrow\))}} \\ \cmidrule(l){3-8} 
 &  & \textbf{Rate (cpm)} & \textbf{Depth (mm)} & \textbf{F1} & \textbf{Acc} & \textbf{F1} & \textbf{Acc} \\ \midrule
\textbf{Egocentric (RGB)} & {MiDaS (monocular) \cite{lasinger2019towards}} & 28.56 & 32.05 & 0.47 & 0.52 & 0.72 & 0.63 \\ 
\textbf{Egocentric (RGB)} & Heuristic (SP) & 28.56 & 25.78 & 0.47 & 0.52 & 0.77 & 0.70 \\ 
\textbf{IMU (Acc)} & Supervised + SP & \textbf{14.89} & 12.13 & 0.51 & 0.64 & \textbf{0.83} & 0.77 \\ 
\textbf{Fusion (Supervised)} & Supervised + SP & 15.66 & \textbf{11.78} & 0.52 & 0.62 & \textbf{0.83} & \textbf{0.77} \\ 
\textbf{Fusion (Simple) $\bigstar$} & Supervised + SP & 19.33 & 18.35 & 0.54 & 0.64 & 0.54 & 0.62 \\ \midrule
% \textbf{Exo (RGBD)} & Heuristic (SP) & 22.74 & 33.11 & 0.57 & 0.64 & 0.77 & 0.70 \\ 
\bottomrule
\end{tabular}%
}
\caption{Quantitative comparison of average CPR rate and depth estimation across different input configurations and depth estimation methods for all subjects for the test split. RMSE errors are reported along with feedback performance metrics (F1 and accuracy). SP = signal processing heuristic. $\bigstar$: Simple fusion utilizes a 50:50 late fusion of Egocentric and IMU modalities.}
\label{tab:cpr_results}
\end{table*}

\noindent\textbf{Results.} Table~\ref{tab:cpr_results} presents the performance of both supervised and heuristic methods for estimating CPR compression depth and rate from egocentric video. For compression depth estimation, the supervised approach using the MiDaS model~\cite{lasinger2019towards} yields an RMSE of 32.05mm, while the heuristic method achieves a lower RMSE of 25.78mm. The improved accuracy of the heuristic approach is likely due to its manual calibration process, which provides a more stable conversion from image-space wrist coordinates to depth values.

\begin{figure}[!ht]
    \centering
    \includegraphics[width=0.5\linewidth]{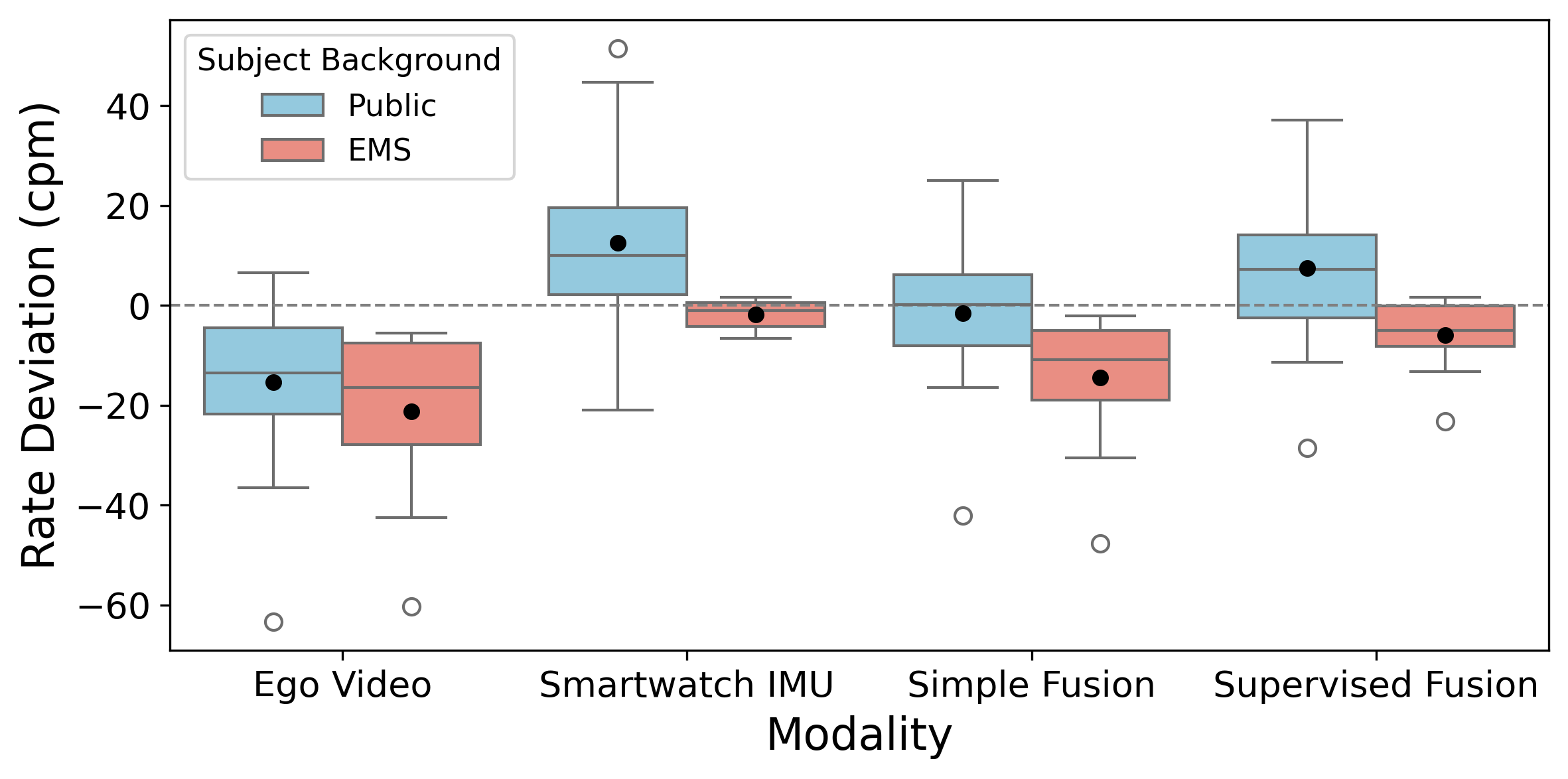}
    \caption{Average deviation of predicted CPR compression rates from the ground truth across different modalities, separated by subject background (Public vs. EMS), on the test split of the dataset.}
    \label{fig:sw_vision_cpr_rate_deviation_comparison}
\end{figure}

Figure~\ref{fig:sw_vision_cpr_rate_deviation_comparison} illustrates the distribution of CPR compression rate prediction errors across modalities, segmented by subject background. Egocentric video shows the largest negative deviations, particularly for EMS responders, indicating that video only estimates tend to underestimate the true compression rate. These errors stem from the complexity of multi-responder simulations, where wrist detectors often track the wrist of a secondary responder instead of the primary responder. Occlusions and moments when the field of view moves away from the manikin during team communication further disrupt the peak‑detection process used to infer compression rates.

\begin{figure}[!ht]
    \centering
    \includegraphics[width=0.5\linewidth]{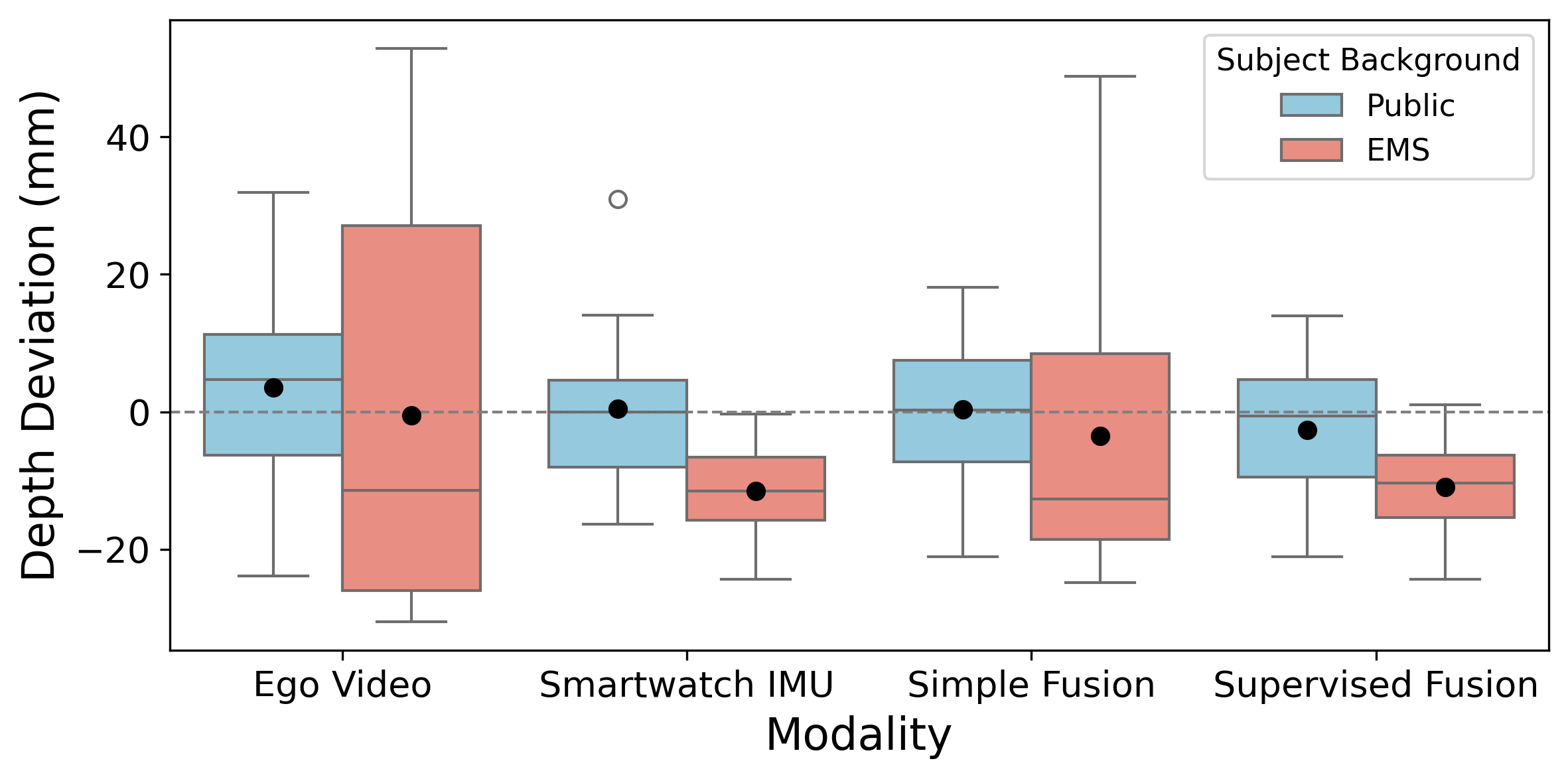}
    \caption{Average deviation of predicted CPR compression depths from the ground truth across different modalities, separated by subject background (Public vs. EMS), on the test split of the dataset.}
    \label{fig:sw_vision_cpr_depth_deviation_comparison}
\end{figure}

A similar trend is observed in compression depth estimation, where both monocular depth estimation and signal‑processing approaches struggle in the egocentric setting. The relative motion between the camera wearer and the hands performing compressions introduces lateral noise in the X and Y axes during wrist localization, which propagates into depth errors. As shown in Figure~\ref{fig:sw_vision_cpr_depth_deviation_comparison}, egocentric video alone produces the largest deviations with wide variability and frequent large errors for EMS responders. These challenges are exacerbated by the same multi‑responder dynamics noted above, and further compounded by the difficulty of obtaining consistent scene geometry with MiDaS monocular depth prediction in cluttered and dynamic environments, which contributes to higher supervised error.

In addition to estimating rate and depth, we evaluate the downstream performance of CPR feedback generation using our rule-based feedback framework. For compression rate feedback, both the supervised and heuristic approaches yield an F1 score of 0.47 and a feedback accuracy of 0.52. For compression depth feedback, the heuristic approach achieves higher performance, consistent with its superior depth estimation accuracy, resulting in more reliable feedback generation.

% --------------------------------------------------- %
\subsubsection{Smartwatch IMU-Based Setting}
We estimate CPR compression rate and depth using motion data captured by the participant’s smartwatch. Specifically, we employ a model composed of two convolutional layers followed by a unidirectional LSTM to predict CPR metrics from normalized IMU signals. 

Let the 3-axis accelerometer data collected during the 5-second window \( x_{\text{window}} \) be denoted as \( \mathbf{SW}^{\text{IMU}}_x = \{\mathbf{s}_t\}_{t=1}^{T} \), where \( \mathbf{s}_t \in \mathbb{R}^3 \) represents the accelerometer reading \((x_t, y_t, z_t)\) at time step \( t \). Ground-truth compression depths are obtained from the VL6180X proximity sensor embedded in the manikin.
The model predicts a depth estimate \( \hat{d} \) for each compression detected within the window. The training objective minimizes the mean squared error (MSE) between the predicted and ground-truth depths:
\[
\mathcal{L}_{\text{depth}} = \left(\hat{d} - d_{\text{gt}}\right)^2
\]

To estimate the compression rate \( r \), we apply a signal processing pipeline. First, the magnitude of the accelerometer signal is computed as \( M_t = \sqrt{x_t^2 + y_t^2 + z_t^2} \). This signal is then passed through a 4th-order Butterworth low-pass filter with a cutoff frequency of 2\,Hz to remove high-frequency noise. Compression rate is determined by counting the number of peaks in the filtered signal over the 5-second window and scaling to compressions per minute:
\[
r = \frac{\text{number of peaks}}{5} \times 60
\]

This procedure yields a robust per-minute estimate of CPR compression rate for each window. \\

\noindent\textbf{Results.}  
Table~\ref{tab:cpr_results} summarizes the performance of compression depth and rate estimation across subjects. The smartwatch IMU-based method achieves the best overall performance for compression rate estimation, yielding the lowest RMSE, and also ranks second in compression depth accuracy. This can be attributed to the smartwatch's ability to capture fine-grained wrist motion patterns during CPR, providing an isolated and consistent signal for CPR quality estimation.

Figures~\ref{fig:sw_vision_cpr_rate_deviation_comparison} and \ref{fig:sw_vision_cpr_depth_deviation_comparison} present the average errors in compression rate and depth estimation for different modalities, grouped by participant background. 
In contrast to egocentric video based estimation, smartwatch IMU based method consistently achieve low errors for both EMS responders and general public participants. 
This improved performance stems from the direct and robust measurement of wrist motion by the IMU, which is largely unaffected by the occlusions, field-of-view limitations, and other challenges that hinder egocentric video based approaches. This is further evident from the substantial reduction in compression rate error for EMS responders when using IMU data compared to egocentric video.
Following a similar trend, the smartwatch IMU based method achieves the highest feedback accuracy for compression rate at 0.64, and for compression depth at 0.77, outperforming other modalities in both metrics.

\subsubsection{Multimodal Fusion}  
\begin{figure}[!h]
    \centering
    \includegraphics[width=0.6\linewidth]{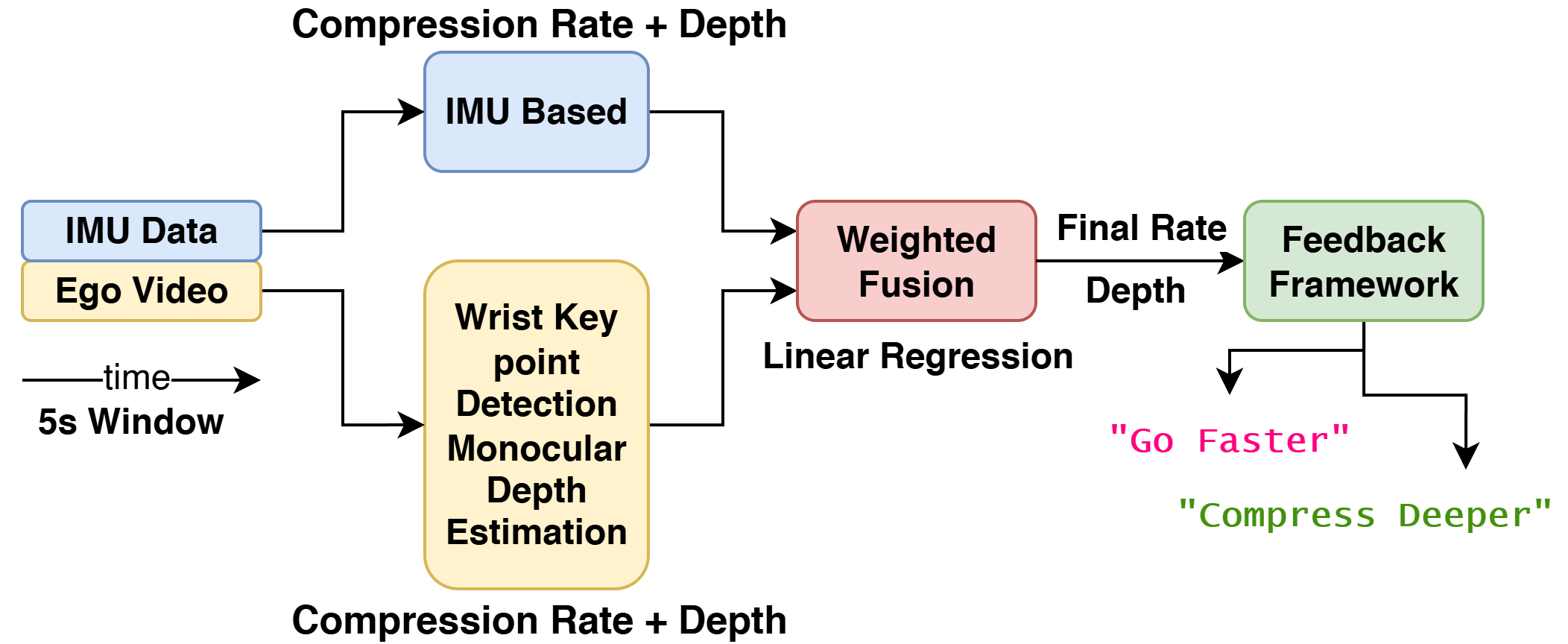}
    \caption{Multimodal fusion approach for CPR quality estimation.}
    \label{fig:multimodal_fusion_cpr_estimation}
\end{figure}

In this setting, we fuse egocentric video with smartwatch IMU data for CPR quality estimation. This is enabled by the time-synchronized nature of our dataset, allowing for a realistic evaluation of ICA’s ability to leverage multimodal information to assist responders. We implement a supervised late fusion strategy in which fusion weights between modalities are learned during training. As a baseline, we also evaluate a simple late fusion method that averages predictions from the egocentric and smartwatch streams with equal weights (50-50), as illustrated in Figure~\ref{fig:multimodal_fusion_cpr_estimation}.

Formally, the model receives synchronized egocentric video \( \mathbf{Ego}^{\text{RGB}}_x \) and smartwatch IMU data \( \mathbf{SW}^{\text{IMU}}_x \) for each 5-second window \( x_{\text{window}} \), and must estimate the compression rate \( r \), compression depths \( \hat{\mathbf{d}} \), and feedback labels. Let the predictions from the egocentric and smartwatch models be denoted as:
\[
r_{\text{ego}}, \hat{\mathbf{d}}_{\text{ego}} \quad \text{and} \quad r_{\text{imu}}, \hat{\mathbf{d}}_{\text{imu}}
\]

In the simple late fusion baseline, the final predictions are computed as the average of the two modalities:
\[
r_{\text{fused}} = \frac{1}{2}(r_{\text{ego}} + r_{\text{imu}}), \quad \hat{\mathbf{d}}_{\text{fused}} = \frac{1}{2}(\hat{\mathbf{d}}_{\text{ego}} + \hat{\mathbf{d}}_{\text{imu}})
\]

In the supervised late fusion setting, we learn scalar weights \( \alpha \in [0,1] \) and \( \beta \in [0,1] \) for rate and depth respectively, subject to \( 1 - \alpha \) and \( 1 - \beta \) for the complementary modality. The fused predictions are given by:
\[
r_{\text{fused}} = \alpha \cdot r_{\text{ego}} + (1 - \alpha) \cdot r_{\text{imu}}, \quad 
\hat{\mathbf{d}}_{\text{fused}} = \beta \cdot \hat{\mathbf{d}}_{\text{ego}} + (1 - \beta) \cdot \hat{\mathbf{d}}_{\text{imu}}
\]

These fused estimates are then used to compute window-level CPR feedback. The fusion weights \( \alpha \) and \( \beta \) are learned jointly with the task loss during training to optimize final prediction accuracy.

\noindent\textbf{Results.}
Table~\ref{tab:cpr_results} shows that multimodal fusion yields strong performance across both CPR rate and depth estimation tasks. The supervised fusion model achieves the best overall depth estimation RMSE of 11.78mm, outperforming both ego-only and IMU-only settings. It also matches the highest depth feedback F1 score of 0.83 and accuracy of 0.77, demonstrating effective integration of spatial and inertial cues. For rate estimation, it achieves an RMSE of 15.66cpm, closely approaching the IMU-only baseline, while improving significantly over ego-only models. The simple 50:50 fusion baseline also performs competitively, particularly in rate feedback, where it achieves the highest F1 score of 0.54. However, its depth estimation and feedback metrics are lower than those of the supervised fusion model, highlighting the benefit of learning optimal fusion weights.

These quantitative results are further supported by the box plots in Figures~\ref{fig:sw_vision_cpr_rate_deviation_comparison} and \ref{fig:sw_vision_cpr_depth_deviation_comparison}, which illustrate deviations in predicted CPR compression rate and depth across modalities and participant backgrounds. For both metrics, fusion methods exhibit consistently lower deviations compared to unimodal egocentric video, especially for EMS responders, whose results with ego video alone show large variability and frequent extreme errors. The addition of IMU data stabilizes predictions, reducing errors caused by wrist occlusions, camera motion, and confusion with secondary responders in multi-responder scenes.

Overall, these results confirm that supervised multimodal fusion provides complementary gains over unimodal inputs, particularly when compared to egocentric video alone, resulting in more accurate and robust CPR quality estimation. While smartwatch IMU alone already delivers strong performance, fusion helps mitigate modality-specific weaknesses and improves consistency across participants. Future work should explore more advanced fusion architectures, such as cross-modal attention, temporal modeling, and alignment mechanisms, to further enhance the integration of visual and inertial signals in real-time CPR assessment.

\bibliographystyle{unsrtnat}
\bibliography{references}  %%% Uncomment this line and comment out the ``thebibliography'' section below to use the external .bib file (using bibtex) .

\end{document}